\pdfoutput=1

\documentclass[11pt]{article}
\usepackage{float}

\usepackage[utf8]{inputenc}
\usepackage{pgfplots}
\usepackage{inconsolata}
\usepackage{booktabs}
\usepackage{multirow}
\usepackage{amsmath}
\usepackage{amsfonts}
\usepackage{amsthm}
\usepackage{pifont}
\usepackage{xspace}
\usepackage{graphicx}
\usepackage{bm}
\usepackage{xcolor, colortbl}
\usepackage[most]{tcolorbox}
\usepackage{csquotes}
\usepackage{anyfontsize}
\usepackage{comment}
\usepackage{float}
\usepackage{cuted}
\usepackage{tikz}
\usepackage{listings,multicol}
\usepackage{filecontents}
\usepackage{dsfont}
\DeclareUnicodeCharacter{2212}{−}
\usepgfplotslibrary{groupplots,dateplot}
\usetikzlibrary{patterns,shapes.arrows}
\pgfplotsset{compat=newest}

\usepackage{tikzscale}
\usepackage{amsmath}
\newcommand{\probP}{\text{I\kern-0.15em P}}

\newcommand{\specialcellleft}[2][l]{%
\begin{tabular}[#1]{@{}l@{}}#2\end{tabular}}

\usepackage{multirow, colortbl}

\usepackage{tabularx,booktabs}
\usepackage{makecell}

\usepackage[normalem]{ulem}
\useunder{\uline}{\ul}{}

\usepackage{graphicx}

\definecolor{ablation6}{HTML}{fcefed}
\definecolor{ablation_tie}{HTML}{fce3e1}

\definecolor{ablation5}{HTML}{fcd8d4}
\definecolor{ablation4}{HTML}{FBC3BC}
\definecolor{ablation3}{HTML}{F7A399}
\definecolor{ablation2}{HTML}{F38375}
\definecolor{ablation1}{HTML}{EF6351}
\definecolor{UMDred}{HTML}{ed1c24}
\definecolor{UMDyellow}{HTML}{ffc20e}
\definecolor{CustomGreen}{HTML}{1FC801}

\usepackage[]{acl2023}

\usepackage{xspace}
\usepackage[utf8]{inputenc}
\usepackage{pgfplots}
\usepackage{dsfont}
\DeclareUnicodeCharacter{2212}{−}
\usepgfplotslibrary{groupplots,dateplot}
\usetikzlibrary{patterns,shapes.arrows}
\pgfplotsset{compat=newest}

\definecolor{bggray}{rgb}{0.95, 0.95, 0.95}
\usepackage[%
    framemethod=tikz,
    skipbelow=\topskip,
    skipabove=\topskip
]{mdframed}
\mdfsetup{%
    leftmargin=0pt,
    rightmargin=0pt,
    backgroundcolor=bggray,
    middlelinecolor=black,
    roundcorner=3
}

\newtcolorbox[list inside=prompt,auto counter,number within=section]{prompt}[1][]{
    colbacktitle=black!60,
    fonttitle=\small,
    coltitle=white,
    fontupper=\footnotesize,
    boxsep=4pt,
    left=0pt,
    right=0pt,
    top=0pt,
    bottom=0pt,
    boxrule=1pt,
    width=\textwidth, 
    enlarge left by=0mm, 
    enlarge right by=0mm, 
    #1,
}

\usepackage[%
    framemethod=tikz,
    skipbelow=\topskip,
    skipabove=\topskip
]{mdframed}
\mdfsetup{%
    leftmargin=0pt,
    rightmargin=0pt,
    backgroundcolor=white,
    middlelinecolor=black,
    roundcorner=3
}

\newtcolorbox[list inside=prompt,auto counter,number within=section]{summary}[1][]{
    colbacktitle=blue!10,
    fonttitle=\small,
    coltitle=black,
    fontupper=\footnotesize,
    boxsep=4pt,
    left=0pt,
    right=0pt,
    top=0pt,
    bottom=0pt,
    boxrule=1pt,
    width=\textwidth, 
    enlarge left by=0mm, 
    enlarge right by=0mm, 
    #1,
}

\usepackage{tikzscale}
\usepackage{amsmath,amssymb}

\newcommand{\model}{\textsc{MoDS}\xspace}
\newcommand{\modelAll}{\textsc{MoDS}-\textit{All}\xspace}

\newcommand{\modelTopic}{\textsc{MoDS}-\textit{Topic}\xspace}

\usepackage{multirow, colortbl}

\usepackage{tabularx,booktabs}
\usepackage{makecell}

\usepackage[normalem]{ulem}
\useunder{\uline}{\ul}{}

\usepackage{graphicx}

\definecolor{yellowcite}{HTML}{e3be05}

\definecolor{ablation6}{HTML}{fcefed}
\definecolor{ablation_tie}{HTML}{fce3e1}

\definecolor{ablation5}{HTML}{fcd8d4}
\definecolor{ablation4}{HTML}{FBC3BC}
\definecolor{ablation3}{HTML}{F7A399}
\definecolor{ablation2}{HTML}{F38375}
\definecolor{ablation1}{HTML}{EF6351}

\usepackage[]{algpseudocode}
\usepackage[]{algorithm}
\usepackage{float}
\algtext*{EndFor}%
\algtext*{EndProcedure}%

\usepackage{booktabs}
\usepackage[normalem]{ulem}
\useunder{\uline}{\ul}{}

\usepackage{times}
\usepackage{latexsym}
\usepackage{adjustbox}

\usepackage[T1]{fontenc}
\usepackage{amsmath}
\usepackage{amssymb}
\usepackage{booktabs}
\usepackage{tikzscale}
\usepackage{amsmath}

\usepackage{multirow, colortbl}

\usepackage{tabularx,booktabs}
\usepackage{makecell}
\usepackage{multirow}
\usepackage{scalerel,xparse}
\usepackage[utf8]{inputenc}

\usepackage{cleveref}
\usepackage{microtype}
\usepackage{enumitem}
\crefformat{section}{\S#2#1#3}
\crefformat{subsection}{\S#2#1#3}
\crefformat{subsubsection}{\S#2#1#3}

\definecolor{UMDred}{HTML}{ed1c24}

\title{\model: Moderating a Mixture of Document Speakers\\to Summarize Debatable Queries in Document Collections }

\author{Nishant Balepur$^{1,}$$^{2*}$ \hspace{0.5cm} Alexa Siu$^{2}$  \hspace{0.5cm} \textbf{Nedim Lipka}$^{2}$ \hspace{0.5cm} \textbf{Franck Dernoncourt}$^{2}$ \\ \hspace{0.5cm} \textbf{Tong Sun}$^{2}$ \hspace{0.5cm} \textbf{Jordan Boyd-Graber}$^{1}$ \hspace{0.5cm} \textbf{Puneet Mathur}$^{2\dagger}$ \vspace{0.1cm}  \\
  $^{1}$University of Maryland \hspace{0.5cm}
  $^{2}$Adobe Research \hspace{0.5cm} \vspace{0.1cm} \\
  \texttt{nbalepur@umd.edu} \hspace{0.5cm} \texttt{{puneetm}@adobe.com}
}

\begin{document}
\maketitle
\renewcommand{\thefootnote}{\fnsymbol{footnote}}
\footnotetext[1]{Work done during internship at Adobe.}
\footnotetext[2]{Primary internship mentor}
\renewcommand{\thefootnote}{\arabic{footnote}}

\begin{abstract} {
Query-focused summarization (QFS) gives a summary of documents to answer a query.
Past QFS work assumes queries have one answer, ignoring debatable ones (\textit{Is law school worth it?}).
We introduce \textbf{Debatable} \textbf{QFS} \textbf{(DQFS)}, a task to create summaries that answer debatable queries via documents with opposing perspectives; summaries must \textit{comprehensively cover} all sources and \textit{balance perspectives}, favoring no side.
These goals elude LLM QFS systems, which: 1) lack structured content plans, failing to guide LLMs to write balanced summaries, and 2) use the same query to retrieve contexts across documents, failing to cover all perspectives specific to each document's content.
To overcome this, we design \model, a multi-LLM framework mirroring human panel discussions.
\model treats documents as individual Speaker LLMs and has a Moderator LLM that picks speakers to respond to tailored queries for planned topics.
Speakers use tailored queries to retrieve relevant contexts from their documents and supply perspectives, which are tracked in a rich outline, yielding a content plan to guide the final summary.
Experiments on ConflictingQA with controversial web queries and DebateQFS, our new dataset of debate queries from Debatepedia, show \model beats SOTA by 38-59\% in topic paragraph coverage and balance, based on new citation metrics. Users also find \model's summaries to be readable and more balanced.\footnote{Code and data will be released on the Adobe Research Github after internal approval: \url{https://github.com/adobe-research}}
}
\end{abstract}

\section{Introduction}
\definecolor{myblue1}{HTML}{a5c3f0}
\definecolor{myred1}{HTML}{e99999}
\definecolor{myyellow1}{HTML}{f8d862}
\begin{figure}
    \centering
    \includegraphics[width=\linewidth]{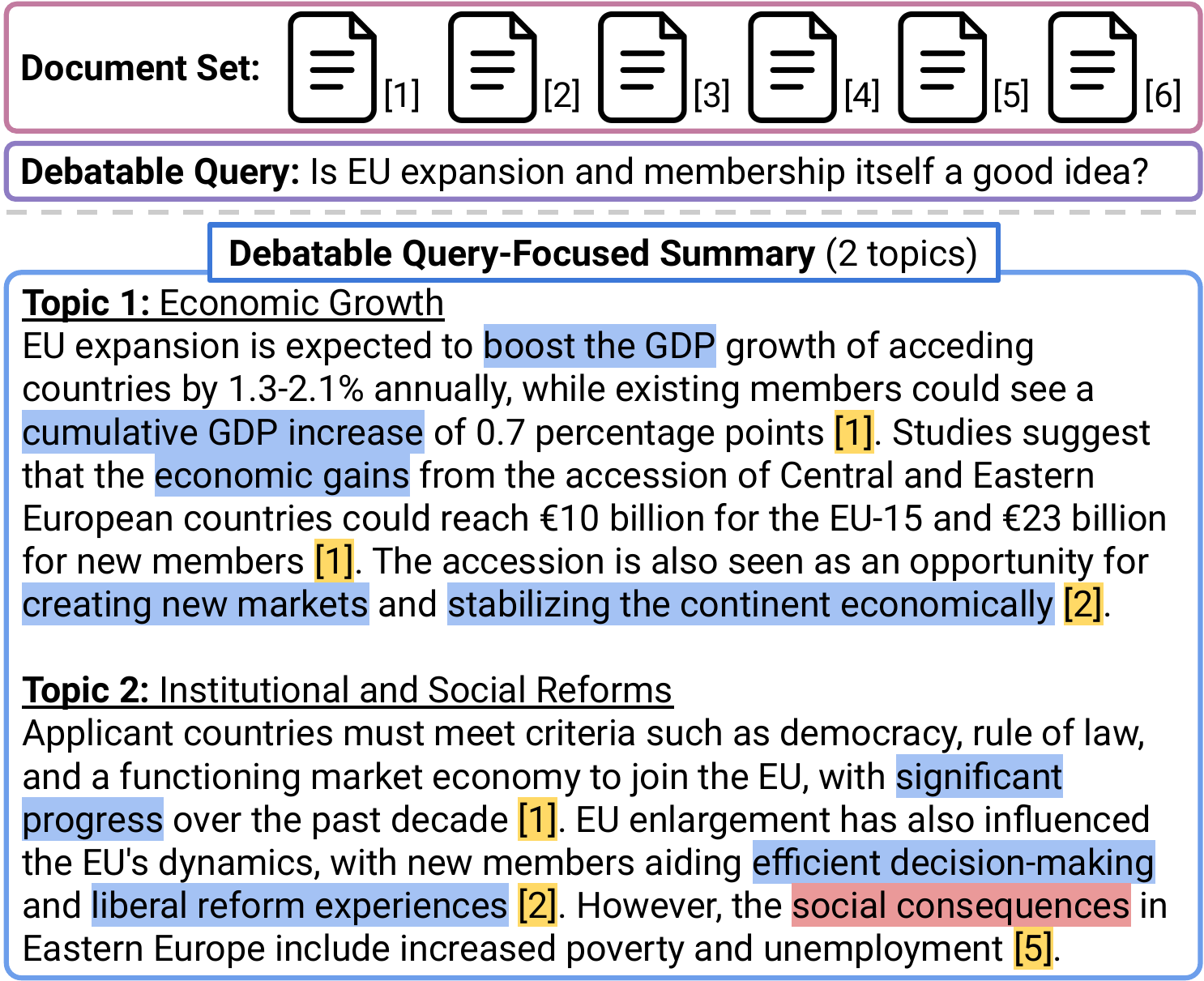}
    \caption{\label{fig:intro} Debatable Query-Focused Summarization (DQFS) with GPT-4 for two topics. The model mainly gives ``yes'' perspectives (\colorbox{myblue1}{Blue}) with few ``no'' perspectives (\colorbox{myred1}{Red}), giving an unbalanced summary. It also has poor coverage, failing to cite (\colorbox{myyellow1}{Yellow}) half the inputs.}
\end{figure}
Query-focused summaries (QFS) give an overview of documents to answer a query~\cite{rosner2008multisum, el2021automatic}.
By combining each document's content useful for answering the query, or their \textbf{perspectives}~\cite{lin2006side}, these summaries can aid decision-making~\cite{hsu2021decision}.
For example, doctors pick treatments based on research paper perspectives~\cite{goff2008patients} and legislators vote based on perspectives in policy reports~\cite{jones1994reconceiving}. 
Past QFS work assumes documents have aligned perspectives~\cite{roy2023review}, but some queries, like ``\emph{Is law school worth it?}'', are debatable, containing opposing perspectives~\cite{wan2024evidence}.
In such cases, it is key to \textit{balance} perspectives from \textit{diverse} sources so users consider all sides before deciding~\cite{dale2015heuristics}.

To address this gap, we propose \textbf{\textit{debatable} QFS (DQFS}).
As input, DQFS uses documents and a debatable query, defined as a yes/no query where documents have opposing, equally-valid\footnote{This is meant to avoid input questions like ``Is the earth flat?'' where ``yes'' and ``no'' are not equally-valid (\cref{subsection:ethics}).} ``yes'' and ``no'' perspectives (Fig~\ref{fig:intro}).
Such queries are broad (\textit{Is law school worth it?}), and decomposing broad concepts into more specific topics (\textit{cost}, \textit{job market}) improves comprehension~\cite{johnson1983mental}.
Thus, DQFS creates a multi-aspect summary, with each paragraph covering one of an input number of topics ($2$ in Fig~\ref{fig:intro}).
The full summary and each paragraph must be \textit{comprehensive} and \textit{balanced}~(\cref{section:task}).
Comprehensive text has perspectives from all documents, while balanced text is not skewed towards the yes or no perspectives; our goals aid informed, unbiased decision-making~\cite{ziems2024measuring}.

While LLMs are deft summarizers~\cite{zhang2024benchmarking}, they cannot directly solve DQFS, as they fail to use diverse sources~\cite{huang-etal-2024-embrace}.
In Figure~\ref{fig:intro}, GPT-4 mainly gives perspectives favoring EU expansion (\textcolor{blue}{\textbf{blue}}), yielding a biased output.
Also, when asked for citations~\cite{huang-chang-2024-citation}, GPT-4 only cites 3/6 (\textcolor{yellowcite}{\textbf{yellow}}), missing half the documents' perspectives.
We intuit this arises since GPT-4 uses one inference step, with all documents in a single prompt.
This can omit document perspectives in certain positions of the prompt~\cite{liu2024lost} or that oppose parametric memory~\cite{jin2024tug}, reducing output coverage and balance.

Multi-LLM summarizers~\cite{chang2024booookscore, adams2023sparse}, which use LLMs to summarize documents individually into intermediate outputs before merging them with another LLM call, are better choices, as they represent documents more equally. 
However, they have two key issues.
\textbf{First}, they use the same topic or query as input to summarize each document, which is subpar if we wish to use retrieval in summarization to reduce LLM costs.
Queries unaligned to a document's unique content and expertise will fail to retrieve all of its most relevant contexts~\cite{sachan2022improving}; this reduces the total number of perspectives in the intermediate output, resulting in lower coverage.
\textbf{Second}, their intermediate outputs are unstructured, free-form texts, which are hard for the LLM to combine into a final output.
Free-form text needs extra reasoning to extract, classify, and compare the texts' perspectives~\cite{barrow2021syntopical}, steps that distract from the final goal of generating a balanced summary.

\begin{figure*}
    \centering
    \includegraphics[width=\linewidth]{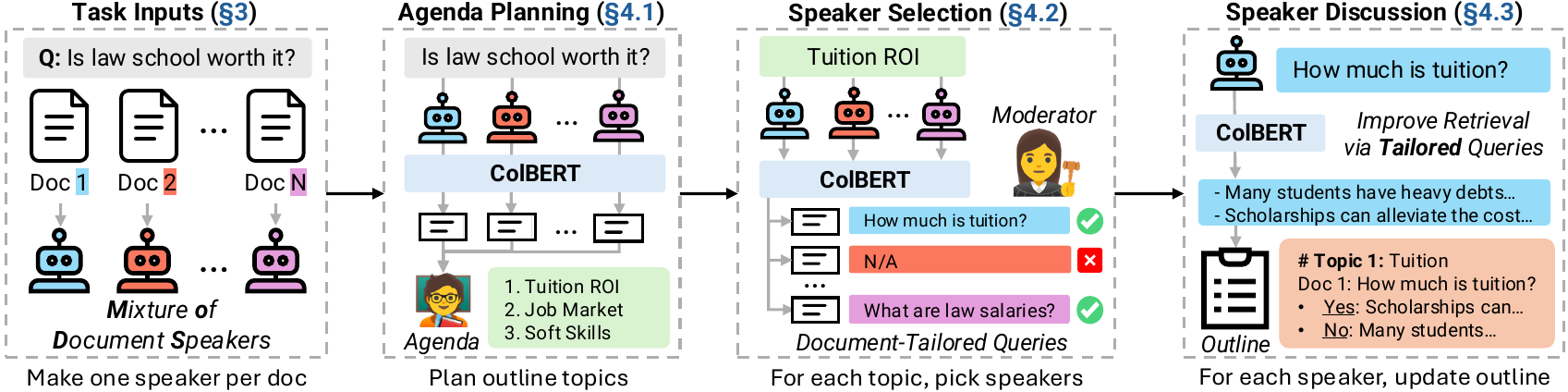}
    \caption{\label{fig:model} \small Using a debatable query and documents as inputs, \model creates an outline of document perspectives via a panel discussion among LLM speakers. First, an Agenda Planner drafts topics for the outline. A Moderator picks speakers for these topics and tailors a query for each speaker. The speakers retrieve contexts for the tailored query and use these contexts to provide their document's perspectives, which are tracked in an outline; this outline is used as a content plan to write the final summary.}
\end{figure*}

To solve our issues, we build \textbf{\model} (Fig~\ref{fig:model}), a multi-LLM system using a \textbf{M}ixture \textbf{o}f \textbf{D}ocument \textbf{S}peakers.
Inspired by panel discussions~\cite{doumont2014english}, \model has a \textit{Speaker} LLM for each document that responds to queries using its document, and a \textit{Moderator} LLM that decides when and how speakers respond.
Specifically, \model: 1) plans an agenda of topics for the outline (\cref{subsection:agenda}); 2) picks a subset of speakers with relevant perspectives for each topic and tailors them a query (\cref{subsection:moderator}); and 3) asks each speaker to obtain its document's context relevant to the tailored query and give the context's ``yes'' and ``no'' perspectives for the topic. 

When a speaker supplies its document's perspectives, the topic, document number, tailored query, and perspectives update an outline, tracking the LLM discourse.
After the discussion, the outline is summarized for a DQFS output.
In all, \model frames DQFS as a discussion of document speakers to represent sources equally, tailors queries for speakers to optimize the retrieval of contexts used to find perspectives, and builds a structured outline of document perspectives to simplify the synthesis of a final output---a novel combination that leads to comprehensive and balanced summaries~(\cref{subsection:ablation}).

We compare \model to eight strong baselines~on ConflictingQA~\cite{wan2024evidence} and \textbf{DebateQFS} (\cref{subsection:datasets}), a new dataset for DQFS drawn from the debate community on Debatepedia~\cite{gottopati2013learning}.
To assess summaries, we have models give citations in their outputs (Fig~\ref{fig:intro}), showing the documents the model intends to use~\cite{huang-chang-2024-citation}.
Many works use citations for factuality~\cite{li2024citation}, but
we repurpose them for coverage and balance---measuring the proportion of documents cited and distribution of ground-truth yes/no perspective stances of cited documents (\cref{subsection:metrics}).

\model has the best document coverage and balance in full summaries and topic paragraphs (\cref{subsection:citation_comp}), surpassing SOTA by 38-58\% in paragraphs.
The Prometheus LLM~\cite{kim2024prometheus} ranks \model as one of the best models in summarization quality 28/30 times, the most of any model (\cref{subsection:summary_comp}).
Users also find \model's outputs to be the most balanced, and preserve readability despite using perspectives from more documents (\cref{subsection:human_eval}).
Lastly, analyses show the utility of tailoring queries and building outlines, which improve \model (\cref{subsection:ablation}) and offer rich, structured tools for users (\cref{subsection:qg}). Our contributions are:

\noindent \textbf{1)} We propose \textbf{debatable query-focused summarization}, a new task to help users navigate yes/no queries in documents with opposing perspectives. \\
\noindent \textbf{2)} We design \model, a multi-LLM DQFS system that treats documents as \textbf{individual} \textbf{LLM speakers}, uses a moderator to \textbf{tailor queries} to apt speakers, and tracks speaker perspectives in an \textbf{outline}. \\
\noindent \textbf{3)} We release \textbf{DebateQFS} for DQFS and \textbf{citation metrics} to capture summary coverage and~balance. \\
\noindent \textbf{4)} Experiments show \model \textbf{beats baselines by 38-58\%} in topic paragraph coverage and balance, while annotators find \model's summaries \textbf{maintain readability} and \textbf{better balance perspectives}.

\section{Related Work}

\paragraph{Diverse Perspectives in Summarization:}

LLMs have shown to struggle with diverse input sources in news~\cite{huang-etal-2024-embrace}, review~\cite{zeng2023scientific}, and dialogue~\cite{zhang-etal-2024-fair} summarization.
While these tasks lack user guidance, DQFS is the first task that summarizes diverse texts while guided by a user's query.
Also, DQFS gives \textit{multi-aspect} summaries that are broken down into more specific paragraphs; this granularity of perspective diversity has not been studied in past work.


Most of these works expose LLM issues without giving solutions other than prompt tweaks~\cite{huang-etal-2024-embrace, zhang-etal-2024-fair}.
Instead, we design \model, a multi-LLM system to better handle diversity (\cref{subsection:citation_comp}), and also release a new dataset (\cref{subsection:datasets}) and citation metrics (\cref{subsection:metrics}) to help build even better summarization systems for diverse sources.

\paragraph{Argument Generation:}

DQFS is a form of argument generation~\cite{zukerman2000using}, producing text to argue for topics and claims~\cite{schiller2020aspect}.
Such tasks include debate~\cite{li2024can, hu2023americano, hu2024unlocking}, key point summarization~\cite{bar2020arguments, li2024exploring}, and argument essay writing~\cite{heinisch2022strategies, bao-etal-2022-aeg}.
These tasks either rely on LLM parametric memory~\cite{li2024can} or passages in evidence corpora~\cite{hua2019argument}. 
Conversely, in DQFS, models give arguments by summarizing and balancing perspectives in \textit{all} documents, rather than finding a \textit{subset} of evidence in large~corpora.

Further, existing datasets like OpenDebateEvidence~\cite{roush2024opendebateevidence} or DebateSum~\cite{roush-balaji-2020-debatesum} have specific claims (\textit{Colonialism made a hierarchy for exclusion}), which are unlike the broad queries in DQFS (\textit{Was colonialism helpful?}).
Thus, we release DebateQFS (\cref{subsection:datasets}), a dataset of broad debate queries grounded in documents.

\paragraph{Multi-LLM Summaries:}

Multi-LLM systems chain LLMs for tasks~\cite{guo2024large}.
\model is a multi-LLM system similar to single-turn debate~\cite{parrish2022single}, with a Moderator LLM routing to Speaker LLMs to supply document perspectives, storing them in memory (outline).
LLM discussions have been used for evaluation~\cite{verga2024replacing}, math~\cite{sun2023query}, and creativity~\cite{lu2024llm}, and we adopt them for DQFS.

The multi-LLM \model system has speakers respond individually to fairly treat documents.
Hierarchical Merging and Incremental Updating similarly summarize documents one at a time~\cite{chang2024booookscore}, but their intermediate outputs are free-form text.
\model instead uses a rich outline of document perspectives, better guiding the final summary~\cite{shao2024assisting}.
These models also summarize documents without catering to their expertise, hampering retriever efficacy; we solve this by tailoring custom queries for speakers (\cref{subsection:moderator}).

\section{Task Definition} \label{section:task}

Debatable query-focused summarization (DQFS) uses as input: 1) documents $\mathcal{D}$, where each source $d_i \in \mathcal{D}$ is a set of context paragraphs; 2) a yes/no query $q$; and 3) a number of summary topics $m>1$.
Source $d_i$ has perspectives $\mathcal{P}_i$, where perspective $(s, f) \in \mathcal{P}_i$ has stance $s \in \{\texttt{yes}, \texttt{no}\}$ and factual sentence $f$ derived via $d_{i}$, where $f$ supports $s$ as the answer to $q$.
We enforce ($\texttt{yes}, f$) and ($\texttt{no}, f$) are common in $\mathcal{P}$ (\cref{subsection:datasets}), meaning $q$ is \textbf{debatable}.

With these inputs, DQFS creates a summary $\mathbb{S}$ for $\mathcal{D}$ that answers $q$.
As seen in Figure~\ref{fig:intro}, $\mathbb{S}$ discusses topics $\mathcal{T} = \{t_1, ..., t_m\}$, each with a paragraph.
To aid trust and evaluation (\ref{subsection:metrics}), $\mathbb{S}$ contains citations (e.g. \texttt{[1]}) after each sentence noting the source document(s) for its information~\cite{huang-chang-2024-citation}.
For a \textit{comprehensive} and \textit{balanced} summary, we aim to cite a high number of documents in $\mathcal{D}$, ensuring no document's perspective is missed, and equally represent yes/no perspectives for $q$, curbing bias.
Comprehensiveness and balance are goals not only for the overall summary but also in each topic paragraph, ensuring a well-cited and balanced discussion within each topic.

\section{\model: Mixture of Document Speakers} \label{section:method}

For DQFS, we build \model (Figure~\ref{fig:model}), which uses content planning to guide generation~\cite{balepur-etal-2023-expository, shao-etal-2024-assisting} via the steps of drafting an outline $\mathcal{O}$; and condensing $\mathcal{O}$ into a summary $\mathbb{S}$.

To build $\mathcal{O}$, \model moderates a panel discussion of LLM speakers $\mathcal{S}$, where each speaker $s_i \in \mathcal{S}$~represents one document $d_i \in \mathcal{D}$.
\model executes: \textbf{1) Agenda Planning} to find $m$ topics $\mathcal{T}$ for $\mathcal{O}$ (\cref{subsection:agenda}); \textbf{2) Speaker Selection} to pick speakers $\mathcal{S}_j \in \mathcal{S}$ to respond to tailored queries for each topic $t_j \in \mathcal{T}$ (\cref{subsection:moderator}); and \textbf{3) Speaker Discussion} to prompt each speaker $s_i \in \mathcal{S}_j$ for its document's perspectives on $t_j$ and tailored query $q_{i,j}$ (\cref{subsection:speaker}), which are added to $\mathcal{O}$. We then prompt an LLM to use $\mathcal{O}$ to make a summary $\mathbb{S}$ (\cref{subsection:summary}). We describe each step below. 

\subsection{Agenda Planning} \label{subsection:agenda}

Before speakers discuss debatable query $q$ (``Is law school worth it?''), we must plan $m$ topics $\mathcal{T}$ (``law school jobs'') for the discussion (Fig~\ref{fig:model}, column 2).
In panel discussions, agendas are planned via \textit{biographies} summarizing speakers' expertise~\cite{pigeonholelive_panel_discussions}.
We also plan $\mathcal{T}$ with biographies $\mathcal{B}$ of our speakers' documents.
Instead of abstractively summarizing a speaker's document $d_i$ for its biography $b_i$ with an LLM, we efficiently create $b_i$ via extractive summarization---retrieving the $k$ contexts in $d_i$ most relevant to $q$ with ColBERT~\cite{khattab2020colbert}.
Then, in a 0-shot prompt, we ask an LLM to plan $m$ topics $\mathcal{T}$ relevant to $q$ and $\mathcal{B}$.

\subsection{Speaker Selection} \label{subsection:moderator}

After planning topics $\mathcal{T}$ for discussion (\cref{subsection:ablation}), we must decide which speakers $\mathcal{S}_j \subseteq \mathcal{S}$ are relevant for each topic $t_j \in \mathcal{T}$ (Fig~\ref{fig:model}, column 3).
We could pick all speakers, but this may hamper efficiency if we want to tailor queries for speakers (Appendix~\ref{subsection:efficiency}).
To illustrate, for the topic ``law school jobs,'' a document with perspectives on ``tuition costs'' can be omitted for efficiency, as it is not topically relevant.  

To solve this, a \textbf{Moderator} LLM picks relevant speakers $\mathcal{S}_{j}$ for each topic $t_j \in \mathcal{T}$. 
It is costly to prompt with all documents just to select speakers, so we use retrieval (\cref{subsection:agenda}) to create a biography $b_{i, j}$ of each speaker $s_i$ for topic $t_j$.
The biographies $\mathcal{B}_j$ are used in a 0-shot prompt, asking the moderator for speakers $\mathcal{S}_{j} \subseteq \mathcal{S}$ with biographies related to $t_j$.

To better cater to speakers' expertise, the Moderator also tailors a query $q_{i, j}$ specific to each selected speaker $s_i \in \mathcal{S}_j$ and topic $t_j$ using biography $b_{i,j}$; in panel discussions, moderators tailor queries to target speaker perspectives~\cite{Fingerhut2002, panel_discussion_questions}.
In \model, the queries form a chain-of-thought~\cite{10.5555/3600270.3602070}, improving our speaker selection (\cref{subsection:ablation}), and can be used for re-ranking~\cite{sachan2022improving}, serving as enhanced retrieval queries versus topic $t_j$ for speaker discussion (\cref{subsection:speaker}).
The tailored queries also further structure our outline $\mathcal{O}$, giving follow-up queries~\cite{liu2019fanda} for free that may interest users~(\cref{subsection:qg}).

\subsection{Speaker Discussion} \label{subsection:speaker}

After selecting relevant speakers $\mathcal{S}_j$ and tailoring them a query for each topic $t_j$ (\cref{subsection:moderator}), we must get the perspectives $\mathcal{P}$ from speakers' documents for the outline $\mathcal{O}$ (Fig~\ref{fig:model}, column 4).
A simple method to get $\mathcal{P}$ is to add all documents from $\mathcal{S}_j$ in one prompt and ask for perspectives on $t_j$, but LLMs often ignore text in the middle of long prompts~\cite{liu2024lost}, which may discard perspectives and reduce coverage. Further, LLMs may disregard the documents that oppose their parametric memory~\cite{jin2024tug}, skewing the outline's balance.

Using fairness ideals in panel discussions~\cite{Fingerhut2002}, speakers $s_i \in \mathcal{S}_j$ are \textit{individually} prompted to supply its document's perspectives for $t_j$ based on its tailored query $q_{i,j}$.
For example, on the topic ``law school jobs,'' we may query one speaker for ``market trends'' and another separately for ``Ivy League placement.''
Thus, each speaker adds its document's unique perspectives one at a time, preventing any one document from dominating, which leads to higher coverage (\cref{subsection:ablation}).


A speaker $s_i$ gives perspectives for a topic $t_j$ in two steps.
First, for efficiency, the speaker retrieves the $k$ contexts $\mathcal{C}$ in its document most relevant to the tailored query $q_{i, j}$.
Using the debatable query $q$, contexts $\mathcal{C}$, tailored query $q_{i, j}$, and topic $t_j$, the speaker is 0-shot prompted to give its yes and no perspectives $\mathcal{P}$ for $q$ based on $\mathcal{C}$, related to $q_{i, j}$ and $t_j$.
Each yes/no stance and fact $(s, f) \in \mathcal{P}$, tailored query $q_{i, j}$, and document number $i$ is added to $\mathcal{O}$ under topic $t_j$.
The yes/no stance predictions in $\mathcal{P}$ have $80$\% accuracy (Appendix~\ref{appendix:outline}), which better organizes $\mathcal{O}$ (\cref{subsection:qg}) to improve summaries~(\cref{subsection:ablation}).

\subsection{Outline Summarization} \label{subsection:summary}

Our outline $\mathcal{O}$ is a rich structure to track perspectives for a debatable query $q$, which we use as a content plan~\cite{balepur-etal-2023-expository} to create the final summary $\mathbb{S}$.
To do so, we test summarizing: 1) all of $\mathcal{O}$ in one prompt; and 2) topic sections of $\mathcal{O}$, i.e. $\{\mathcal{O}_j, \forall t_j \in \mathcal{T}\}$, one prompt at a time. 
We call these models 1) \modelAll and 2) \modelTopic. We detail the full \model system in Appendix~\ref{subsection:model}.


\section{Experimental Setup}

\subsection{Dataset Collection} \label{subsection:datasets}

DQFS needs entries of documents $\mathcal{D}$ with facts for ``yes'' and ``no'' answers to a query $q$. An apt dataset is \textbf{ConflictingQA}~\cite{wan2024evidence}, with controversial yes/no web search queries (``Do fires benefit forests?'') and labeled support/refute web~pages.

Other summarization diversity datasets are unsuited for DQFS.
Opinion summarization~\cite{zhang-etal-2024-fair} is grounded in subjective tweets/reviews, while DQFS needs fact-based texts.
DiverseSumm~\cite{huang-etal-2024-embrace} has diverse news articles, but lacks queries with opposing perspectives.
Debate datasets~\cite{roush2024opendebateevidence} are factual with opposing sides, but rely on argument mining corpora with specific claims (``Colonialism made an exclusion hierarchy''), which are hard to group into broad DQFS queries (``Is colonialism~good?'').


We create \textbf{DebateQFS}---a new dataset based on Debatepedia, the ``Wikipedia of debates''~\cite{gottopati2013learning}.
Debatepedia pages have broad topics (``carbon tax''), where users curate documents arguing pros/cons.
We turn topics into yes/no queries and collect the text of sites cited as pro/con sources.
We get 290 document sets for ConflictingQA and 183 for DebateQFS, each with a debatable query, with mean document set sizes of 10.47 and 9.86.
We also have ground-truth yes/no stances for the full documents, with mean majority/minority splits of 0.65/0.35 and 0.62/0.38.
We use these stances for summary balance (\cref{subsection:metrics}), but users also assess balance (\cref{subsection:human_eval}).
Appendix~\ref{appendix:data} has dataset details.

\begin{table*}[t]
\footnotesize
\centering
\setlength{\tabcolsep}{2.75pt}
\renewcommand{\arraystretch}{0.8}
\begin{tabular}{@{}clcccccccc@{}}
\multicolumn{1}{l}{} &  & \multicolumn{3}{c}{\textit{Summary Level}} & \multicolumn{3}{c}{\textit{Topic Paragraph Level}} & \multicolumn{2}{c}{\textit{Confounders}} \\ \toprule
\textbf{\# Top.} & \multicolumn{1}{l|}{\textbf{Model}} & \textbf{DC ($\uparrow$)} & \textbf{Fair ($\downarrow$)} & \multicolumn{1}{c|}{\textbf{\begin{tabular}[c]{@{}c@{}}Faithful ($\downarrow$)\end{tabular}}} & \textbf{DC ($\uparrow$)} & \textbf{Fair ($\downarrow$)} & \multicolumn{1}{c|}{\textbf{\begin{tabular}[c]{@{}c@{}}Faithful ($\downarrow$)\end{tabular}}} & \multicolumn{1}{l}{\textbf{Cite Acc. ($\uparrow$)}} & \textbf{All / Avg Sents} \\ \midrule
 & \multicolumn{1}{l|}{\modelTopic \textbf{(Ours)}} & \textbf{0.8961*} & {\ul 0.0998*} & \multicolumn{1}{c|}{\textbf{0.0320*}} & \textbf{0.6056*} & \textbf{0.1650*} & \multicolumn{1}{c|}{\textbf{0.0979*}} & 0.985 & 8.99 / 3.00 \\
\multirow{8}{*}{3} & \multicolumn{1}{l|}{\modelAll \textbf{(Ours)}} & {\ul 0.8664} & 0.1062* & \multicolumn{1}{c|}{0.0359*} & {\ul 0.5420} & {\ul 0.1896*} & \multicolumn{1}{c|}{{\ul 0.1217}} & 0.988 & 8.97 / 2.99 \\
 & \multicolumn{1}{l|}{Long-Context} & 0.5242 & 0.2047 & \multicolumn{1}{c|}{0.1733} & 0.2566 & 0.3816 & \multicolumn{1}{c|}{0.3503} & 0.958 & 9.00 / 3.00 \\
 & \multicolumn{1}{l|}{RAG-\textit{All}} & 0.6565 & 0.1664 & \multicolumn{1}{c|}{0.0911} & 0.3300 & 0.3296 & \multicolumn{1}{c|}{0.2547} & 0.990 & 9.01 / 3.00 \\
 & \multicolumn{1}{l|}{RAG-\textit{Doc}} & 0.7532 & 0.1364 & \multicolumn{1}{c|}{0.0668} & 0.3741 & 0.3023 & \multicolumn{1}{c|}{0.2352} & 0.949 & 9.01 / 3.00 \\
 & \multicolumn{1}{l|}{Hierarchical} & 0.8158 & \textbf{0.0956*} & \multicolumn{1}{c|}{{\ul 0.0333*}} & 0.3679 & 0.3136 & \multicolumn{1}{c|}{0.2523} & 0.981 & 8.99 / 3.00 \\
 & \multicolumn{1}{l|}{Incremental-\textit{All}} & 0.5037 & 0.2466 & \multicolumn{1}{c|}{0.1924} & 0.3467 & 0.3019 & \multicolumn{1}{c|}{0.2488} & 0.961 & 8.99 / 3.00 \\
 & \multicolumn{1}{l|}{Incremental-\textit{Topic}} & 0.5635 & 0.2288 & \multicolumn{1}{c|}{0.1720} & 0.4209 & 0.2796 & \multicolumn{1}{c|}{0.2236} & 0.963 & 9.01 / 3.00 \\
 & \multicolumn{1}{l|}{Cluster} & 0.7142 & 0.1203* & \multicolumn{1}{c|}{0.0662} & 0.3502 & 0.3016 & \multicolumn{1}{c|}{0.2517} & 0.927 & 9.04 / 3.01 \\
 & \multicolumn{1}{l|}{RAG+Cluster} & 0.7694 & 0.1332 & \multicolumn{1}{c|}{0.0620} & 0.3906 & 0.2808 & \multicolumn{1}{c|}{0.2101} & 0.976 & 9.02 / 3.01 \\ \midrule

 & \multicolumn{1}{l|}{\modelTopic \textbf{(Ours)}} & \textbf{0.9549*} & \textbf{0.0884*} & \multicolumn{1}{c|}{\textbf{0.0239*}} & \textbf{0.5924*} & \textbf{0.1661*} & \multicolumn{1}{c|}{\textbf{0.1051*}} & 0.986 & 15.00 / 3.00 \\
\multirow{8}{*}{5} & \multicolumn{1}{l|}{\modelAll \textbf{(Ours)}} & {\ul 0.9156} & {\ul 0.0966*} & \multicolumn{1}{c|}{{\ul 0.0272*}} & {\ul 0.4809} & {\ul 0.1972} & \multicolumn{1}{c|}{{\ul 0.1297}} & 0.990 & 14.88 / 2.98 \\
 & \multicolumn{1}{l|}{Long-Context} & 0.5779 & 0.2038 & \multicolumn{1}{c|}{0.1622} & 0.2164 & 0.4620 & \multicolumn{1}{c|}{0.4213} & 0.966 & 15.00 / 3.00 \\
 & \multicolumn{1}{l|}{RAG-\textit{All}} & 0.7331 & 0.1581 & \multicolumn{1}{c|}{0.0814} & 0.2755 & 0.3850 & \multicolumn{1}{c|}{0.3101} & 0.996 & 15.03 / 3.01 \\
 & \multicolumn{1}{l|}{RAG-\textit{Doc}} & 0.7898 & 0.1464 & \multicolumn{1}{c|}{0.0706} & 0.3018 & 0.3691 & \multicolumn{1}{c|}{0.2945} & 0.975 & 15.06 / 3.01 \\
 & \multicolumn{1}{l|}{Hierarchical} & 0.8871 & 0.0931* & \multicolumn{1}{c|}{0.0276*} & 0.2951 & 0.3670 & \multicolumn{1}{c|}{0.3038} & 0.987 & 15.01 / 3.00 \\
 & \multicolumn{1}{l|}{Incremental-\textit{All}} & 0.5392 & 0.2327 & \multicolumn{1}{c|}{0.1738} & 0.3083 & 0.3236 & \multicolumn{1}{c|}{0.2672} & 0.948 & 14.91 / 2.98 \\
 & \multicolumn{1}{l|}{Incremental-\textit{Topic}} & 0.6239 & 0.1899 & \multicolumn{1}{c|}{0.1337} & 0.3961 & 0.2902 & \multicolumn{1}{c|}{0.2348} & 0.958 & 14.99 / 3.00 \\
 & \multicolumn{1}{l|}{Cluster} & 0.8480 & 0.0968* & \multicolumn{1}{c|}{0.0464} & 0.3365 & 0.3093 & \multicolumn{1}{c|}{0.2625} & 0.933 & 15.04 / 3.01 \\
 & \multicolumn{1}{l|}{RAG+Cluster} & 0.8717 & 0.1084* & \multicolumn{1}{c|}{0.0436} & 0.3499 & 0.3136 & \multicolumn{1}{c|}{0.2511} & 0.971 & 15.03 / 3.01 \\ \bottomrule
\end{tabular}
\vspace{-1.5ex}
\caption{\label{table:doc_cover_cqa} ConflictingQA citation coverage, balance, and accuracy. Best model is \textbf{bold}, second best is \underline{underlined}. Models with * are significantly the best (2-sample $t$-test, $p<0.05$ with Bonferroni correction \cite{dror-etal-2018-hitchhikers}). }
\vspace{-1.25ex}
\end{table*}

\subsection{Baselines} \label{subsection:baselines}

We compare \model to SOTA LLM summarizers:\\
\noindent \textbf{1) Long-Context:} All documents $\mathcal{D}$ are used as the input in a single prompt~\cite{wang2024beyond}. \\
\noindent \textbf{2) RAG-\textit{All}:} Top-($k|\mathcal{D}|$) contexts in $\mathcal{D}$ relevant to $q$ are retrieved as input prompt~\cite{lewis2020retrieval}. \\
\noindent \textbf{3) RAG-\textit{Doc}}: Same as RAG-\textit{All}, but we retrieve the $k$-most relevant contexts in \textit{each} source in~$\mathcal{D}$. \\
\noindent \textbf{4) Hierarchical-\textit{All}:} Each document in $\mathcal{D}$ is summarized using $q$; these are summarized again into a final output under $m$ topics~\cite{chang2024booookscore}. \\
\noindent \textbf{5) Incremental-\textit{All}:} We plan topics $\mathcal{T}$ (\cref{subsection:agenda}) and iterate over each document in $\mathcal{D}$ to incrementally update the paragraphs for $\mathcal{T}$~\cite{chang2024booookscore}. For the final summary, we self-refine all paragraphs at once like chain-of-density~\cite{adams2023sparse}. \\
\noindent \textbf{6) Incremental-\textit{Topic}:} Same as Incremental-\textit{All}, but we self-refine topic paragraphs independently. \\
\noindent \textbf{7) Cluster:} We sort $\mathcal{D}$ into $m$ clusters, summarized to form topic paragraphs~\cite{hayashi-etal-2021-wikiasp}. \\
\noindent \textbf{8) RAG+Cluster:} Same as Cluster, but we retrieve the top-($k|\mathcal{D}|$) relevant contexts using $q$ before clustering, similar to LLooM~\cite{lam2024concept}.

These cover the main summarization paradigms: seq2seq~\cite{sutskever2014sequence}, clustering~\cite{zhang2009automatic}, content selection~\cite{louis2010discourse}, and multi-model frameworks~\cite{chang2024booookscore}.

\begin{table*}[t]
\centering
\footnotesize
\setlength{\tabcolsep}{2.75pt}
\renewcommand{\arraystretch}{0.8}
\begin{tabular}{@{}clcccccccc@{}}
\toprule
\textbf{\# Top.} & \multicolumn{1}{l|}{\textbf{Model}} & \textbf{DC ($\uparrow$)} & \textbf{Fair ($\downarrow$)} & \multicolumn{1}{c|}{\textbf{Faithful ($\downarrow$)}} & \textbf{DC ($\uparrow$)} & \textbf{Fair ($\downarrow$)} & \multicolumn{1}{c|}{\textbf{Faithful ($\downarrow$)}} & \multicolumn{1}{l}{\textbf{Cite Acc. ($\uparrow$)}} & \textbf{All / Avg Sents} \\ \midrule
\multirow{10}{*}{3} & \multicolumn{1}{l|}{\modelTopic \textbf{(Ours)}} & \textbf{0.8724*} & \textbf{0.0701*} & \multicolumn{1}{c|}{\textbf{0.0235*}} & \textbf{0.6066*} & \textbf{0.1255*} & \multicolumn{1}{c|}{\textbf{0.0789*}} & 0.982 & 8.99 / 3.00 \\
 & \multicolumn{1}{l|}{\modelAll \textbf{(Ours)}} & {\ul 0.8457*} & {\ul 0.0786*} & \multicolumn{1}{c|}{{\ul 0.0273*}} & {\ul 0.5508} & {\ul 0.1463*} & \multicolumn{1}{c|}{{\ul 0.0938*}} & 0.987 & 8.87 / 2.96 \\
 & \multicolumn{1}{l|}{Long-Context} & 0.5877 & 0.2094 & \multicolumn{1}{c|}{0.1790} & 0.2798 & 0.4336 & \multicolumn{1}{c|}{0.4028} & 0.953 & 9.02 / 3.01 \\
 & \multicolumn{1}{l|}{RAG-\textit{All}} & 0.6125 & 0.1544 & \multicolumn{1}{c|}{0.1040} & 0.3229 & 0.3176 & \multicolumn{1}{c|}{0.2701} & 0.997 & 9.01 / 3.00 \\
 & \multicolumn{1}{l|}{RAG-\textit{Doc}} & 0.7171 & 0.1180 & \multicolumn{1}{c|}{0.0664} & 0.3504 & 0.3233 & \multicolumn{1}{c|}{0.2748} & 0.961 & 9.01 / 3.00 \\
 & \multicolumn{1}{l|}{Hierarchical} & 0.7868 & 0.0907 & \multicolumn{1}{c|}{0.0374} & 0.3639 & 0.2980 & \multicolumn{1}{c|}{0.2452} & 0.983 & 9.02 / 3.01 \\
 & \multicolumn{1}{l|}{Incremental-\textit{All}} & 0.5566 & 0.2579 & \multicolumn{1}{c|}{0.2089} & 0.3919 & 0.3243 & \multicolumn{1}{c|}{0.2765} & 0.950 & 8.91 / 2.97 \\
 & \multicolumn{1}{l|}{Incremental-\textit{Topic}} & 0.6152 & 0.2415 & \multicolumn{1}{c|}{0.1970} & 0.4707 & 0.3128 & \multicolumn{1}{c|}{0.2674} & 0.954 & 9.03 / 3.01 \\
 & \multicolumn{1}{l|}{Cluster} & 0.7102 & 0.1106 & \multicolumn{1}{c|}{0.0725} & 0.3632 & 0.3106 & \multicolumn{1}{c|}{0.2737} & 0.931 & 9.04 / 3.01 \\
 & \multicolumn{1}{l|}{RAG+Cluster} & 0.6811 & 0.1405 & \multicolumn{1}{c|}{0.0894} & 0.3428 & 0.3200 & \multicolumn{1}{c|}{0.2689} & 0.977 & 9.01 / 3.00 \\ \midrule

\multirow{10}{*}{5} & \multicolumn{1}{l|}{\modelTopic \textbf{(Ours)}} & \textbf{0.9137*} & {\ul 0.0651*} & \multicolumn{1}{c|}{\textbf{0.0208*}} & \textbf{0.5793*} & \textbf{0.1420*} & \multicolumn{1}{c|}{\textbf{0.0998*}} & 0.986 & 14.99 / 3.00 \\
 & \multicolumn{1}{l|}{\modelAll \textbf{(Ours)}} & {\ul 0.8847*} & \textbf{0.0640*} & \multicolumn{1}{c|}{{\ul 0.0236*}} & {\ul 0.4991} & {\ul 0.1502*} & \multicolumn{1}{c|}{{\ul 0.1096*}} & 0.990 & 14.46 / 2.89 \\
 & \multicolumn{1}{l|}{Long-Context} & 0.6686 & 0.1724 & \multicolumn{1}{c|}{0.1392} & 0.2312 & 0.4965 & \multicolumn{1}{c|}{0.4640} & 0.966 & 15.01 / 3.00 \\
 & \multicolumn{1}{l|}{RAG-\textit{All}} & 0.6721 & 0.1423 & \multicolumn{1}{c|}{0.0912} & 0.2668 & 0.3927 & \multicolumn{1}{c|}{0.3438} & 0.996 & 15.02 / 3.00 \\
 & \multicolumn{1}{l|}{RAG-\textit{Doc}} & 0.7765 & 0.1053 & \multicolumn{1}{c|}{0.0618} & 0.3005 & 0.3584 & \multicolumn{1}{c|}{0.3147} & 0.975 & 15.01 / 3.00 \\
 & \multicolumn{1}{l|}{Hierarchical} & 0.8565 & 0.0761* & \multicolumn{1}{c|}{0.0239*} & 0.2896 & 0.3713 & \multicolumn{1}{c|}{0.3192} & 0.987 & 15.04 / 3.01 \\
 & \multicolumn{1}{l|}{Incremental-\textit{All}} & 0.6122 & 0.2000 & \multicolumn{1}{c|}{0.1629} & 0.3716 & 0.2936 & \multicolumn{1}{c|}{0.2572} & 0.948 & 14.77 / 2.95 \\
 & \multicolumn{1}{l|}{Incremental-\textit{Topic}} & 0.6767 & 0.1659 & \multicolumn{1}{c|}{0.1198} & 0.4446 & 0.2897 & \multicolumn{1}{c|}{0.2443} & 0.958 & 15.05 / 3.01 \\
 & \multicolumn{1}{l|}{Cluster} & 0.8098 & 0.1116 & \multicolumn{1}{c|}{0.0624} & 0.3292 & 0.3383 & \multicolumn{1}{c|}{0.2921} & 0.933 & 15.03 / 3.01 \\
 & \multicolumn{1}{l|}{RAG+Cluster} & 0.7811 & 0.1233 & \multicolumn{1}{c|}{0.0738} & 0.3129 & 0.3588 & \multicolumn{1}{c|}{0.3107} & 0.971 & 15.03 / 3.01 \\ \bottomrule
\end{tabular}
\vspace{-1.5ex}
\caption{\label{table:doc_cover_debate} DebateQFS citation coverage, balance, and accuracy. Best model is \textbf{bold}, second best is \underline{underlined}. Models with * are significantly the best (2-sample $t$-test, $p<0.05$ with Bonferroni correction \cite{dror-etal-2018-hitchhikers}). \model consistently has the highest citation coverage, fairness, and faithfulness for summaries and topic paragraphs.}
\vspace{-1ex}
\end{table*}
\begin{table*}[t]
\footnotesize
\definecolor{myblue}{HTML}{DAE8FC}
\centering
\setlength{\tabcolsep}{2.8pt}
\renewcommand{\arraystretch}{0.8}
\begin{tabular}{@{}clccccccccccccccc|c@{}}
\multicolumn{1}{l}{} &  & \multicolumn{5}{c}{\textit{Summary Quality}} & \multicolumn{5}{c}{\textit{Topic Paragraph Quality}} & \multicolumn{5}{c}{\textit{Topic Quality}} & \multicolumn{1}{c}{\textit{Dist.}} \\ \midrule
\textbf{Dataset} & \multicolumn{1}{l|}{\textbf{Model}} & \textbf{Int} & \textbf{Coh} & \textbf{Rel} & \textbf{Cov} & \multicolumn{1}{c|}{\textbf{Div}} & \textbf{Int} & \textbf{Coh} & \textbf{Rel} & \textbf{Cov} & \multicolumn{1}{c|}{\textbf{Div}} & \textbf{Int} & \textbf{Coh} & \textbf{Rel} & \textbf{Cov} & \textbf{Div}  & \textbf{SB} \\ \midrule
 & \multicolumn{1}{l|}{\modelTopic} & \cellcolor[HTML]{DAE8FC}4.24 & \cellcolor[HTML]{DAE8FC}4.34 & \cellcolor[HTML]{DAE8FC}4.64 & \cellcolor[HTML]{DAE8FC}4.49 & \multicolumn{1}{c|}{\cellcolor[HTML]{DAE8FC}\textbf{4.42}} & \cellcolor[HTML]{DAE8FC}\textbf{4.08} & \cellcolor[HTML]{DAE8FC}\textbf{4.33} & \cellcolor[HTML]{DAE8FC}\textbf{4.69} & \cellcolor[HTML]{DAE8FC}\textbf{4.34} & \multicolumn{1}{c|}{\cellcolor[HTML]{DAE8FC}\textbf{3.89}} & \cellcolor[HTML]{DAE8FC}3.47 & \cellcolor[HTML]{DAE8FC}\textbf{4.12} & \cellcolor[HTML]{DAE8FC}\textbf{4.69} & \cellcolor[HTML]{DAE8FC}\textbf{3.61} & \cellcolor[HTML]{DAE8FC}\textbf{4.02} & 0.69 \\
 & \multicolumn{1}{l|}{\modelAll} & \cellcolor[HTML]{DAE8FC}\textbf{4.27} & \cellcolor[HTML]{DAE8FC}4.33 & \cellcolor[HTML]{DAE8FC}4.63 & \cellcolor[HTML]{DAE8FC}4.49 & \multicolumn{1}{c|}{\cellcolor[HTML]{DAE8FC}4.40} & 3.88 & \cellcolor[HTML]{DAE8FC}4.27 & 4.60 & 4.19 & \multicolumn{1}{c|}{3.70} & \cellcolor[HTML]{DAE8FC}\textbf{3.49} & \cellcolor[HTML]{DAE8FC}4.09 & \cellcolor[HTML]{DAE8FC}4.62 & \cellcolor[HTML]{DAE8FC}3.46 & \cellcolor[HTML]{DAE8FC}3.99 & 0.65 \\
 & \multicolumn{1}{l|}{Hierarchical} & \cellcolor[HTML]{DAE8FC}4.24 & \cellcolor[HTML]{DAE8FC}\textbf{4.37} & \cellcolor[HTML]{DAE8FC}4.73 & \cellcolor[HTML]{DAE8FC}4.50 & \multicolumn{1}{c|}{\cellcolor[HTML]{DAE8FC}4.38} & 3.78 & 4.21 & 4.62 & 4.14 & \multicolumn{1}{c|}{3.57} & \cellcolor[HTML]{DAE8FC}3.43 & \cellcolor[HTML]{DAE8FC}4.07 & \cellcolor[HTML]{DAE8FC}4.65 & \cellcolor[HTML]{DAE8FC}3.49 & \cellcolor[HTML]{DAE8FC}3.94 & 0.58 \\
 & \multicolumn{1}{l|}{Increm-\textit{Topic}} & \cellcolor[HTML]{DAE8FC}4.17 & \cellcolor[HTML]{DAE8FC}4.37 & \cellcolor[HTML]{DAE8FC}\textbf{4.74} & \cellcolor[HTML]{DAE8FC}\textbf{4.57} & \multicolumn{1}{c|}{\cellcolor[HTML]{DAE8FC}4.39} & 3.91 & 4.29 & 4.62 & 4.25 & \multicolumn{1}{c|}{3.65} & 3.36 & 3.79 & 4.31 & 3.21 & 3.73 & 0.61 \\ \midrule
\multirow{-8}{*}{ConflictingQA} 
 & \multicolumn{1}{l|}{\modelTopic} & \cellcolor[HTML]{DAE8FC}4.02 & \cellcolor[HTML]{DAE8FC}4.20 & 4.49 & \cellcolor[HTML]{DAE8FC}\textbf{4.44} & \multicolumn{1}{c|}{\cellcolor[HTML]{DAE8FC}4.34} & \cellcolor[HTML]{DAE8FC}\textbf{3.97} & \cellcolor[HTML]{DAE8FC}\textbf{4.21} & \cellcolor[HTML]{DAE8FC}\textbf{4.55} & \cellcolor[HTML]{DAE8FC}\textbf{4.14} & \multicolumn{1}{c|}{\cellcolor[HTML]{DAE8FC}\textbf{3.82}} & \cellcolor[HTML]{DAE8FC}3.54 & \cellcolor[HTML]{DAE8FC}4.09 & \cellcolor[HTML]{DAE8FC}4.64 & 3.39 & \cellcolor[HTML]{DAE8FC}3.93 & 0.67 \\
 & \multicolumn{1}{l|}{\modelAll} & \cellcolor[HTML]{DAE8FC}4.11 & \cellcolor[HTML]{DAE8FC}4.21 & \cellcolor[HTML]{DAE8FC}4.60 & \cellcolor[HTML]{DAE8FC}4.34 & \multicolumn{1}{c|}{\cellcolor[HTML]{DAE8FC}\textbf{4.36}} & \cellcolor[HTML]{DAE8FC}3.83 & \cellcolor[HTML]{DAE8FC}4.15 & \cellcolor[HTML]{DAE8FC}4.51 & \cellcolor[HTML]{DAE8FC}4.10 & \multicolumn{1}{c|}{3.63} & \cellcolor[HTML]{DAE8FC}\textbf{3.61} & \cellcolor[HTML]{DAE8FC}4.11 & \cellcolor[HTML]{DAE8FC}4.67 & \cellcolor[HTML]{DAE8FC}\textbf{3.71} & \cellcolor[HTML]{DAE8FC}4.02 & 0.64  \\
 & \multicolumn{1}{l|}{Hierarchical} & \cellcolor[HTML]{DAE8FC}4.15 & \cellcolor[HTML]{DAE8FC}4.17 & \cellcolor[HTML]{DAE8FC}\textbf{4.69} & \cellcolor[HTML]{DAE8FC}4.35 & \multicolumn{1}{c|}{\cellcolor[HTML]{DAE8FC}4.33} & 3.74 & 4.09 & \cellcolor[HTML]{DAE8FC}4.53 & 3.96 & \multicolumn{1}{c|}{3.48} & \cellcolor[HTML]{DAE8FC}3.56 & \cellcolor[HTML]{DAE8FC}\textbf{4.22} & \cellcolor[HTML]{DAE8FC}\textbf{4.70} & \cellcolor[HTML]{DAE8FC}3.63 & \cellcolor[HTML]{DAE8FC}\textbf{4.16} & 0.58 \\
 & \multicolumn{1}{l|}{Increm-\textit{Topic}} & \cellcolor[HTML]{DAE8FC}\textbf{4.25} & \cellcolor[HTML]{DAE8FC}4.19 & \cellcolor[HTML]{DAE8FC}4.61 & \cellcolor[HTML]{DAE8FC}4.41 & \multicolumn{1}{c|}{\cellcolor[HTML]{DAE8FC}4.23} & \cellcolor[HTML]{DAE8FC}3.91 & \cellcolor[HTML]{DAE8FC}4.17 & \cellcolor[HTML]{DAE8FC}4.55 & \cellcolor[HTML]{DAE8FC}4.06 & \multicolumn{1}{c|}{\cellcolor[HTML]{DAE8FC}3.68} & 3.09 & 3.66 & 4.30 & 3.03 & 3.56 & 0.60 \\ \bottomrule
\multirow{-7}{*}{DebateQFS} 
\end{tabular}
\vspace{-3.5ex}
\caption{\label{table:llm_eval} Interest, coherence, relevance, coverage, and diversity for summaries, topic paragraphs, and topics ($m=3$). Best scores are \textbf{bold}. Significant scores are \colorbox{myblue}{blue} (2-sample $t$-test, $p<0.05$). Tables~\ref{appendix:table:llm_cqa} and \ref{appendix:table:llm_debate} have all results. \model is consistently ranked as having the significantly best quality for summaries, topic paragraphs, and topics.
}
\end{table*}

\subsection{Implementation Details} \label{subsection:implementation}

All models use 0-shot gpt-4-1106-preview~\cite{achiam2023gpt} with 0 temperature.~We write prompts using best practices on a small held-out set with fixed instructions for models~\cite{schulhoff2024prompt}.
LLMs are prompted to ``Use as many documents as possible'' and write three-sentence topic paragraphs.
The former ensures LLMs have the goal of coverage, while the latter fixes length confounders (\cref{subsection:citation_comp}).
Both of these strategies (specifying instructions, three-sentence text) have been used to improve summary balance~\cite{zhang-etal-2024-fair}.
We give mode details in Appendix~\ref{appendix:implementation}.

We retrieve via ColBERT~\cite{khattab2020colbert}, a retriever trained on MS-MARCO~\cite{Campos2016MSMA}, with $k=3$, and cluster using BERTopic and KMeans~\cite{macqueen1967some, grootendorst2022bertopic}.
Other parameters are default without tuning.
Results are from a single~run.

\subsection{Quantitative Evaluation via Citations} \label{subsection:metrics}

DQFS tests if models can cover and balance document perspectives.
To assess this, works use \textit{post-hoc} attribution, mapping summaries to sources they are believed to derive from~\cite{wolhandler2022multi, zhang-etal-2024-fair}.
But this does~not mean the model \textit{intends} to use all attributed texts.
A model may give perspectives using one source that is post-hoc attributable to many, gaming coverage and balance metrics without truly reflecting these qualities.

To solve this, we use \textit{pre-hoc} attributions~\cite{huang-chang-2024-citation}, i.e. citations, as they can better capture which documents the model intends to use.
Since each baseline gives document citations after each sentence (\cref{section:task}), and we know the ground-truth yes/no stances of these documents (\cref{subsection:datasets}), we can evaluate summary coverage and balance using the coverage and balance of the cited documents.

Let $\mathcal{D}_{cite} \subseteq \mathcal{D}$ be the cited documents in a text.
For \textit{comprehensiveness}, we let \textbf{document coverage (DC)} be the percent of sources in $\mathcal{D}$ cited.
For \textit{balance}, we use the ground-truth yes/no document stances.
We compute KL divergence of the distribution of $\mathcal{D}_{cite}$ stances to: 1) a uniform distribution; and 2) the stance distribution of all input documents $\mathcal{D}$.
(1) sees if $\mathcal{D}_{cite}$ splits perspectives equally, i.e. \textbf{fairness}~\cite{zhang-etal-2024-fair} and (2) tests if $\mathcal{D}_{cite}$ captures the input document split, i.e. \textbf{faithfulness}~\cite{fischer2022measuring}.
In DQFS, fairness is more critical for summary balance, but as our input documents have fairly balanced stance splits (\cref{subsection:datasets}), improving on both metrics is feasible.
We present citation faithfulness as another aspect of DQFS for research to explore.
These three metrics are aggregated over full summaries and topic paragraphs, as high-quality DQFS outputs should be balanced and comprehensive overall and within each paragraph.

\section{Results}

We generate DQFS summaries with two to five topics $m$, a traditional range for argumentative essays~\cite{mery2019use}.
Due to space constraints, we only show $m \in \{3, 5\}$ in the following sections, with all experiments repeated in Appendix~\ref{appendix:results}.

\subsection{Citation Coverage and Balance} \label{subsection:citation_comp}

\model excels at coverage and balance for summaries and topic paragraphs (Tables~\ref{table:doc_cover_cqa}, \ref{table:doc_cover_debate}).
Notably, \modelTopic leads in all metrics 22/24 times and is \textbf{always} a significantly best model.
\modelAll is also strong, a top-2 model in 22/24 cases.
Some models have high full summary scores, but \modelTopic largely improves DC/Fair/Faithful in \textit{topic paragraphs}, with 38/48/59\% mean increases over the next-best model. 
LLMs struggle in summarization coverage and diversity~\cite{huang-etal-2024-embrace, zhang-etal-2024-fair}, but we show that these issues are more pronounced in multi-aspect texts.
Our results verify \model's strategy of moderating \textit{single-turn} LLM discussions, so \textit{multi-turn} debates~\cite{khan2024debating} may produce even better summaries.

We also check two confounders: \textit{citation accuracy}, how often cited documents support claims in the sentence, and \textit{sentence count}.
Summaries can game our metrics via inaccurate citations or more sentences, as models cite post-sentence.
We assess citation accuracy via an LLM entailment model, a standard approach~\cite{gao2023enabling, balepur-etal-2023-text}, with 87\% human agreement on 200 held-out examples (Appendix~\ref{appendix:metrics}).
Citation accuracy and sentence count are consistent across models, so \model's strong coverage and balance are not due to extra sentences or non-existent perspectives.

Lastly, models can plan different topics, as we believe the open-aspect nature of topics in DQFS is interesting for future work~\cite{amar2023openasp}.
To ensure \model's gains are not from planning topics (\cref{subsection:agenda}) naturally more balanced or comprehensive, in Appendix~\ref{appendix:fixed_topics}, all models produce summaries for the same topics from our agenda planning step. 
\model is superior, validating its strength is from the LLM speaker design (\cref{section:method}), not topic selection.

\subsection{Summary Quality} \label{subsection:summary_comp}

Our citation metrics show \model excels in summary coverage and balance, so we now ensure that these gains are not at the cost of traditional measures of summary quality.
To do so, we conduct a sanity check and evaluate outputs
with five typical summary quality metrics~\cite{lloret2018challenging}: \textbf{int}erest, \textbf{coh}erence, \textbf{rel}evance, \textbf{cov}erage, \textbf{div}ersity.
The first 4 are from~\citet{shao2024assisting}, who use them on Wikipedia writing, while diversity is new for DQFS, testing the balance of yes and no perspectives.
We use Prometheus, an LLM with 72-85\% human agreement~\cite{kim2024prometheus}, for 1-5 scoring (Appendix~\ref{appendix:metrics}).
We score summaries, topic paragraphs, and topics using these metrics.

Prometheus just uses the summary, topic paragraph, or topic title as input and does not have access to the input documents.
Thus, our evaluation of coverage through citation metrics (\cref{subsection:metrics}) measures the coverage of the input documents, while Prometheus assesses coverage using its parametric knowledge, specifically evaluating if the outputs provide ``an in-depth exploration of the query and have good coverage.''
LLM evaluators can be biased~\cite{wang-etal-2024-large-language-models-fair}, so we also conduct a human evaluation in \cref{subsection:human_eval}.
We also use \textbf{S}elf-\textbf{B}leu~\cite{zhu2018texygen} ($n=4$) to assess the semantic distance between paragraphs~\cite{liu-etal-2023-dimongen}. 

\modelTopic and \modelAll have significantly high-quality summaries, topic paragraphs, and topics 28/30 and 25/30 times (Table~\ref{table:llm_eval}).
In summaries and paragraphs, \model has the best coverage 5/6 times and diversity 6/6 times, aligning with our citation metrics (\cref{subsection:citation_comp}).
\model has a slightly higher SB, meaning more paragraph similarity.
This similarity does not largely impede readability (\cref{subsection:human_eval}), and this occurs as \model adapts similar perspectives for distinct topics\footnote{For example, a document's perspective that ``electric cars must be recharged often'' relates to ``consumer utility'' and ``energy use''--- distinct topics. Appendix~\ref{appendix:outputs} has examples.}.
Given the tradeoff in paragraph coverage and dissimilarity~\cite{alguliev2012gendocsum+}, a small SB increase is worth the large coverage and balance gains (\cref{subsection:citation_comp}).
Overall, \model exhibits strong coverage, balance, and quality.

\begin{table}[]
\small
\centering
\setlength{\tabcolsep}{3pt}
\renewcommand{\arraystretch}{0.8}
\begin{tabular}{@{}lcccc@{}}
 & \multicolumn{2}{c}{\textit{ConflictingQA}} & \multicolumn{2}{c}{\textit{DebateQFS}} \\ \toprule
\textbf{Model} & \multicolumn{1}{l}{\textbf{Read (S/P) }} & \multicolumn{1}{l|}{\textbf{Bal (S/P) }} & \multicolumn{1}{l}{\textbf{Read (S/P) }} & \multicolumn{1}{l}{\textbf{Bal (S/P) }} \\ \midrule
\model & 4.45/4.39 & \multicolumn{1}{c|}{\textbf{4.04/3.93}} & 4.43/4.21 & \textbf{4.03/3.84} \\
Long-Cont. & 4.40/\textbf{4.45} & \multicolumn{1}{c|}{3.54/2.65} & 4.44/\textbf{4.44} & 3.55/2.78 \\
Hierarch. & \textbf{4.63}/4.43 & \multicolumn{1}{c|}{3.94/3.85} & 4.46/4.23 & 3.70/3.10 \\
Inc-\textit{Topic} & 4.29/4.21 & \multicolumn{1}{c|}{3.94/3.13} & \textbf{4.53}/4.28 & 3.51/3.01 \\ \bottomrule
\end{tabular}
\vspace{-1.5ex}
\caption{\label{table:human} Readability/balance of summaries/paragraphs. \model has the most balanced text and despite including more perspectives, \model has competitive readability. }
\vspace{-1.5ex}
\end{table}

\subsection{Human Evaluation} \label{subsection:human_eval} 

We have 76 users compare 20 DQFS outputs per dataset from \modelTopic to Hierarchical and Incremental-\textit{Topic}, the next-best models, and long-context, the simplest model.
\model cites more documents (\cref{subsection:citation_comp}), so users rate \textit{\textbf{Read}ability}~\cite{ribeiro2023generating} to ensure the extra perspectives do not harm comprehension.
Users also rate \textit{\textbf{Bal}ance}, as DQFS must fairly show yes/no stances. Scores are from 1-5 (Appendix~\ref{appendix:human}) and are used for full summaries and paragraphs on the same topic.

\model has similar readability to baselines (Table~\ref{table:human}), meaning our additionally cited perspectives are clearly conveyed, and users find \model's summaries/paragraphs the most balanced.
In 3/4 cases, \model has the highest average of readability and balance.
Thus, \model is the best DQFS model,~citing more documents and better balancing perspectives versus SOTA, all while preserving readability.

\begin{table}[!t]
\centering
\small
\setlength{\tabcolsep}{3.9pt}
\renewcommand{\arraystretch}{0.8}
\begin{tabular}{@{}ccccccc@{}}
\multicolumn{1}{l}{} & \multicolumn{3}{c}{\textit{ConflitingQA}} & \multicolumn{3}{c}{\textit{DebateQFS}} \\ \toprule
\multicolumn{1}{l|}{\textbf{Model}} & DC & |$\mathcal{P}$| & \multicolumn{1}{l|}{|$\mathcal{P}$| / Doc} & DC & |$\mathcal{P}$| & 
 |$\mathcal{P}$| / Doc \\ \midrule
\multicolumn{1}{l|}{\model} & 79.1 & \textbf{30.3} & \multicolumn{1}{c|}{\textbf{3.61}} & 77.8 & \textbf{25.5} & \textbf{3.36} \\
\multicolumn{1}{l|}{No Tailor} & 77.4 & 27.5 & \multicolumn{1}{c|}{3.37} & 72.6 & 22.2 & 3.07 \\
\multicolumn{1}{l|}{No CoT} & 78.8 & 27.8 & \multicolumn{1}{c|}{3.35} & 73.3 & 22.7 & 3.07 \\ 
\multicolumn{1}{l|}{No Speak} & 47.4 & 13.3 & \multicolumn{1}{c|}{3.28} & 45.6 & 11.6 & 3.30 \\
\multicolumn{1}{l|}{No Mod} & \textbf{84.8} & 29.8 & \multicolumn{1}{c|}{3.17} & \textbf{ 79.8} & 24.0 & 3.06 \\ 
\bottomrule
\end{tabular}
\vspace{-1.5ex}
\caption{\label{table:ablation_outline} \model ablation outline metrics ($m=3$).~We show the document coverage, the total perspectives, and perspectives per document under each outline topic.}
\vspace{-1ex}
\end{table}
\begin{table}[!t]
\centering
\small
\setlength{\tabcolsep}{4pt}
\renewcommand{\arraystretch}{0.8}
\begin{tabular}{@{}l|cccc@{}}
\multicolumn{1}{l}{} & \multicolumn{2}{c}{\textit{ConflitingQA}} & \multicolumn{2}{c}{\textit{DebateQFS}} \\
\toprule
\textbf{Model} & \textbf{DC (S/P)} & \multicolumn{1}{c|}{\textbf{Fair (S/P)}} & \textbf{DC (S/P)} & \textbf{Fair (S/P)} \\ \midrule

No Mod & \textbf{0.96/0.75} & \multicolumn{1}{c|}{\textbf{0.02/0.07}} & \textbf{0.89/0.65} & \textbf{0.03/0.10} \\
- $\mathcal{O}$ &  0.88/0.60 & \multicolumn{1}{c|}{0.11/0.17} & 0.78/0.50 & 0.10/0.22  \\ \midrule

\model & \textbf{0.90}/0.61 & \multicolumn{1}{c|}{\textbf{0.03/0.10}} & \textbf{0.87/0.61} & \textbf{0.02/0.08} \\
- Stance & 0.88/0.58 & \multicolumn{1}{c|}{0.08/0.12} & 0.85/0.58 & 0.08/0.15 \\ 
- Tailor & 0.89/\textbf{0.62} & \multicolumn{1}{c|}{0.10/0.17} & \textbf{0.87}\textbf{/}\textbf{0.61} & 0.08/0.13 \\
\bottomrule

\end{tabular}
\vspace{-1.5ex}
\caption{\label{table:ablation_summary} \model summary ablations ($m=3$). Using an outline and richer outline structures both improve coverage and fairness for summaries and topic paragraphs.}
\vspace{-3ex}
\end{table}
\definecolor{myblue1}{HTML}{a5c3f0}
\definecolor{myred1}{HTML}{e99999}
\definecolor{myyellow1}{HTML}{f8d862}
\begin{figure}[!t]
    \centering
    \includegraphics[width=\linewidth]{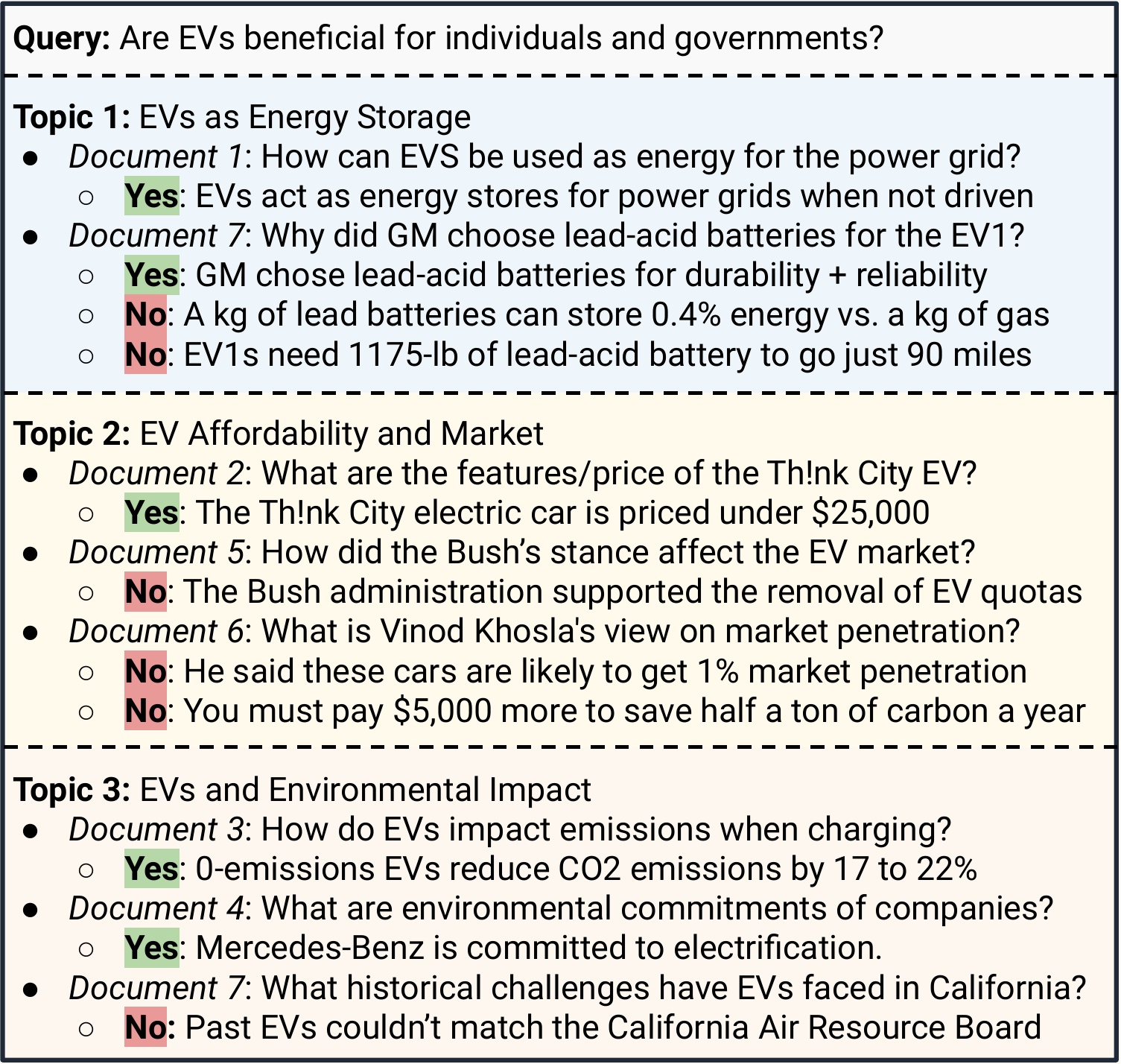}
    \vspace{-4ex}
    \caption{\label{fig:outline} Example outline subset from \model, which clearly tracks topics, documents, perspectives (facts and stances), and follow-up queries for the user to explore. }
\end{figure}

\subsection{Ablation Study} \label{subsection:ablation}

We ensure all parts of \model are useful by ablating our outline creation and summarization steps.
In outline creation, having individual speakers respond versus combining all speaker biographies in a prompt (No Speak), tailoring queries (No Tailor), and picking speakers via CoT (No CoT) all improve outlines (Table~\ref{table:ablation_outline}).
No Speak has the worst outlines, confirming the strength of equally treating document speakers for DQFS.
We also test our moderator's abilities by having all speakers respond (No Mod) instead of selecting speakers.
No Mod has higher DC as all speakers respond, but fewer perspectives per document, meaning our moderator adeptly selects speakers with relevant perspectives.

To see how outline $\mathcal{O}$ alters summaries, we compare \model (with no moderator) updating an outline to updating free-form paragraphs (-$\mathcal{O}$). 
Using $\mathcal{O}$ greatly improves coverage and fairness (Table~\ref{table:ablation_summary}, top), showing structured outlines are better intermediate outputs than free-form text in multi-LLM systems.
Further, extra organization in $\mathcal{O}$ (stances, tailored queries) aids summarization (Table~\ref{table:ablation_summary}, bottom), so richer outlines yield better summaries.




\subsection{Outline Analysis Case Study} \label{subsection:qg}

\model builds an outline $\mathcal{O}$ as a content plan pre-summarization~\cite{shao2024assisting}, but $\mathcal{O}$ is also a valuable tool for users~\cite{barrow2021syntopical}.
Figure~\ref{fig:outline} shows part of an outline, which organizes perspectives with their source documents and yes/no stances.
$\mathcal{O}$ outlines all seven input documents and a range of perspectives for a thorough, balanced view of the debatable query.
Further, the tailored document queries in $\mathcal{O}$ can inspire users to explore follow-up queries to ask.
Overall, $\mathcal{O}$ gives a rich, structured representation of perspectives, enabling in-depth explorations of document collections.

\section{Conclusion}

We propose DQFS and design \model, which controls individual document speakers, to write well-covered, balanced summaries.
\model has potential utility past DQFS, such as in code-switching~\cite{gao2019code}, multi-modal generation~\cite{dai2022enabling}, and full-stack design~\cite{si2024design2code}, where balancing diverse inputs (languages, modalities) is crucial.
We also show that content planning with outlines largely enhances DQFS quality.
Future work can explore the direct application of our outline to tasks with opposing stances, like pro/con generation~\cite{kumar2023apcs}, document contradiction detection~\cite{deusser2023contradiction}, or key point analysis~\cite{kunneman2018aspect}.
While \model excels in DQFS, romising extensions to our task would still challenge our model, such as document misinformation~\cite{sung2023not} or aligning summaries with and against expressed or observed \textit{user} perspectives~\cite{balepur2024smart}.
These insights, along with our new DebateQFS dataset and citation metrics, will be key toward~building models that can handle diverse, opposing perspectives.



\section{Limitations}

One limitation of \model is cost, as it uses multiple LLM calls. To reduce costs, we use top-3 retrieval at each step and a moderator to avoid inference on irrelevant documents (\cref{subsection:moderator}).
Appendix~\ref{subsection:efficiency} has a cost analysis, which shows \model is cheaper than Incremental-\emph{Topic} and is comparable to Hierarchical Merging for fewer topics.
Most of the cost from \model stems from outline creation, rather than outline summarization.
Our outline is a rich resource for users (\cref{subsection:qg}) and can also be useful for other tasks like key point analysis~\cite{bar2020arguments, kumar2023apcs}, pro/con summarization~\cite{hu2009classification}, and document contradiction detection~\cite{deusser2023contradiction}, which we believe justifies its high creation expense.

Further, all baseline implementations are based on GPT-4, as smaller LLMs like LLaMA-2 and GPT-3.5 struggled with following the instructions given in our 0-shot prompts (Appendix~\ref{appendix:implementation}), particularly in generating structured JSON outputs~\cite{xia2024fofo}. To overcome this, researchers could generate synthetic training data to improve format-following in smaller models~\cite{long2024llms}.
We also show some preliminary results with a version of \model using GPT-4 mini in Appendix~\ref{appendix:results_mini}, which can still compete with GPT-4 baselines. 

LLMs are also sensitive to prompt formats~\cite{sclar2023quantifying}, so our results may vary with prompt changes.
To mitigate this issue, we follow best practices in prompt engineering~\cite{schulhoff2024prompt}, ensuring consistent instructions across models, including input/output definitions, output format (JSON), and output requirements.
This ensures \model's gains in coverage and balance (\cref{subsection:citation_comp}) are due to its overall design, rather than advantages in prompt engineering.
We also plan to release all of our prompts for reproducibility (Appendix~\ref{appendix:implementation}).

Finally, while human evaluation across many aspects of DQFS quality would be valuable, we are limited by time and resources. To make the most of our human evaluation, we focus on readability and balance. Since MoDS objectively cites more documents and is thus more comprehensive, we ensure that this does not reduce readability. Further, since DQFS aims to support unbiased decision-making, we assess whether human judgments of summary balance align with our offline citation metrics. We acknowledge that further human evaluation, including how DQFS outputs impact decision-makers, would be an exciting direction for future research.

\section{Ethical Considerations} \label{subsection:ethics}

The goal of debatable query-focused summarization is to provide comprehensive and balanced summaries for yes/no queries that fairly represent both ``yes'' and ``no'' perspectives.
However, we acknowledge that not all yes/no queries should be balanced in a summary.
Balancing some queries could spread misinformation (e.g. ``Is the earth flat?''), or the user might prefer to focus on one side of the issue.
For misinformation, we limit DQFS to queries with \textit{equally-valid} opposing perspectives, as reflected in our high-quality DebateQFS dataset, which is annotated by the debate community. 
For user preferences, future work could study using the \textit{user's} perspective as input, tasking models to generate summaries that align with or challenge the user's viewpoint. This would enable fine-grained control, allowing users to decide when to balance diverse perspectives or focus on a preferred one.

Further, we assume our input documents are factual and written in good faith for DQFS, but this is not always guaranteed in practice.
To detect document misinformation, future DQFS research could explore adversarial settings where input documents contain factual errors, requiring models to incorporate a fact verification module to filter out factual inaccuracies.
In \model, a fact verification system could be run on the facts in \model's outline before summarization to discard factual inaccuracies.
We believe such efforts are essential for developing safe, factual, and reliable summarization systems.

\section*{Acknowledgments}

We would like to thank our collaborators at Adobe Research and the University of Maryland for their valuable contributions and insights that helped shape this work, including Jack Wang, Rajiv Jain, Jennifer Healey, Vishy Swaminathan, and Rachel Rudinger.
Nishant is also grateful to the amazing cohort of interns at Adobe Research---including Dang Nguyen, Vishakh Padmakumar, Dayeon Ki, Hyunji Lee, Yoonjoo Lee, and Paiheng Xu---not just for their helpful feedback, but also for making the internship a fun and memorable experience.

\bibliography{custom}
\bibliographystyle{acl_natbib}

\clearpage

\appendix

\section{Appendix} \label{appendix}

\subsection{Dataset Details} \label{appendix:data}

To collect a dataset based on Debatepedia~\cite{gottopati2013learning}, we use Wayback Machine\footnote{\url{https://web.archive.org/}}, as the original website is no longer hosted.
We iterate through each debate articles page on the website with BeautifulSoup\footnote{\url{https://pypi.org/project/beautifulsoup4/}} and collect the 1) topic of the debate; 2) list of URLs under ``Supporting References''; and 3) list of URLs under ``Refuting References''.
We then use jusText\footnote{\url{https://pypi.org/project/jusText/}} to extract the text content from each web page, ignoring websites that are not free-to-access.

After this, we filter out instances that have less than five sources or do not have at least a 75/25 majority/minority split of perspective labels.
We then remove web pages that do not have any of the non-stopword tokens in the query, implemented with nltk, to ensure the web pages form a set of relevant documents.
We run this same process on ConflictingQA~\cite{wan2024evidence}.

Dataset statistics after data processing are in Table~\ref{appendix:table:dataset}.
Since all websites were publicly-accessible, our collected artifacts are within their intended use and licenses.
We sampled a subset of five document collections and manually checked them for PII and offensive content, which we did not find; we also found all text to be in English.

\subsection{The \model Algorithm} \label{subsection:model}

We detail \model in Algorithm~\ref{algo:mods}. For a debatable query $q$, document collection $\mathcal{D}$, number of topics $m$, and retrieval parameter $k$, we create speakers $\mathcal{S}$ for $\mathcal{D}$.
First, we retrieve speaker biographies $\mathcal{B}$ related to $q$ and plan $m$ topics $\mathcal{T}$ for $\mathcal{O}$ (\cref{subsection:agenda}). For each topic $t_j \in \mathcal{T}$, we pick relevant speakers $\mathcal{S}_j \subseteq \mathcal{S}$ and tailor them questions $\mathcal{Q}_j$ using their topic biographies $\mathcal{B}_j$ (\cref{subsection:moderator}). Each speaker supplies stance/fact perspectives $\mathcal{P}$, which are tracked in $\mathcal{O}$ (\cref{subsection:speaker}).
Finally, $\mathcal{O}$ is summarized all at once ($\mathbb{S}_{all}$) or per topic ($\mathbb{S}_{top}$) and returned to the user (\cref{subsection:summary}).

\subsection{Experimental Setup Details} \label{appendix:implementation}

All of our baseline implementations use GPT-4 (\texttt{gpt-4-1106-preview}) with 0 temperature and a maximum input token length of 127,000 tokens.
All baselines use zero-shot prompting, and the prompts will be released with our code after internal approval.
For costs associated with using GPT-4, see Appendix~\ref{subsection:efficiency}.

All models using retrieval, including \model, use ColBERT~\cite{khattab2020colbert}, a state-of-the-art retriever. For hyperparameters, we use a maximum document length of 300 tokens, a maximum query length of 64 tokens, 8 bits, and the \texttt{colbert-ir/colbertv2.0} checkpoint; none of these parameters were tuned during experimentation.
The clustering methods were implemented with BERTopic~\cite{grootendorst2022bertopic}, using all default values.

All experiments were run on a single H100 GPU, but as the only GPU usage comes from retrieval, we found \model and all baselines can be run on a Google Collaboratory T4 GPU (16GB of GPU memory).
Each baseline was allocated 24 hours for a single run.
We give more details about the runtime of \model in Appendix~\ref{subsection:efficiency}.


\subsection{Metric Details} \label{appendix:metrics}

We extract all citations via regex\footnote{\url{https://docs.python.org/3/library/re.html}} by first finding text between square brackets (\texttt{[} and \texttt{]}) and then extracting integers between these spans.
The document coverage, faithfulness, and fairness metrics are all implemented with numpy\footnote{\url{https://numpy.org/}}.

We implement citation accuracy through entailment; entailment has shown to be a viable strategy to measure the factuality of text~\cite{maynez2020faithfulness}.
We use GPT-3.5 (\texttt{gpt-35-turbo-1106}) with 0 temperature to classify whether a generated sentence is entailed by the document it cited, using a 0-shot prompt shown in Prompt~\ref{prompt:cite_acc}
To evaluate the accuracy of this metric, we manually annotate 200 held-out examples (100 examples GPT predicted to be accurate citations, and 100 examples predicted to be inaccurate citations) of generated summaries for DQFS from all models (not used in evaluation).
We annotate these blindly, without knowing the output classification of GPT-3.5.
On this set, we obtain 87\% agreement with GPT-3.5, close to the agreement of 88\%, 90\%, and 96\% shown by human annotators in~\citet{min2023factscore}.
Further, this value is near the entailment-based accuracy given in other factuality tasks~\cite{balepur-etal-2023-expository, balepur-etal-2023-text}.

For the summary quality evaluation (\cref{subsection:summary_comp}), we use the Prometheus-v2 LLM evaluator\footnote{\url{https://github.com/prometheus-eval}}.
Example rubrics given to this evaluator are in Table~\ref{table:rubric}, which are adapted directly from~\citet{shao2024assisting}.

\subsection{Efficiency and Cost Comparison} \label{subsection:efficiency}

In Tables~\ref{appendix:table:cost_cqa} and \ref{appendix:table:cost_debate}, we present the cost (LLM input/output tokens, number of calls) and efficiency (seconds taken for inference) of \modelTopic, the slightly more expensive model out of the two \model baselines, versus Hierarchical Merging and Incremental Updating~\cite{chang2024booookscore, adams2023sparse}, the two other best-performing baselines, which also happen to be multi-LLM systems. Despite \model using more LLM calls through single-turn LLM debate, our use of retrieval and a moderator LLM greatly reduces the number of input tokens \model otherwise would have consumed, keeping GPT-4 cost competitive with Hierarchical Merging, and making our model cheaper than Incremental-\textit{Topic}.
The inference time of multi-LLM summarization systems like \model could be improved, a common limitation of agentic systems~\cite{li2024personal}, and one possible strategy could be to use multi-threading or batched decoding to parallelize the discussions of LLM speakers. 

\subsection{Results for All Topics} \label{appendix:results}

We run \model and all baselines where the number of topics $m$ ranges between $2$ and $5$ inclusive, a typical range of paragraphs in argumentative essays~\cite{mery2019use}.
Tables~\ref{table:doc_cover_cqa_all} and \ref{table:doc_cover_debate_all} display the citation coverage and balance metrics from \cref{subsection:citation_comp} for all $m$, while Tables~\ref{appendix:table:llm_cqa} and \ref{appendix:table:llm_debate} display the summary quality metrics from \cref{subsection:summary_comp} results for all $m$.
Our claims hold for these varied values of $m$; \model generates comprehensive and balanced summaries while maintaining traditional output quality metrics, regardless of the number of topic paragraphs it must generate.

\subsection{Results for Hierarchical Merging over Topic Paragraphs} \label{appendix:results_hm}

Further, the Hierarchical Merging baseline we use does not generate summaries one topic at a time.
We believe that such a model (i.e. Hierarchical-\emph{Topic}) is too costly and inefficient to deploy, so we do not compare against it in the main body of the work. 
In Tables~\ref{table:doc_cover_cqa_all_comp} and \ref{table:doc_cover_debate_all_comp} we provide some results for this model, which still underperforms \modelTopic.
Further, we show in Table~\ref{appendix:table:cost_weird} that this model is much more costly compared to \model.
It is also more costly than a version of \model that iterates through all speakers, highlighting the utility of retrieval to keep inference time and LLM cost low.

\subsection{Results with GPT-4 Mini} \label{appendix:results_mini}

All of our models are implemented with GPT-4, but we also run some preliminary experiments with \modelTopic using GPT-4 mini.
In citation coverage, fairness, and faithfulness (Tables~\ref{table:doc_cover_cqa_all_comp_mini} and \ref{table:doc_cover_debate_all_comp_mini}), \modelTopic using GPT-4 mini underperforms the model using GPT-4, suggesting that larger models are better suited for multi-LLM systems like \model. However, the GPT-4 mini system still exhibits strong performance, and is even comparable to several of the baselines using GPT-4 in Tables~\ref{table:doc_cover_cqa_all_comp} and \ref{table:doc_cover_debate_all_comp}, further showcasing the efficacy of our framework.

\subsection{Results with Fixed Topics} \label{appendix:fixed_topics}

Each baseline in \cref{subsection:baselines} produces distinct topics while planning a summary.
To ensure the citation coverage and balance gains in \model are not just derived from our agenda planning step (\cref{subsection:agenda}), we implement a version of each baseline that is asked to generate summaries for the same topics that \modelTopic generates.
We present these results in Tables~\ref{table:doc_cover_cqa_fixedtopic} and \ref{table:doc_cover_debate_fixedtopic}, and find that \model still largely outperforms baselines even when using our topics, suggesting that our agenda planning is not the source of gains in the framework.


\subsection{Outline Perspective Accuracy} \label{appendix:outline}

During speaker discussion (\cref{subsection:speaker}), we ask speakers to provide perspectives in the form of facts in the document.
These facts are grouped by whether the fact gives evidence for why the answer to the query is ``yes'' or ``no'', which also provides another layer of organization to enrich the user's understanding of the outline (\cref{subsection:qg}).
To assess the accuracy of these yes/no labels, we ask human annotators to label if each paragraph in 10 document collections (5 from DebateQFS, 5 from ConflictingQA) strongly supports, weakly supports, strongly refutes, weakly refutes, or is neutral toward the input query.
In total, we collect 7592 annotations, and aggregate them into one of three labels: supports, refutes, or neutral.\footnote{For each annotator, we score a paragraph as $ \pm1$ for strongly support/reject, $\pm0.5$ for weakly support/reject, and 0 for neutral. We take the sum of these scores over all annotators, and set the final label to support/reject if the sum is greater/less than 0. A score of 0 yields a neutral label.}
We will also release these paragraph-level annotations, which may be useful for training DQFS models.
We use the same procedure in Appendix~\ref{appendix:human} for this user study.

After collecting ground truth paragraph labels, we take the outlines produced by \model on this subset of 10 examples. For each predicted yes/no fact in the outline, we post-hoc attribute~\cite{huang-chang-2024-citation} the paragraph in the speaker's document that was the source of the information in the fact (with ColBERT).
We compare the accuracy of the LLM's yes/no label using the ground truth labels from human annotators, which are 0.798, 0.806, 0.781, and 0.803 for $m = 2, 3, 4, 5$, respectively.
Our accuracy is near the accuracy of LLMs on existing stance detection benchmarks~\cite{lan2024stance}, meaning our yes/no labels provide a useful and fairly accurate signal for users.

\subsection{Human Evaluation Setup} \label{appendix:human}
We conducted user evaluations to compare the readability and balance of summaries produced by different models (\model, Long-Context, Hierarchical, Incremental-Topic). The evaluation was divided into two parts: one focusing on the entire summary and the other on topic paragraphs.

\subsubsection{Recruitment \& Procedure}
We recruited 76 participants via Prolific, all of whom were based in the United States and required to have fluency in English. Each participant rated a total of 20 summaries, assessing the output from each of the four models for a given debate query. 
Participants were paid \$12/hour, the recommended rate on the website.
To mitigate order and fatigue effects, the presentation order of summaries was counterbalanced. Each summary was rated by 3-5 different participants. Additionally, the task included two baseline comprehension checks to ensure participants understood the instructions and metric definitions. Participants who did not pass these checks were excluded from the final analysis.
These annotations did not require review from an Institutional Review Board (IRB).
We collect no Personal Identifiable Information during the study.

\subsubsection{Rating Criteria}
The task included two Likert ratings for Readability and Balance. Additionally, participants could provide open comments for feedback or to report any issues. For the Likert items, participants saw the following questions:

\begin{itemize}
    \item \textbf{Readability.} Is the summary easy to read and understand?
    \begin{enumerate}
        \item The summary is very unclear, with consistent grammatical errors and disjointed ideas.
        \item The summary is often unclear, with frequent grammatical errors and poor flow.
        \item The summary is moderately clear but has some grammatical errors and awkward transitions.
        \item The summary is mostly clear, with minor grammatical errors and mostly smooth transitions.
        \item The summary is exceptionally clear, grammatically perfect, and flows seamlessly.
    \end{enumerate}

    \item \textbf{Balance.} Does the summary address both sides of the debatable query by using counterarguments to present a well-rounded view?
    \begin{enumerate}
        \item The summary is heavily biased, with little to no use of counterarguments and only one side addressed effectively.
        \item The summary is poorly balanced, significantly favoring one side and using counterarguments ineffectively.
        \item The summary is somewhat balanced but has noticeable bias and some awkward or less effective counterarguments.
        \item The summary is mostly balanced, with minor bias and effective use of counterarguments.
        \item The summary is perfectly balanced, equally addressing both sides and effectively using counterarguments.
    \end{enumerate}
\end{itemize}

\subsubsection{Results}

\begin{figure*}
    \centering
    \fbox{
    \includegraphics[width=\linewidth]{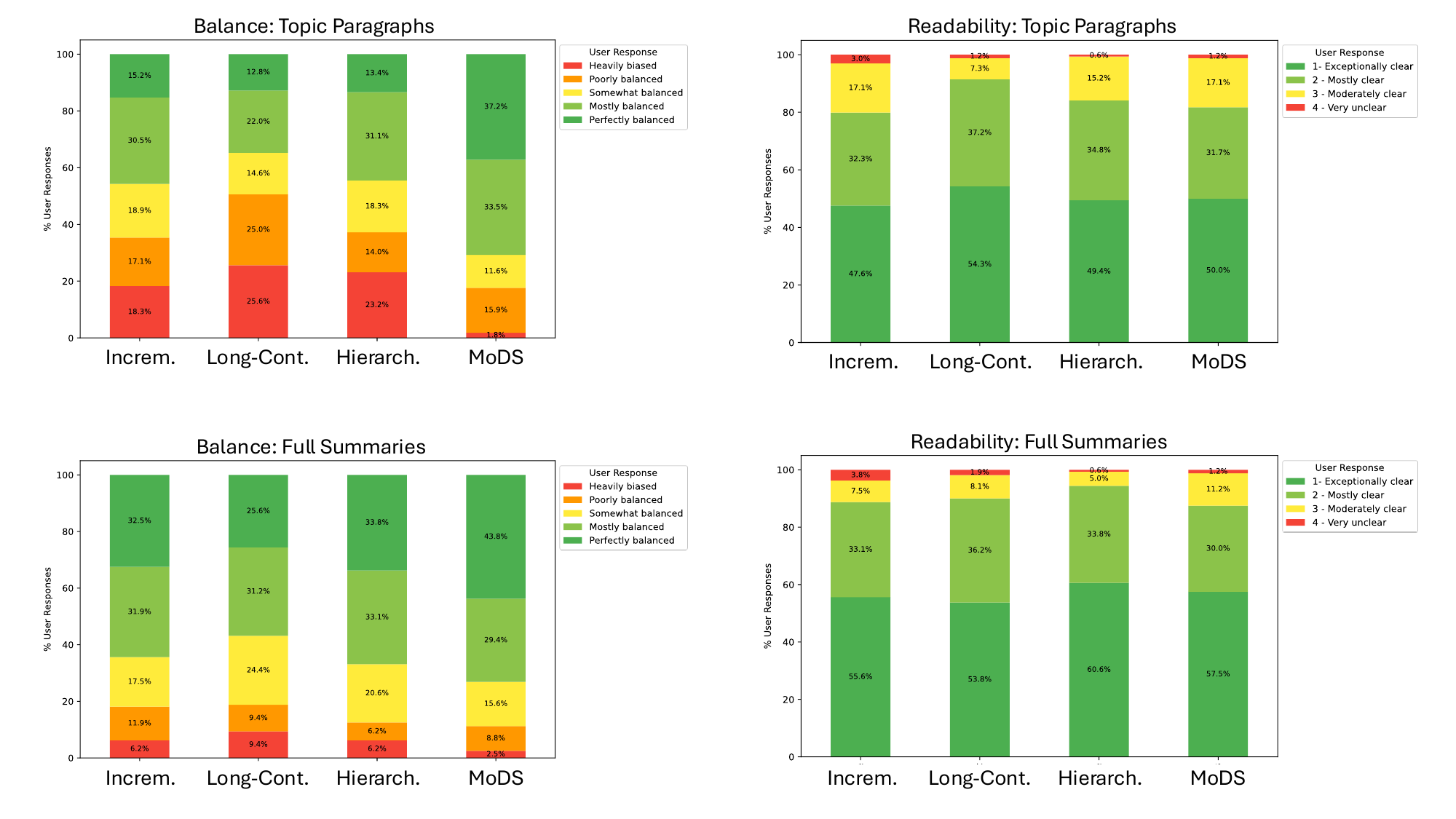}}
    \caption{\label{fig:annot} Distribution of Readability and Balance for Full Summaries and Topic Paragraphs from Prolific.}
\end{figure*}

Figure~\ref{fig:annot} shows the full distribution of Prolific annotations for Balance and Readability across Summaries and Topic Paragraphs.

\subsection{Sample Outputs} \label{appendix:outputs}

We present sample outputs generated by \model on ConflictingQA (Summary~\ref{summary1}, \ref{summary2}) and DebateQFS (Summary~\ref{summary3}, \ref{summary4}).
The summaries from \model have high coverage, citing several documents from the input collection, while also being balanced.
Further, the summary quality of \model remains high.
After comparing the summary for the EU expansion query in Figure~\ref{fig:intro} from 0-shot GPT-4 versus the summary from \model in Summary~\ref{summary3}, the balance, comprehensiveness, and quality gains from our method are clear.

\clearpage
\begin{table*}[]
\centering
\begin{tabular}{@{}lcccc@{}}
\toprule
\textbf{Dataset} & \textbf{\# Entries} & \textbf{Avg \# Docs} & \textbf{Avg \# Para. / Doc} & \textbf{Majority / Minority Stance Split} \\ \midrule
ConflictingQA & 290 & 10.468 & 57.725 & 0.649 / 0.351 \\
DebateQFS & 183 & 9.857 & 26.320 & 0.620 / 0.380 \\ \bottomrule
\end{tabular}
\caption{\label{appendix:table:dataset} Dataset statistics for ConflictingQA and DebateQFS.}
\end{table*}

\begin{algorithm*}
\caption{\model}\label{algo:mods}
\footnotesize
\begin{algorithmic}[1]
\Procedure{MoDS}{$q, \mathcal{D}, m, k$}

\State Initialize $\mathcal{O}$\Comment{Create outline}
\State $\mathcal{S} \gets \{\textsc{Speaker}(d_i) : d_i \in \mathcal{D} \}$\Comment{Create speakers}

\State \texttt{\# Agenda Planning (\cref{subsection:agenda})}
\State $\mathcal{B} \gets \{\textsc{Retrieve}(d_i, q, k) : d_i \in \mathcal{D} \}$

\State $\mathcal{T} \gets \textsc{Planner}(q, \mathcal{B}, m)$

\For{$t_j \in \mathcal{T}$}
    \State \texttt{\# Speaker Selection (\cref{subsection:moderator})}
    \State $\mathcal{B}_j \gets \{\textsc{Retrieve}(d_i, t_j, k) : d_i \in \mathcal{D} \}$
    \State $\mathcal{S}_j, \mathcal{Q}_j \gets \textsc{Moderator}(q, t_j, \mathcal{B}_j)$

    \For{$s_{i,j}, q_{i,j} \in (\mathcal{S}_j, \mathcal{Q}_j)$}
        \State \texttt{\# Speaker Discussion (\cref{subsection:speaker})}
        \State $\mathcal{P} \gets s_{i,j}(q, q_{i,j}, t_j)$
        \State $\mathcal{O} \gets \mathcal{O} \cup \{t_j, i, \mathcal{P}, q_{i, j}\}$ \Comment{Update outline}
    \EndFor
\EndFor

\State \texttt{\# Outline Summarization (\cref{subsection:summary})}
\State $\mathbb{S}_{all} \gets \textsc{Summarize}(\mathcal{O})$
\State $\mathbb{S}_{top} \gets \{\textsc{Summarize}(\mathcal{O}_j) : t_j \in \mathcal{T}\}$
\State \Return $\mathbb{S}_{all}, \mathbb{S}_{top}$\Comment{Return summaries to the user}

\EndProcedure
\end{algorithmic}
\end{algorithm*}
\begin{table*}[]
\small 
\centering
\begin{tabular}{@{}ll@{}}
\toprule
 & Rubric Text \\ \midrule
Criteria & Interest Level: How engaging and thought-provoking is the summary? \\
Score 1 & Not engaging at all; no attempt to capture the reader’s attention. \\
Score 2 & Fairly engaging with a basic narrative but lacking depth. \\
Score 3 & Moderately engaging with several interesting points. \\
Score 4 & Quite engaging with a well-structured narrative and noteworthy points that frequently capture and retain attention \\
Score 5 & Exceptionally engaging throughout, with a compelling narrative that consistently stimulates interest. \\ \midrule
Criteria & Coherence and Organization: Is the summary well-organized and logically structured? \\
Score 1 & Disorganized; lacks logical structure and coherence. \\
Score 2 & Fairly organized; a basic structure is present but not consistently followed. \\
Score 3 & Organized; a clear structure is mostly followed with some lapses in coherence. \\
Score 4 & Good organization; a clear structure with minor lapses in coherence. \\
Score 5 & Excellently organized; the summary is logically structured with seamless transitions and a clear argument. \\ \midrule
Criteria & Relevance and Focus: Does the summary stay on topic to the query and maintain a clear focus? \\
Score 1 & Off-topic; the content does not align with the query. \\
Score 2 & Somewhat on topic but with several digressions; the answer to the query is evident but not consistently adhered to. \\
Score 3 & Generally on topic, despite a few unrelated details. \\
Score 4 & Mostly on topic and focused; the narrative has a consistent relevance to the query with infrequent digressions. \\
Score 5 & \specialcellleft{Exceptionally focused and entirely on topic; the article is tightly centered on the query,\\with every piece of information contributing to a comprehensive understanding of the query.} \\ \midrule
Criteria & Broad Coverage: Does the article provide an in-depth exploration of the query and have good coverage? \\
Score 1 & Severely lacking; offers little to no coverage of the query's primary aspects, resulting in a very narrow perspective. \\
Score 2 & Partial coverage; includes some of the query's main aspects but misses others, resulting in an incomplete portrayal. \\
Score 3 & \specialcellleft{Acceptable breadth; covers most main aspects, though it may stray into minor unnecessary details\\ or overlook some relevant points.} \\
Score 4 & \specialcellleft{Good coverage; achieves broad coverage of the query,\\hitting on all major points with minimal extraneous information.} \\
Score 5 & \specialcellleft{Exemplary in breadth; delivers outstanding coverage,\\thoroughly detailing all crucial aspects of the query without including irrelevant information.} \\ \midrule
Criteria & \specialcellleft{Diversity of Perspectives: Does the summary adequately describe\\why the answer to the query could be yes and why it could be no?} \\
Score 1 & No diversity; the summary presents only one perspective without addressing the opposing viewpoint. \\
Score 2 & Limited diversity; the summary acknowledges both perspectives but lacks depth in the explanation of one side. \\
Score 3 & Moderate diversity; the summary covers both perspectives, but one side is more thoroughly explored than the other. \\
Score 4 & Good diversity; the summary fairly represents both perspectives with balanced and detailed explanations. \\
Score 5 & \specialcellleft{Excellent diversity; the summary provides a comprehensive and balanced exploration of both perspectives,\\offering in-depth explanations for why the answer could be yes and why it could be no.} \\ \bottomrule
\end{tabular}
\caption{\label{table:rubric} Rubrics for Interest, Coherence, Relevance, Coverage, and Diversity for DQFS summaries. Rubrics are adapted for topic paragraphs and topics (e.g. ``Relevance'' becomes relevance to the topic in topic paragraph evaluation, rather than relevance to the query).}
\end{table*}
\begin{table*}[!h]
\footnotesize
\centering
\setlength{\tabcolsep}{3.5pt}
\renewcommand{\arraystretch}{0.8}
\begin{tabular}{@{}clcccccccc@{}}
\multicolumn{1}{l}{} &  & \multicolumn{3}{c}{\textit{Summary Level}} & \multicolumn{3}{c}{\textit{Topic Paragraph Level}} & \multicolumn{2}{c}{\textit{Confounders}} \\ \midrule
\textbf{\# Pts} & \multicolumn{1}{l|}{\textbf{Model}} & \textbf{DC ($\uparrow$)} & \textbf{Fair ($\downarrow$)} & \multicolumn{1}{c|}{\textbf{Faithful ($\downarrow$)}} & \textbf{DC ($\uparrow$)} & \textbf{Fair ($\downarrow$)} & \multicolumn{1}{c|}{\textbf{Faithful ($\downarrow$)}} & \multicolumn{1}{l}{\textbf{Cite Acc.}} & \textbf{All / Avg Sents} \\ \midrule
\multirow{10}{*}{2} & \multicolumn{1}{l|}{\textbf{\modelAll (\textbf{Ours})}} & {\ul 0.811*} & 0.113* & \multicolumn{1}{c|}{{\ul 0.046*}} & {\ul 0.578*} & {\ul 0.171*} & \multicolumn{1}{c|}{{\ul 0.106*}} & 0.988 & 5.99 / 3.00 \\
 & \multicolumn{1}{l|}{\textbf{\modelTopic (\textbf{Ours})}} & \textbf{0.821*} & \textbf{0.108*} & \multicolumn{1}{c|}{\textbf{0.043*}} & \textbf{0.623} & \textbf{0.153*} & \multicolumn{1}{c|}{\textbf{0.090*}} & 0.985 & 6.01 / 3.01 \\
 & \multicolumn{1}{l|}{Long-Context} & 0.447 & 0.242 & \multicolumn{1}{c|}{0.198} & 0.277 & 0.369 & \multicolumn{1}{c|}{0.326} & 0.950 & 5.99 / 3.00 \\
 & \multicolumn{1}{l|}{RAG-\textit{All}} & 0.603 & 0.166 & \multicolumn{1}{c|}{0.098} & 0.378 & 0.285 & \multicolumn{1}{c|}{0.219} & 0.992 & 6.00 / 3.00 \\
 & \multicolumn{1}{l|}{RAG-\textit{Doc}} & 0.668 & 0.148 & \multicolumn{1}{c|}{0.078} & 0.415 & 0.273 & \multicolumn{1}{c|}{0.204} & 0.970 & 6.02 / 3.01 \\
 & \multicolumn{1}{l|}{Hierarchical} & 0.765 & {\ul 0.111*} & \multicolumn{1}{c|}{0.048*} & 0.454 & 0.265 & \multicolumn{1}{c|}{0.204} & 0.985 & 6.00 / 3.00 \\
 & \multicolumn{1}{l|}{Incremental-\textit{All}} & 0.464 & 0.249 & \multicolumn{1}{c|}{0.202} & 0.357 & 0.289 & \multicolumn{1}{c|}{0.244} & 0.971 & 5.99 / 3.00 \\
 & \multicolumn{1}{l|}{Incremental-\textit{Topic}} & 0.512 & 0.230 & \multicolumn{1}{c|}{0.182} & 0.419 & 0.262 & \multicolumn{1}{c|}{0.215} & 0.977 & 6.00 / 3.00 \\
 & \multicolumn{1}{l|}{Cluster} & 0.586 & 0.168 & \multicolumn{1}{c|}{0.126} & 0.356 & 0.309 & \multicolumn{1}{c|}{0.269} & 0.927 & 6.01 / 3.01 \\
 & \multicolumn{1}{l|}{RAG+Cluster} & 0.665 & 0.151 & \multicolumn{1}{c|}{0.078} & 0.417 & 0.269 & \multicolumn{1}{c|}{0.198} & 0.979 & 6.04 / 3.02 \\ \midrule
\multirow{10}{*}{3} & \multicolumn{1}{l|}{\textbf{\modelAll (\textbf{Ours})}} & {\ul 0.8664} & 0.1062* & \multicolumn{1}{c|}{0.0359*} & {\ul 0.5420} & {\ul 0.1896*} & \multicolumn{1}{c|}{{\ul 0.1217}} & 0.988 & 8.97 / 2.99 \\
 & \multicolumn{1}{l|}{\textbf{\modelTopic (\textbf{Ours})}} & \textbf{0.8961*} & {\ul 0.0998*} & \multicolumn{1}{c|}{\textbf{0.0320*}} & \textbf{0.6056*} & \textbf{0.1650*} & \multicolumn{1}{c|}{\textbf{0.0979}} & 0.985 & 8.99 / 3.00 \\
 & \multicolumn{1}{l|}{Long-Context} & 0.5242 & 0.2047 & \multicolumn{1}{c|}{0.1733} & 0.2566 & 0.3816 & \multicolumn{1}{c|}{0.3503} & 0.958 & 9.00 / 3.00 \\
 & \multicolumn{1}{l|}{RAG-\textit{All}} & 0.6565 & 0.1664 & \multicolumn{1}{c|}{0.0911} & 0.3300 & 0.3296 & \multicolumn{1}{c|}{0.2547} & 0.990 & 9.01 / 3.00 \\
 & \multicolumn{1}{l|}{RAG-\textit{Doc}} & 0.7532 & 0.1364 & \multicolumn{1}{c|}{0.0668} & 0.3741 & 0.3023 & \multicolumn{1}{c|}{0.2352} & 0.949 & 9.01 / 3.00 \\
 & \multicolumn{1}{l|}{Hierarchical} & 0.8158 & \textbf{0.0956*} & \multicolumn{1}{c|}{{\ul 0.0333*}} & 0.3679 & 0.3136 & \multicolumn{1}{c|}{0.2523} & 0.981 & 8.99 / 3.00 \\
 & \multicolumn{1}{l|}{Incremental-\textit{All}} & 0.5037 & 0.2466 & \multicolumn{1}{c|}{0.1924} & 0.3467 & 0.3019 & \multicolumn{1}{c|}{0.2488} & 0.961 & 8.99 / 3.00 \\
 & \multicolumn{1}{l|}{Incremental-\textit{Topic}} & 0.5635 & 0.2288 & \multicolumn{1}{c|}{0.1720} & 0.4209 & 0.2796 & \multicolumn{1}{c|}{0.2236} & 0.963 & 9.01 / 3.00 \\
 & \multicolumn{1}{l|}{Cluster} & 0.7142 & 0.1203* & \multicolumn{1}{c|}{0.0662} & 0.3502 & 0.3016 & \multicolumn{1}{c|}{0.2517} & 0.927 & 9.04 / 3.01 \\
 & \multicolumn{1}{l|}{RAG+Cluster} & 0.7694 & 0.1332 & \multicolumn{1}{c|}{0.0620} & 0.3906 & 0.2808 & \multicolumn{1}{c|}{0.2101} & 0.976 & 9.02 / 3.01 \\ \midrule
\multirow{10}{*}{4} & \multicolumn{1}{l|}{\textbf{\modelAll (\textbf{Ours})}} & {\ul 0.8991} & {\ul 0.0976*} & \multicolumn{1}{c|}{{\ul 0.0301*}} & {\ul 0.5107} & {\ul 0.1886*} & \multicolumn{1}{c|}{{\ul 0.1225*}} & 0.987 & 11.92 / 2.98 \\
 & \multicolumn{1}{l|}{\textbf{\modelTopic (\textbf{Ours})}} & \textbf{0.9307*} & \textbf{0.0907*} & \multicolumn{1}{c|}{\textbf{0.0263*}} & \textbf{0.5954*} & \textbf{0.1653*} & \multicolumn{1}{c|}{\textbf{0.1022*}} & 0.982 & 12.00 / 3.00 \\
 & \multicolumn{1}{l|}{Long-Context} & 0.5594 & 0.1953 & \multicolumn{1}{c|}{0.1501} & 0.2342 & 0.4204 & \multicolumn{1}{c|}{0.3779} & 0.953 & 12.03 / 3.01 \\
 & \multicolumn{1}{l|}{RAG-\textit{All}} & 0.7065 & 0.1485 & \multicolumn{1}{c|}{0.0801} & 0.2987 & 0.3556 & \multicolumn{1}{c|}{0.2891} & 0.997 & 12.02 / 3.00 \\
 & \multicolumn{1}{l|}{RAG-\textit{Doc}} & 0.7638 & 0.1357 & \multicolumn{1}{c|}{0.0631} & 0.3293 & 0.3427 & \multicolumn{1}{c|}{0.2725} & 0.961 & 12.01 / 3.00 \\
 & \multicolumn{1}{l|}{Hierarchical} & 0.8643 & 0.1008* & \multicolumn{1}{c|}{0.0325*} & 0.3204 & 0.3439 & \multicolumn{1}{c|}{0.2768} & 0.983 & 12.02 / 3.01 \\
 & \multicolumn{1}{l|}{Incremental-\textit{All}} & 0.4994 & 0.2589 & \multicolumn{1}{c|}{0.1999} & 0.3208 & 0.3200 & \multicolumn{1}{c|}{0.2602} & 0.950 & 11.97 / 2.99 \\
 & \multicolumn{1}{l|}{Incremental-\textit{Topic}} & 0.5611 & 0.2274 & \multicolumn{1}{c|}{0.1703} & 0.3896 & 0.2931 & \multicolumn{1}{c|}{0.2365} & 0.954 & 12.00 / 3.00 \\
 & \multicolumn{1}{l|}{Cluster} & 0.7907 & 0.1108* & \multicolumn{1}{c|}{0.0577} & 0.3485 & 0.3068 & \multicolumn{1}{c|}{0.2557} & 0.931 & 12.02 / 3.01 \\
 & \multicolumn{1}{l|}{RAG+Cluster} & 0.8266 & 0.1175 & \multicolumn{1}{c|}{0.0527} & 0.3614 & 0.3002 & \multicolumn{1}{c|}{0.2393} & 0.977 & 12.03 / 3.01 \\ \midrule
\multirow{10}{*}{5} & \multicolumn{1}{l|}{\textbf{\modelAll (\textbf{Ours})}} & {\ul 0.9156} & {\ul 0.0966*} & \multicolumn{1}{c|}{{\ul 0.0272*}} & {\ul 0.4809} & {\ul 0.1972} & \multicolumn{1}{c|}{{\ul 0.1297}} & 0.990 & 14.88 / 2.98 \\
 & \multicolumn{1}{l|}{\textbf{\modelTopic (\textbf{Ours})}} & \textbf{0.9549*} & \textbf{0.0884*} & \multicolumn{1}{c|}{\textbf{0.0239*}} & \textbf{0.5924*} & \textbf{0.1661*} & \multicolumn{1}{c|}{\textbf{0.1051*}} & 0.986 & 15.00 / 3.00 \\
 & \multicolumn{1}{l|}{Long-Context} & 0.5779 & 0.2038 & \multicolumn{1}{c|}{0.1622} & 0.2164 & 0.4620 & \multicolumn{1}{c|}{0.4213} & 0.966 & 15.00 / 3.00 \\
 & \multicolumn{1}{l|}{RAG-\textit{All}} & 0.7331 & 0.1581 & \multicolumn{1}{c|}{0.0814} & 0.2755 & 0.3850 & \multicolumn{1}{c|}{0.3101} & 0.996 & 15.03 / 3.01 \\
 & \multicolumn{1}{l|}{RAG-\textit{Doc}} & 0.7898 & 0.1464 & \multicolumn{1}{c|}{0.0706} & 0.3018 & 0.3691 & \multicolumn{1}{c|}{0.2945} & 0.975 & 15.06 / 3.01 \\
 & \multicolumn{1}{l|}{Hierarchical} & 0.8871 & 0.0931* & \multicolumn{1}{c|}{0.0276*} & 0.2951 & 0.3670 & \multicolumn{1}{c|}{0.3038} & 0.987 & 15.01 / 3.00 \\
 & \multicolumn{1}{l|}{Incremental-\textit{All}} & 0.5392 & 0.2327 & \multicolumn{1}{c|}{0.1738} & 0.3083 & 0.3236 & \multicolumn{1}{c|}{0.2672} & 0.948 & 14.91 / 2.98 \\
 & \multicolumn{1}{l|}{Incremental-\textit{Topic}} & 0.6239 & 0.1899 & \multicolumn{1}{c|}{0.1337} & 0.3961 & 0.2902 & \multicolumn{1}{c|}{0.2348} & 0.958 & 14.99 / 3.00 \\
 & \multicolumn{1}{l|}{Cluster} & 0.8480 & 0.0968* & \multicolumn{1}{c|}{0.0464} & 0.3365 & 0.3093 & \multicolumn{1}{c|}{0.2625} & 0.933 & 15.04 / 3.01 \\
 & \multicolumn{1}{l|}{RAG+Cluster} & 0.8717 & 0.1084* & \multicolumn{1}{c|}{0.0436} & 0.3499 & 0.3136 & \multicolumn{1}{c|}{0.2511} & 0.971 & 15.03 / 3.01 \\ \bottomrule
\end{tabular}
\caption{\label{table:doc_cover_cqa_all}ConflictingQA citation coverage, balance, and accuracy. Best model is \textbf{bold}, second best is \underline{underlined}. Models with * are significantly the best (2-sample $t$-test, $p<0.05$ with Bonferroni correction).}
\end{table*}

\begin{table*}[!h]
\footnotesize
\centering
\setlength{\tabcolsep}{3.5pt}
\renewcommand{\arraystretch}{0.8}
\begin{tabular}{@{}clcccccccc@{}}
\multicolumn{1}{l}{} &  & \multicolumn{3}{c}{\textit{Summary Level}} & \multicolumn{3}{c}{\textit{Topic Paragraph Level}} & \multicolumn{2}{c}{\textit{Confounders}} \\ \midrule
\textbf{\# Pts} & \multicolumn{1}{l|}{\textbf{Model}} & \textbf{DC ($\uparrow$)} & \textbf{Fair ($\downarrow$)} & \multicolumn{1}{c|}{\textbf{Faithful ($\downarrow$)}} & \textbf{DC ($\uparrow$)} & \textbf{Fair ($\downarrow$)} & \multicolumn{1}{c|}{\textbf{Faithful ($\downarrow$)}} & \multicolumn{1}{l}{\textbf{Cite Acc}} & \textbf{All / Avg Sents} \\ \midrule
\multirow{10}{*}{2} & \multicolumn{1}{l|}{\modelTopic (\textbf{Ours})} & \textbf{0.798*} & \textbf{0.088*} & \multicolumn{1}{c|}{\textbf{0.036*}} & \textbf{0.614*} & \textbf{0.132*} & \multicolumn{1}{c|}{\textbf{0.078*}} & 0.991 & 5.99 / 3.00 \\
 & \multicolumn{1}{l|}{\modelAll (\textbf{Ours})} & {\ul 0.789*} & {\ul 0.098*} & \multicolumn{1}{c|}{{\ul 0.040*}} & {\ul 0.582*} & {\ul 0.150*} & \multicolumn{1}{c|}{{\ul 0.092*}} & 0.992 & 5.96 / 2.98 \\
 & \multicolumn{1}{l|}{Long-Context} & 0.506 & 0.254 & \multicolumn{1}{c|}{0.212} & 0.302 & 0.423 & \multicolumn{1}{c|}{0.385} & 0.976 & 6.01 / 3.00 \\
 & \multicolumn{1}{l|}{RAG-\textit{All}} & 0.529 & 0.183 & \multicolumn{1}{c|}{0.139} & 0.347 & 0.295 & \multicolumn{1}{c|}{0.251} & 0.995 & 6.01 / 3.00 \\
 & \multicolumn{1}{l|}{RAG-\textit{Doc}} & 0.630 & 0.142 & \multicolumn{1}{c|}{0.095} & 0.374 & 0.325 & \multicolumn{1}{c|}{0.280} & 0.991 & 5.99 / 3.00 \\
 & \multicolumn{1}{l|}{Hierarchical} & 0.710 & 0.104* & \multicolumn{1}{c|}{0.053*} & 0.421 & 0.261 & \multicolumn{1}{c|}{0.209} & 0.983 & 6.00 / 3.00 \\
 & \multicolumn{1}{l|}{Incremental-\textit{All}} & 0.497 & 0.326 & \multicolumn{1}{c|}{0.291} & 0.405 & 0.348 & \multicolumn{1}{c|}{0.313} & 0.981 & 6.01 / 3.00 \\
 & \multicolumn{1}{l|}{Incremental-\textit{Topic}} & 0.548 & 0.297 & \multicolumn{1}{c|}{0.266} & 0.459 & 0.338 & \multicolumn{1}{c|}{0.307} & 0.982 & 6.00 / 3.00 \\
 & \multicolumn{1}{l|}{Cluster} & 0.610 & 0.133 & \multicolumn{1}{c|}{0.102} & 0.384 & 0.297 & \multicolumn{1}{c|}{0.266} & 0.966 & 6.01 / 3.00 \\
 & \multicolumn{1}{l|}{RAG+Cluster} & 0.572 & 0.166 & \multicolumn{1}{c|}{0.121} & 0.354 & 0.306 & \multicolumn{1}{c|}{0.260} & 0.986 & 6.02 / 3.01 \\ \midrule
\multirow{10}{*}{3} & \multicolumn{1}{l|}{\modelTopic (\textbf{Ours})} & \textbf{0.8724*} & \textbf{0.0701*} & \multicolumn{1}{c|}{\textbf{0.0235*}} & \textbf{0.6066*} & \textbf{0.1255*} & \multicolumn{1}{c|}{\textbf{0.0789*}} & 0.982 & 8.99 / 3.00 \\
\multicolumn{1}{l}{} & \multicolumn{1}{l|}{\modelAll (\textbf{Ours})} & {\ul 0.8457*} & {\ul 0.0786*} & \multicolumn{1}{c|}{{\ul 0.0273*}} & {\ul 0.5508} & {\ul 0.1463*} & \multicolumn{1}{c|}{{\ul 0.0938*}} & 0.987 & 8.87 / 2.96 \\
\multicolumn{1}{l}{} & \multicolumn{1}{l|}{Long-Context} & 0.5877 & 0.2094 & \multicolumn{1}{c|}{0.1790} & 0.2798 & 0.4336 & \multicolumn{1}{c|}{0.4028} & 0.953 & 9.02 / 3.01 \\
\multicolumn{1}{l}{} & \multicolumn{1}{l|}{RAG-\textit{All}} & 0.6125 & 0.1544 & \multicolumn{1}{c|}{0.1040} & 0.3229 & 0.3176 & \multicolumn{1}{c|}{0.2701} & 0.997 & 9.01 / 3.00 \\
\multicolumn{1}{l}{} & \multicolumn{1}{l|}{RAG-\textit{Doc}} & 0.7171 & 0.1180 & \multicolumn{1}{c|}{0.0664} & 0.3504 & 0.3233 & \multicolumn{1}{c|}{0.2748} & 0.961 & 9.01 / 3.00 \\
\multicolumn{1}{l}{} & \multicolumn{1}{l|}{Hierarchical} & 0.7868 & 0.0907 & \multicolumn{1}{c|}{0.0374} & 0.3639 & 0.2980 & \multicolumn{1}{c|}{0.2452} & 0.983 & 9.02 / 3.01 \\
\multicolumn{1}{l}{} & \multicolumn{1}{l|}{Incremental-\textit{All}} & 0.5566 & 0.2579 & \multicolumn{1}{c|}{0.2089} & 0.3919 & 0.3243 & \multicolumn{1}{c|}{0.2765} & 0.950 & 8.91 / 2.97 \\
\multicolumn{1}{l}{} & \multicolumn{1}{l|}{Incremental-\textit{Topic}} & 0.6152 & 0.2415 & \multicolumn{1}{c|}{0.1970} & 0.4707 & 0.3128 & \multicolumn{1}{c|}{0.2674} & 0.954 & 9.03 / 3.01 \\
\multicolumn{1}{l}{} & \multicolumn{1}{l|}{Cluster} & 0.7102 & 0.1106 & \multicolumn{1}{c|}{0.0725} & 0.3632 & 0.3106 & \multicolumn{1}{c|}{0.2737} & 0.931 & 9.04 / 3.01 \\
\multicolumn{1}{l}{} & \multicolumn{1}{l|}{RAG+Cluster} & 0.6811 & 0.1405 & \multicolumn{1}{c|}{0.0894} & 0.3428 & 0.3200 & \multicolumn{1}{c|}{0.2689} & 0.977 & 9.01 / 3.00 \\ \midrule
\multirow{10}{*}{4} & \multicolumn{1}{l|}{\modelTopic (\textbf{Ours})} & \textbf{0.8895*} & {\ul 0.0724*} & \multicolumn{1}{c|}{\textbf{0.0209*}} & \textbf{0.5844*} & \textbf{0.1385*} & \multicolumn{1}{c|}{\textbf{0.0868*}} & 0.987 & 11.98 / 3.00 \\
 & \multicolumn{1}{l|}{\modelAll (\textbf{Ours})} & {\ul 0.8653*} & \textbf{0.0697*} & \multicolumn{1}{c|}{{\ul 0.0216*}} & {\ul 0.5230} & {\ul 0.1419*} & \multicolumn{1}{c|}{{\ul 0.0925*}} & 0.990 & 11.86 / 2.96 \\
 & \multicolumn{1}{l|}{Long-Context} & 0.6361 & 0.1691 & \multicolumn{1}{c|}{0.1471} & 0.2473 & 0.4733 & \multicolumn{1}{c|}{0.4479} & 0.977 & 12.03 / 3.01 \\
 & \multicolumn{1}{l|}{RAG-\textit{All}} & 0.6595 & 0.1440 & \multicolumn{1}{c|}{0.0969} & 0.2916 & 0.3603 & \multicolumn{1}{c|}{0.3149} & 0.995 & 12.03 / 3.01 \\
 & \multicolumn{1}{l|}{RAG-\textit{Doc}} & 0.7335 & 0.1218 & \multicolumn{1}{c|}{0.0723} & 0.3113 & 0.3635 & \multicolumn{1}{c|}{0.3171} & 0.991 & 12.03 / 3.01 \\
 & \multicolumn{1}{l|}{Hierarchical} & 0.8338 & 0.0845* & \multicolumn{1}{c|}{0.0325} & 0.3269 & 0.3331 & \multicolumn{1}{c|}{0.2813} & 0.986 & 12.02 / 3.01 \\
 & \multicolumn{1}{l|}{Incremental-\textit{All}} & 0.5716 & 0.2352 & \multicolumn{1}{c|}{0.1874} & 0.3795 & 0.3193 & \multicolumn{1}{c|}{0.2736} & 0.963 & 11.87 / 2.97 \\
 & \multicolumn{1}{l|}{Incremental-\textit{Topic}} & 0.6331 & 0.2129 & \multicolumn{1}{c|}{0.1629} & 0.4514 & 0.3133 & \multicolumn{1}{c|}{0.2658} & 0.970 & 11.98 / 2.99 \\
 & \multicolumn{1}{l|}{Cluster} & 0.7744 & 0.1129 & \multicolumn{1}{c|}{0.0698} & 0.3451 & 0.3181 & \multicolumn{1}{c|}{0.2752} & 0.964 & 12.03 / 3.01 \\
 & \multicolumn{1}{l|}{RAG+Cluster} & 0.7305 & 0.1218 & \multicolumn{1}{c|}{0.0746} & 0.3237 & 0.3459 & \multicolumn{1}{c|}{0.3029} & 0.989 & 12.04 / 3.01 \\ \midrule
\multirow{10}{*}{5} & \multicolumn{1}{l|}{\modelTopic (\textbf{Ours})} & \textbf{0.9137*} & {\ul 0.0651*} & \multicolumn{1}{c|}{\textbf{0.0208*}} & \textbf{0.5793*} & \textbf{0.1420*} & \multicolumn{1}{c|}{\textbf{0.0998*}} & 0.986 & 14.99 / 3.00 \\
 & \multicolumn{1}{l|}{\modelAll (\textbf{Ours})} & {\ul 0.8847*} & \textbf{0.0640*} & \multicolumn{1}{c|}{{\ul 0.0236*}} & {\ul 0.4991} & {\ul 0.1502*} & \multicolumn{1}{c|}{{\ul 0.1096*}} & 0.990 & 14.46 / 2.89 \\
 & \multicolumn{1}{l|}{Long-Context} & 0.6686 & 0.1724 & \multicolumn{1}{c|}{0.1392} & 0.2312 & 0.4965 & \multicolumn{1}{c|}{0.4640} & 0.966 & 15.01 / 3.00 \\
 & \multicolumn{1}{l|}{RAG-\textit{All}} & 0.6721 & 0.1423 & \multicolumn{1}{c|}{0.0912} & 0.2668 & 0.3927 & \multicolumn{1}{c|}{0.3438} & 0.996 & 15.02 / 3.00 \\
 & \multicolumn{1}{l|}{RAG-\textit{Doc}} & 0.7765 & 0.1053 & \multicolumn{1}{c|}{0.0618} & 0.3005 & 0.3584 & \multicolumn{1}{c|}{0.3147} & 0.975 & 15.01 / 3.00 \\
 & \multicolumn{1}{l|}{Hierarchical} & 0.8565 & 0.0761* & \multicolumn{1}{c|}{0.0239*} & 0.2896 & 0.3713 & \multicolumn{1}{c|}{0.3192} & 0.987 & 15.04 / 3.01 \\
 & \multicolumn{1}{l|}{Incremental-\textit{All}} & 0.6122 & 0.2000 & \multicolumn{1}{c|}{0.1629} & 0.3716 & 0.2936 & \multicolumn{1}{c|}{0.2572} & 0.948 & 14.77 / 2.95 \\
 & \multicolumn{1}{l|}{Incremental-\textit{Topic}} & 0.6767 & 0.1659 & \multicolumn{1}{c|}{0.1198} & 0.4446 & 0.2897 & \multicolumn{1}{c|}{0.2443} & 0.958 & 15.05 / 3.01 \\
 & \multicolumn{1}{l|}{Cluster} & 0.8098 & 0.1116 & \multicolumn{1}{c|}{0.0624} & 0.3292 & 0.3383 & \multicolumn{1}{c|}{0.2921} & 0.933 & 15.03 / 3.01 \\
 & \multicolumn{1}{l|}{RAG+Cluster} & 0.7811 & 0.1233 & \multicolumn{1}{c|}{0.0738} & 0.3129 & 0.3588 & \multicolumn{1}{c|}{0.3107} & 0.971 & 15.03 / 3.01 \\ \bottomrule
\end{tabular}
\caption{\label{table:doc_cover_debate_all}DebateQFS citation coverage, balance, and accuracy. Best model is \textbf{bold}, second best is \underline{underlined}. Models with * are significantly the best (2-sample $t$-test, $p<0.05$ with Bonferroni correction).}
\end{table*}

\begin{table*}[!h]
\footnotesize
\centering
\setlength{\tabcolsep}{3.5pt}
\renewcommand{\arraystretch}{0.8}
\begin{tabular}{@{}clcccccccc@{}}
\multicolumn{1}{l}{} &  & \multicolumn{3}{c}{{\textit{Summary Level}}} & \multicolumn{3}{c}{{\textit{Topic Paragraph Level}}} & \multicolumn{2}{c}{{\textit{Confounders}}} \\ \midrule
{\# Pts} & \multicolumn{1}{l|}{{Model}} & {\textbf{DC} ($\uparrow$)} & {\textbf{Fair} ($\downarrow$)} & \multicolumn{1}{c|}{{\textbf{Faithful} ($\downarrow$)}} & {\textbf{DC} ($\uparrow$)} & {\textbf{Fair} ($\downarrow$)} & \multicolumn{1}{c|}{{\textbf{Faithful} ($\downarrow$)}} & \multicolumn{1}{l}{{\textbf{Cite Acc}.}} & {\textbf{All / Avg Sents}} \\ \midrule

\multirow{2}{*}{3} & \multicolumn{1}{l|}{{\modelTopic ({Ours})}} & {0.8961} & {0.0998} & \multicolumn{1}{c|}{{0.0320}} & {0.6056} & {0.1650} & \multicolumn{1}{c|}{{0.0979}} & 0.985 & 8.99 / 3.00 \\
 & \multicolumn{1}{l|}{Hierarchical-\textit{Topic}} & 0.8761 & {0.1065} & \multicolumn{1}{c|}{{0.0467	}} & 0.6003 & 0.1688 & \multicolumn{1}{c|}{0.1130} & 0.985 & 8.98 / 2.99 \\ \midrule

\multirow{2}{*}{5} & \multicolumn{1}{l|}{{\modelTopic ({Ours})}} & {0.9549} & {0.0884} & \multicolumn{1}{c|}{{0.0239}} & {0.5924} & {0.1661} & \multicolumn{1}{c|}{{0.1051}} & 0.986 & 15.00 / 3.00 \\
 & \multicolumn{1}{l|}{Hierarchical-\textit{Topic}} & 0.9386 & 0.0996 & \multicolumn{1}{c|}{0.0310} & 0.5774 & 0.1952 & \multicolumn{1}{c|}{0.1304} & 0.987 & 15.01 / 3.00  \\ \bottomrule
\end{tabular}
\caption{\label{table:doc_cover_cqa_all_comp}ConflictingQA citation coverage, balance, and accuracy of \modelTopic versus Hierarchical Merging-\emph{Topic}, which runs hierarchical merging for each topic paragraph. \model consistently outperforms Hierarchal Merging.}
\end{table*}

\begin{table*}[!h]
\footnotesize
\centering
\setlength{\tabcolsep}{3.5pt}
\renewcommand{\arraystretch}{0.8}
\begin{tabular}{@{}clcccccccc@{}}
\multicolumn{1}{l}{} &  & \multicolumn{3}{c}{{\textit{Summary Level}}} & \multicolumn{3}{c}{{\textit{Topic Paragraph Level}}} & \multicolumn{2}{c}{{\textit{Confounders}}} \\ \midrule
{\# Pts} & \multicolumn{1}{l|}{{Model}} & {\textbf{DC} ($\uparrow$)} & {\textbf{Fair} ($\downarrow$)} & \multicolumn{1}{c|}{{\textbf{Faithful} ($\downarrow$)}} & {\textbf{DC} ($\uparrow$)} & {\textbf{Fair} ($\downarrow$)} & \multicolumn{1}{c|}{{\textbf{Faithful} ($\downarrow$)}} & \multicolumn{1}{l}{{\textbf{Cite Acc}}} & {\textbf{All / Avg Sents}} \\ \midrule

\multirow{2}{*}{3} & \multicolumn{1}{l|}{\modelTopic ({Ours})} & {0.8724} & {0.0701} & \multicolumn{1}{c|}{{0.0235}} & {0.6066} & {0.1255} & \multicolumn{1}{c|}{{0.0789}} & 0.982 & 8.99 / 3.00 \\
\multicolumn{1}{l}{} & \multicolumn{1}{l|}{Hierarchical-\textit{Topic}} & 0.7776 & 0.0965 & \multicolumn{1}{c|}{0.0483} & 0.4964 & 0.2177 & \multicolumn{1}{c|}{0.1688} & 0.983 & 9.00 / 3.00\\ \midrule

\multirow{2}{*}{5} & \multicolumn{1}{l|}{\modelTopic ({Ours})} & {0.9137} & { 0.0651} & \multicolumn{1}{c|}{{0.0208}} & {0.5793} & {0.1420} & \multicolumn{1}{c|}{{0.0998}} & 0.986 & 14.99 / 3.00 \\
 
 & \multicolumn{1}{l|}{Hierarchical-\textit{Topic}} & 0.8427 & 0.0951 & \multicolumn{1}{c|}{0.0431} & 0.4669 & 0.2397 & \multicolumn{1}{c|}{0.1909} & 0.984 & 14.90 / 2.98 \\ \bottomrule
\end{tabular}
\caption{\label{table:doc_cover_debate_all_comp}DebateQFS citation coverage, balance, and accuracy of \modelTopic versus Hierarchical Merging-\emph{Topic}, which runs hierarchical merging for each topic paragraph. \model consistently outperforms Hierarchal Merging.}
\end{table*}

\begin{table*}[!h]
\footnotesize
\centering
\setlength{\tabcolsep}{3.5pt}
\renewcommand{\arraystretch}{0.8}
\begin{tabular}{@{}clcccccc@{}}
\multicolumn{1}{l}{} &  & \multicolumn{3}{c}{{\textit{Summary Level}}} & \multicolumn{3}{c}{{\textit{Topic Paragraph Level}}} \\ \midrule
{\# Pts} & \multicolumn{1}{l|}{{Model}} & {\textbf{DC} ($\uparrow$)} & {\textbf{Fair} ($\downarrow$)} & \multicolumn{1}{c|}{{\textbf{Faithful} ($\downarrow$)}} & {\textbf{DC} ($\uparrow$)} & {\textbf{Fair} ($\downarrow$)} & \multicolumn{1}{c}{{\textbf{Faithful} ($\downarrow$)}} \\ \midrule

\multirow{2}{*}{3} & \multicolumn{1}{l|}{{\modelTopic (GPT-4) }} & {0.8961} & {0.0998} & \multicolumn{1}{c|}{{0.0320}} & {0.6056} & {0.1650} & \multicolumn{1}{c}{{0.0979}} \\
 & \multicolumn{1}{l|}{\modelTopic (GPT-4 mini)} & 0.8761 & {0.1065} & \multicolumn{1}{c|}{{0.0467	}} & 0.6003 & 0.1688 & \multicolumn{1}{c}{0.1130} \\ \midrule

\multirow{2}{*}{5} & \multicolumn{1}{l|}{{\modelTopic (GPT-4) }} & {0.9549} & {0.0884} & \multicolumn{1}{c|}{{0.0239}} & {0.5924} & {0.1661} & \multicolumn{1}{c}{{0.1051}} \\
 & \multicolumn{1}{l|}{\modelTopic (GPT-4 mini)} & 0.7841 & 0.1226 & \multicolumn{1}{c|}{0.0634} & 0.4320 & 0.2112	 & \multicolumn{1}{c}{0.1533}  \\ \bottomrule
\end{tabular}
\caption{\label{table:doc_cover_cqa_all_comp_mini}ConflictingQA citation coverage, balance, and accuracy of \modelTopic using GPT-4 versus \modelTopic using GPT-4 mini.}
\end{table*}

\begin{table*}[!h]
\footnotesize
\centering
\setlength{\tabcolsep}{3.5pt}
\renewcommand{\arraystretch}{0.8}
\begin{tabular}{@{}clcccccc@{}}
\multicolumn{1}{l}{} &  & \multicolumn{3}{c}{{\textit{Summary Level}}} & \multicolumn{3}{c}{{\textit{Topic Paragraph Level}}} \\ \midrule
{\# Pts} & \multicolumn{1}{l|}{{Model}} & {\textbf{DC} ($\uparrow$)} & {\textbf{Fair} ($\downarrow$)} & \multicolumn{1}{c|}{{\textbf{Faithful} ($\downarrow$)}} & {\textbf{DC} ($\uparrow$)} & {\textbf{Fair} ($\downarrow$)} & \multicolumn{1}{c}{{\textbf{Faithful} ($\downarrow$)}} \\ \midrule

\multirow{2}{*}{3} & \multicolumn{1}{l|}{\modelTopic (GPT-4) } & {0.8724} & {0.0701} & \multicolumn{1}{c|}{{0.0235}} & {0.6066} & {0.1255} & \multicolumn{1}{c}{{0.0789}} \\
\multicolumn{1}{l}{} & \multicolumn{1}{l|}{\modelTopic (GPT-4 mini)} & 0.7322	 & 0.1284 & \multicolumn{1}{c|}{0.1059} & 0.4788 & 0.2271 & \multicolumn{1}{c}{0.2066} \\ \midrule

\multirow{2}{*}{5} & \multicolumn{1}{l|}{\modelTopic (GPT-4) } & {0.9137} & { 0.0651} & \multicolumn{1}{c|}{{0.0208}} & {0.5793} & {0.1420} & \multicolumn{1}{c}{{0.0998}} \\
 
 & \multicolumn{1}{l|}{\modelTopic (GPT-4 mini)} & 0.8324	 & 0.0686	 & \multicolumn{1}{c|}{0.0686} & 0.4818 & 0.2260	 & \multicolumn{1}{c}{0.2260	} \\ \bottomrule
\end{tabular}
\caption{\label{table:doc_cover_debate_all_comp_mini}DebateQFS citation coverage, balance, and accuracy of \modelTopic using GPT-4 versus \modelTopic using GPT-4 mini.}
\end{table*}

\begin{table*}[t]
\footnotesize
\centering
\setlength{\tabcolsep}{2.75pt}
\renewcommand{\arraystretch}{0.6}
\begin{tabular}{@{}clcccccc@{}}
\multicolumn{1}{l}{} &  & \multicolumn{3}{c}{\textit{Summary Level}} & \multicolumn{3}{c}{\textit{Topic Paragraph Level}} \\ \toprule
\textbf{\# Top.} & \multicolumn{1}{l|}{\textbf{Model}} & \textbf{DC ($\uparrow$)} & \textbf{Fair ($\downarrow$)} & \multicolumn{1}{c|}{\textbf{\begin{tabular}[c]{@{}c@{}}Faithful ($\downarrow$)\end{tabular}}} & \textbf{DC ($\uparrow$)} & \textbf{Fair ($\downarrow$)} & \multicolumn{1}{c}{\textbf{\begin{tabular}[c]{@{}c@{}}Faithful ($\downarrow$)\end{tabular}}} \\ \midrule
 & \multicolumn{1}{l|}{\modelTopic \textbf{(Ours)}} & \textbf{0.8961*} & {\textbf{0.0998*}} & \multicolumn{1}{c|}{\textbf{0.0320*}} & \textbf{0.6056*} & \textbf{0.1650*} & \multicolumn{1}{c}{\textbf{0.0979*}} \\
\multirow{8}{*}{3} & \multicolumn{1}{l|}{\modelAll \textbf{(Ours)}} & {\ul 0.8664*} & {\ul 0.1062}* & \multicolumn{1}{c|}{{\ul 0.0359*}} & {\ul 0.5420} & {\ul 0.1896*} & \multicolumn{1}{c}{{\ul 0.1217}} \\
 & \multicolumn{1}{l|}{Long-Context} & 0.5320 & 0.1834 & \multicolumn{1}{c|}{0.1395} & 0.2662 & 0.3614 & \multicolumn{1}{c}{0.3173} \\
 & \multicolumn{1}{l|}{RAG-\textit{All}} & 0.6325 & 0.1557 & \multicolumn{1}{c|}{0.0898} & 0.3098 & 0.3499 & \multicolumn{1}{c}{0.2825} \\
 & \multicolumn{1}{l|}{RAG-\textit{Doc}} & 0.6909 & 0.1529 & \multicolumn{1}{c|}{0.0776} & 0.3356 & 0.3476 & \multicolumn{1}{c}{0.2752} \\
 & \multicolumn{1}{l|}{Hierarchical} & 0.7647 & 0.1191 & \multicolumn{1}{c|}{{0.0575}} & 0.3509 & 0.3032 & \multicolumn{1}{c}{0.2523} \\
 & \multicolumn{1}{l|}{Incremental-\textit{All}} & 0.5037 & 0.2466 & \multicolumn{1}{c|}{0.1924} & 0.3467 & 0.3019 & \multicolumn{1}{c}{0.2488} \\
 & \multicolumn{1}{l|}{Incremental-\textit{Topic}} & 0.5635 & 0.2288 & \multicolumn{1}{c|}{0.1720} & 0.4209 & 0.2796 & \multicolumn{1}{c}{0.2236} \\ \bottomrule
\end{tabular}
\caption{\label{table:doc_cover_cqa_fixedtopic}ConflictingQA citation coverage, balance, and accuracy when models have fixed topics (except RAG and RAG+Cluster). Best model is \textbf{bold}, second best is \underline{underlined}. Models with * are significantly the best (2-sample $t$-test, $p<0.05$ with Bonferroni correction. \model consistently has the highest citation coverage, fairness, and faithfulness for summaries and topic paragraphs, even when baselines use the same topics, suggesting that our gains are not derived from the agenda planning step, but rather question tailoring and outline construction. }
\end{table*}

\begin{table*}[t]
\centering
\footnotesize
\setlength{\tabcolsep}{2.75pt}
\renewcommand{\arraystretch}{0.6}
\begin{tabular}{@{}clcccccc@{}}
\multicolumn{1}{l}{} &  & \multicolumn{3}{c}{\textit{Summary Level}} & \multicolumn{3}{c}{\textit{Topic Paragraph Level}} \\ 
\toprule
\textbf{\# Top.} & \multicolumn{1}{l|}{\textbf{Model}} & \textbf{DC ($\uparrow$)} & \textbf{Fair ($\downarrow$)} & \multicolumn{1}{c|}{\textbf{Faithful ($\downarrow$)}} & \textbf{DC ($\uparrow$)} & \textbf{Fair ($\downarrow$)} & \multicolumn{1}{c}{\textbf{Faithful ($\downarrow$)}} \\ \midrule
\multirow{10}{*}{3} & \multicolumn{1}{l|}{\modelTopic \textbf{(Ours)}} & \textbf{0.8724*} & \textbf{0.0701*} & \multicolumn{1}{c|}{\textbf{0.0235*}} & \textbf{0.6066*} & \textbf{0.1255*} & \multicolumn{1}{c}{\textbf{0.0789*}} \\
 & \multicolumn{1}{l|}{\modelAll \textbf{(Ours)}} & {\ul 0.8457*} & {\ul 0.0786*} & \multicolumn{1}{c|}{{\ul 0.0273*}} & {\ul 0.5508} & {\ul 0.1463*} & \multicolumn{1}{c}{{\ul 0.0938*}} \\
 & \multicolumn{1}{l|}{Long-Context} & 0.6025 & 0.1919 & \multicolumn{1}{c|}{0.1559} & 0.2956 & 0.3865 & \multicolumn{1}{c}{0.3517} \\
 & \multicolumn{1}{l|}{RAG-\textit{All}} & 0.6200 & 0.1502 & \multicolumn{1}{c|}{0.0968} & 0.3103 & 0.3421 & \multicolumn{1}{c}{0.2896} \\
 & \multicolumn{1}{l|}{RAG-\textit{Doc}} & 0.6728 & 0.1216 & \multicolumn{1}{c|}{0.0683} & 0.3254 & 0.3226 & \multicolumn{1}{c}{0.2694} \\
 & \multicolumn{1}{l|}{Hierarchical} & 0.7676 & 0.0954 & \multicolumn{1}{c|}{0.0443} & 0.3650 & 0.2729 & \multicolumn{1}{c}{0.2207} \\
 & \multicolumn{1}{l|}{Incremental-\textit{All}} & 0.5566 & 0.2579 & \multicolumn{1}{c|}{0.2089} & 0.3919 & 0.3243 & \multicolumn{1}{c}{0.2765} \\
 & \multicolumn{1}{l|}{Incremental-\textit{Topic}} & 0.6152 & 0.2415 & \multicolumn{1}{c|}{0.1970} & 0.4707 & 0.3128 & \multicolumn{1}{c}{0.2674}\\ \bottomrule
\end{tabular}

\caption{\label{table:doc_cover_debate_fixedtopic}DebateQFS citation coverage, balance, and accuracy when models have fixed topics (except RAG and RAG+Cluster). Best model is \textbf{bold}, second best is \underline{underlined}. Models with * are significantly the best (2-sample $t$-test, $p<0.05$ with Bonferroni correction. \model consistently has the highest citation coverage, fairness, and faithfulness for summaries and topic paragraphs, even when baselines use the same topics, suggesting that our gains are not derived from the agenda planning step, but rather question tailoring and outline construction. }
\vspace{-2ex}
\end{table*}
\begin{table*}[]
\definecolor{myblue}{HTML}{DAE8FC}
\small
\centering
\setlength{\tabcolsep}{3.5pt}
\renewcommand{\arraystretch}{0.8}
\begin{tabular}{@{}clrrrrrrrrrrrrrrrc@{}}
\multicolumn{1}{l}{} &  & \multicolumn{5}{c}{\textit{Summary Quality}} & \multicolumn{5}{c}{\textit{Topic Paragraph Quality}} & \multicolumn{5}{c}{\textit{Topic Quality}} & \multicolumn{1}{l}{\textit{Sep.}} \\ \midrule
\textbf{\textbf{\# Topics}} & \multicolumn{1}{l|}{\textbf{Model}} & \multicolumn{1}{c}{\textbf{Int}} & \multicolumn{1}{c}{\textbf{Coh}} & \multicolumn{1}{c}{\textbf{Rel}} & \multicolumn{1}{l}{\textbf{Cov}} & \multicolumn{1}{l|}{\textbf{Div}} & \multicolumn{1}{c}{\textbf{Int}} & \multicolumn{1}{c}{\textbf{Coh}} & \multicolumn{1}{c}{\textbf{Rel}} & \multicolumn{1}{l}{\textbf{Cov}} & \multicolumn{1}{l|}{\textbf{Div}} & \multicolumn{1}{c}{\textbf{Int}} & \multicolumn{1}{c}{\textbf{Coh}} & \multicolumn{1}{c}{\textbf{Rel}} & \multicolumn{1}{l}{\textbf{Cov}} & \multicolumn{1}{l|}{\textbf{Div}} & \textbf{SB} \\ \midrule
 & \multicolumn{1}{l|}{\textbf{\modelTopic}} & \cellcolor[HTML]{DAE8FC}\textbf{4.22} & \cellcolor[HTML]{DAE8FC}4.24 & \cellcolor[HTML]{DAE8FC}4.59 & \cellcolor[HTML]{DAE8FC}4.46 & \multicolumn{1}{r|}{\cellcolor[HTML]{DAE8FC}\textbf{4.23}} & \cellcolor[HTML]{DAE8FC}\textbf{4.09} & \cellcolor[HTML]{DAE8FC}4.30 & \cellcolor[HTML]{DAE8FC}\textbf{4.70} & \cellcolor[HTML]{DAE8FC}\textbf{4.38} & \multicolumn{1}{r|}{\cellcolor[HTML]{DAE8FC}\textbf{3.93}} & \cellcolor[HTML]{DAE8FC}3.22 & \cellcolor[HTML]{DAE8FC}3.88 & \cellcolor[HTML]{DAE8FC}4.56 & \cellcolor[HTML]{DAE8FC}3.00 & \multicolumn{1}{r|}{\cellcolor[HTML]{DAE8FC}3.48} & 0.52 \\
 & \multicolumn{1}{l|}{\textbf{\modelAll}} & \cellcolor[HTML]{DAE8FC}4.12 & \cellcolor[HTML]{DAE8FC}\textbf{4.27} & \cellcolor[HTML]{DAE8FC}\textbf{4.68} & \cellcolor[HTML]{DAE8FC}\textbf{4.49} & \multicolumn{1}{r|}{\cellcolor[HTML]{DAE8FC}4.14} & \cellcolor[HTML]{DAE8FC}3.99 & \cellcolor[HTML]{DAE8FC}4.31 & \cellcolor[HTML]{DAE8FC}4.64 & \cellcolor[HTML]{DAE8FC}4.29 & \multicolumn{1}{r|}{\cellcolor[HTML]{DAE8FC}3.80} & \cellcolor[HTML]{DAE8FC}\textbf{3.27} & \cellcolor[HTML]{DAE8FC}3.93 & \cellcolor[HTML]{DAE8FC}4.52 & \cellcolor[HTML]{DAE8FC}\textbf{3.19} & \multicolumn{1}{r|}{\cellcolor[HTML]{DAE8FC}\textbf{3.70}} & 0.50 \\
 & \multicolumn{1}{l|}{Long-Context} & 3.96 & \cellcolor[HTML]{DAE8FC}4.18 & \cellcolor[HTML]{DAE8FC}4.55 & 4.31 & \multicolumn{1}{r|}{3.85} & 3.72 & 4.14 & 4.51 & 4.03 & \multicolumn{1}{r|}{3.25} & 3.00 & \cellcolor[HTML]{DAE8FC}3.86 & 4.47 & 2.90 & \multicolumn{1}{r|}{\cellcolor[HTML]{DAE8FC}3.47} & 0.45 \\
 & \multicolumn{1}{l|}{RAG-\textit{All}} & \cellcolor[HTML]{DAE8FC}4.06 & \cellcolor[HTML]{DAE8FC}4.24 & \cellcolor[HTML]{DAE8FC}4.55 & \cellcolor[HTML]{DAE8FC}4.43 & \multicolumn{1}{r|}{4.00} & 3.80 & \cellcolor[HTML]{DAE8FC}4.25 & 4.60 & 4.13 & \multicolumn{1}{r|}{3.63} & \cellcolor[HTML]{DAE8FC}3.08 & \cellcolor[HTML]{DAE8FC}3.86 & \cellcolor[HTML]{DAE8FC}4.51 & 2.81 & \multicolumn{1}{r|}{3.42} & 0.47 \\
 & \multicolumn{1}{l|}{RAG-\textit{Doc}} & \cellcolor[HTML]{DAE8FC}4.17 & \cellcolor[HTML]{DAE8FC}4.22 & \cellcolor[HTML]{DAE8FC}4.56 & \cellcolor[HTML]{DAE8FC}4.39 & \multicolumn{1}{r|}{\cellcolor[HTML]{DAE8FC}4.16} & 3.86 & \cellcolor[HTML]{DAE8FC}4.26 & \cellcolor[HTML]{DAE8FC}4.64 & 4.24 & \multicolumn{1}{r|}{3.71} & \cellcolor[HTML]{DAE8FC}3.10 & \cellcolor[HTML]{DAE8FC}3.88 & \cellcolor[HTML]{DAE8FC}4.59 & 2.84 & \multicolumn{1}{r|}{3.41} & 0.47 \\
 & \multicolumn{1}{l|}{Hierarchical} & \cellcolor[HTML]{DAE8FC}4.16 & \cellcolor[HTML]{DAE8FC}4.24 & \cellcolor[HTML]{DAE8FC}4.58 & \cellcolor[HTML]{DAE8FC}4.46 & \multicolumn{1}{r|}{\cellcolor[HTML]{DAE8FC}4.14} & 3.93 & \cellcolor[HTML]{DAE8FC}\textbf{4.33} & \cellcolor[HTML]{DAE8FC}\textbf{4.70} & 4.27 & \multicolumn{1}{r|}{3.76} & \cellcolor[HTML]{DAE8FC}3.21 & \cellcolor[HTML]{DAE8FC}3.90 & \cellcolor[HTML]{DAE8FC}\textbf{4.61} & \cellcolor[HTML]{DAE8FC}3.18 & \multicolumn{1}{r|}{\cellcolor[HTML]{DAE8FC}3.47} & 0.47 \\
 & \multicolumn{1}{l|}{Increm-\textit{All}} & 3.95 & \cellcolor[HTML]{DAE8FC}4.14 & \cellcolor[HTML]{DAE8FC}4.58 & 4.28 & \multicolumn{1}{r|}{3.90} & 3.64 & 4.11 & 4.57 & 4.01 & \multicolumn{1}{r|}{3.31} & \cellcolor[HTML]{DAE8FC}3.14 & \cellcolor[HTML]{DAE8FC}\textbf{3.97} & \cellcolor[HTML]{DAE8FC}4.60 & \cellcolor[HTML]{DAE8FC}3.07 & \multicolumn{1}{r|}{3.46} & 0.46 \\
 & \multicolumn{1}{l|}{Increm-\textit{Topic}} & \cellcolor[HTML]{DAE8FC}4.11 & \cellcolor[HTML]{DAE8FC}4.21 & \cellcolor[HTML]{DAE8FC}4.60 & \cellcolor[HTML]{DAE8FC}4.44 & \multicolumn{1}{r|}{\cellcolor[HTML]{DAE8FC}4.18} & \cellcolor[HTML]{DAE8FC}4.05 & \cellcolor[HTML]{DAE8FC}4.30 & \cellcolor[HTML]{DAE8FC}4.66 & 4.21 & \multicolumn{1}{r|}{3.76} & \cellcolor[HTML]{DAE8FC}3.03 & 3.63 & 4.37 & 2.83 & \multicolumn{1}{r|}{3.30} & 0.49 \\
 & \multicolumn{1}{l|}{Cluster} & 3.89 & 4.08 & 4.45 & 4.22 & \multicolumn{1}{r|}{3.94} & 3.73 & 4.11 & 4.49 & 4.04 & \multicolumn{1}{r|}{3.50} & 2.41 & 3.16 & 3.89 & 2.29 & \multicolumn{1}{r|}{2.47} & 0.48 \\
\multirow{-10}{*}{2} & \multicolumn{1}{l|}{RAG+Cluster} & \cellcolor[HTML]{DAE8FC}4.13 & \cellcolor[HTML]{DAE8FC}\textbf{4.27} & \cellcolor[HTML]{DAE8FC}4.59 & \cellcolor[HTML]{DAE8FC}4.38 & \multicolumn{1}{r|}{\cellcolor[HTML]{DAE8FC}4.07} & \cellcolor[HTML]{DAE8FC}3.97 & \cellcolor[HTML]{DAE8FC}4.29 & \cellcolor[HTML]{DAE8FC}4.67 & \cellcolor[HTML]{DAE8FC}4.30 & \multicolumn{1}{r|}{\cellcolor[HTML]{DAE8FC}3.87} & 2.53 & 3.26 & 4.04 & 2.49 & \multicolumn{1}{r|}{2.60} & 0.52 \\ \midrule
 & \multicolumn{1}{l|}{\textbf{\modelTopic}} & \cellcolor[HTML]{DAE8FC}4.24 & \cellcolor[HTML]{DAE8FC}4.34 & \cellcolor[HTML]{DAE8FC}4.64 & \cellcolor[HTML]{DAE8FC}4.49 & \multicolumn{1}{r|}{\cellcolor[HTML]{DAE8FC}\textbf{4.42}} & \cellcolor[HTML]{DAE8FC}\textbf{4.08} & \cellcolor[HTML]{DAE8FC}\textbf{4.33} & \cellcolor[HTML]{DAE8FC}4.69 & \cellcolor[HTML]{DAE8FC}\textbf{4.34} & \multicolumn{1}{r|}{\cellcolor[HTML]{DAE8FC}\textbf{3.89}} & \cellcolor[HTML]{DAE8FC}3.47 & \cellcolor[HTML]{DAE8FC}\textbf{4.12} & \cellcolor[HTML]{DAE8FC}\textbf{4.69} & \cellcolor[HTML]{DAE8FC}\textbf{3.61} & \multicolumn{1}{r|}{\cellcolor[HTML]{DAE8FC}\textbf{4.02}} & 0.69 \\
 & \multicolumn{1}{l|}{\textbf{\modelAll}} & \cellcolor[HTML]{DAE8FC}\textbf{4.27} & \cellcolor[HTML]{DAE8FC}4.33 & \cellcolor[HTML]{DAE8FC}4.63 & \cellcolor[HTML]{DAE8FC}4.49 & \multicolumn{1}{r|}{\cellcolor[HTML]{DAE8FC}4.40} & 3.88 & \cellcolor[HTML]{DAE8FC}4.27 & 4.60 & 4.19 & \multicolumn{1}{r|}{3.70} & \cellcolor[HTML]{DAE8FC}\textbf{3.49} & \cellcolor[HTML]{DAE8FC}4.09 & \cellcolor[HTML]{DAE8FC}4.62 & \cellcolor[HTML]{DAE8FC}3.46 & \multicolumn{1}{r|}{\cellcolor[HTML]{DAE8FC}3.99} & 0.65 \\
 & \multicolumn{1}{l|}{Long-Context} & 4.02 & \cellcolor[HTML]{DAE8FC}4.34 & \cellcolor[HTML]{DAE8FC}4.63 & \cellcolor[HTML]{DAE8FC}4.44 & \multicolumn{1}{r|}{4.23} & 3.62 & 4.14 & 4.51 & 3.89 & \multicolumn{1}{r|}{3.21} & 3.24 & \cellcolor[HTML]{DAE8FC}4.03 & \cellcolor[HTML]{DAE8FC}4.55 & 3.25 & \multicolumn{1}{r|}{3.76} & 0.58 \\
 & \multicolumn{1}{l|}{RAG-\textit{All}} & \cellcolor[HTML]{DAE8FC}4.16 & \cellcolor[HTML]{DAE8FC}4.33 & \cellcolor[HTML]{DAE8FC}4.67 & \cellcolor[HTML]{DAE8FC}4.49 & \multicolumn{1}{r|}{\cellcolor[HTML]{DAE8FC}4.29} & 3.80 & 4.16 & 4.61 & 4.06 & \multicolumn{1}{r|}{3.53} & \cellcolor[HTML]{DAE8FC}3.41 & \cellcolor[HTML]{DAE8FC}4.08 & \cellcolor[HTML]{DAE8FC}4.57 & \cellcolor[HTML]{DAE8FC}3.47 & \multicolumn{1}{r|}{\cellcolor[HTML]{DAE8FC}3.95} & 0.60 \\
 & \multicolumn{1}{l|}{RAG-\textit{Doc}} & \cellcolor[HTML]{DAE8FC}4.15 & \cellcolor[HTML]{DAE8FC}\textbf{4.37} & \cellcolor[HTML]{DAE8FC}4.68 & \cellcolor[HTML]{DAE8FC}4.47 & \multicolumn{1}{r|}{\cellcolor[HTML]{DAE8FC}\textbf{4.42}} & 3.76 & 4.22 & 4.60 & 4.10 & \multicolumn{1}{r|}{3.56} & \cellcolor[HTML]{DAE8FC}3.33 & \cellcolor[HTML]{DAE8FC}4.08 & \cellcolor[HTML]{DAE8FC}4.63 & \cellcolor[HTML]{DAE8FC}3.39 & \multicolumn{1}{r|}{\cellcolor[HTML]{DAE8FC}3.91} & 0.60 \\
 & \multicolumn{1}{l|}{Hierarchical} & 4.24 & \cellcolor[HTML]{DAE8FC}\textbf{4.37} & \cellcolor[HTML]{DAE8FC}4.73 & \cellcolor[HTML]{DAE8FC}4.50 & \multicolumn{1}{r|}{\cellcolor[HTML]{DAE8FC}4.38} & 3.78 & 4.21 & 4.62 & 4.14 & \multicolumn{1}{r|}{3.57} & \cellcolor[HTML]{DAE8FC}3.43 & \cellcolor[HTML]{DAE8FC}4.07 & \cellcolor[HTML]{DAE8FC}4.65 & \cellcolor[HTML]{DAE8FC}3.49 & \multicolumn{1}{r|}{\cellcolor[HTML]{DAE8FC}3.94} & 0.58 \\
 & \multicolumn{1}{l|}{Increm-\textit{All}} & 3.98 & \cellcolor[HTML]{DAE8FC}4.29 & \cellcolor[HTML]{DAE8FC}4.67 & 4.42 & \multicolumn{1}{r|}{4.21} & 3.54 & 4.09 & 4.56 & 3.79 & \multicolumn{1}{r|}{3.26} & \cellcolor[HTML]{DAE8FC}3.44 & \cellcolor[HTML]{DAE8FC}4.02 & \cellcolor[HTML]{DAE8FC}4.65 & \cellcolor[HTML]{DAE8FC}3.52 & \multicolumn{1}{r|}{\cellcolor[HTML]{DAE8FC}3.94} & 0.58 \\
 & \multicolumn{1}{l|}{Increm-\textit{Topic}} & 4.17 & \cellcolor[HTML]{DAE8FC}\textbf{4.37} & \cellcolor[HTML]{DAE8FC}\textbf{4.74} & \cellcolor[HTML]{DAE8FC}\textbf{4.57} & \multicolumn{1}{r|}{\cellcolor[HTML]{DAE8FC}4.39} & 3.91 & \cellcolor[HTML]{DAE8FC}4.29 & 4.62 & 4.25 & \multicolumn{1}{r|}{3.65} & \cellcolor[HTML]{DAE8FC}3.36 & 3.79 & 4.31 & 3.21 & \multicolumn{1}{r|}{3.73} & 0.61 \\
 & \multicolumn{1}{l|}{Cluster} & 3.81 & 4.03 & 4.25 & 4.19 & \multicolumn{1}{r|}{3.94} & 3.69 & 4.08 & 4.45 & 3.95 & \multicolumn{1}{r|}{3.53} & 2.42 & 2.86 & 3.73 & 2.13 & \multicolumn{1}{r|}{2.47} & 0.61 \\
\multirow{-10}{*}{3} & \multicolumn{1}{l|}{RAG+Cluster} & \cellcolor[HTML]{DAE8FC}4.14 & 4.22 & 4.60 & \cellcolor[HTML]{DAE8FC}4.52 & \multicolumn{1}{r|}{4.22} & 3.96 & \cellcolor[HTML]{DAE8FC}4.31 & \cellcolor[HTML]{DAE8FC}\textbf{4.71} & 4.25 & \multicolumn{1}{r|}{3.77} & 2.43 & 3.11 & 3.82 & 2.44 & \multicolumn{1}{r|}{2.64} & 0.64 \\ \midrule
 & \multicolumn{1}{l|}{\textbf{\modelTopic}} & \cellcolor[HTML]{DAE8FC}\textbf{4.30} & \cellcolor[HTML]{DAE8FC}4.21 & \cellcolor[HTML]{DAE8FC}4.54 & \cellcolor[HTML]{DAE8FC}\textbf{4.54} & \multicolumn{1}{r|}{\cellcolor[HTML]{DAE8FC}\textbf{4.48}} & \cellcolor[HTML]{DAE8FC}\textbf{4.09} & \cellcolor[HTML]{DAE8FC}\textbf{4.29} & \cellcolor[HTML]{DAE8FC}\textbf{4.66} & \cellcolor[HTML]{DAE8FC}\textbf{4.35} & \multicolumn{1}{r|}{\cellcolor[HTML]{DAE8FC}\textbf{3.89}} & \cellcolor[HTML]{DAE8FC}\textbf{3.93} & \cellcolor[HTML]{DAE8FC}4.17 & \cellcolor[HTML]{DAE8FC}4.65 & \cellcolor[HTML]{DAE8FC}4.04 & \multicolumn{1}{r|}{\cellcolor[HTML]{DAE8FC}\textbf{4.31}} & 0.72 \\
 & \multicolumn{1}{l|}{\textbf{\modelAll}} & \cellcolor[HTML]{DAE8FC}4.24 & \cellcolor[HTML]{DAE8FC}4.26 & \cellcolor[HTML]{DAE8FC}4.53 & \cellcolor[HTML]{DAE8FC}4.49 & \multicolumn{1}{r|}{\cellcolor[HTML]{DAE8FC}4.38} & 3.93 & \cellcolor[HTML]{DAE8FC}4.24 & \cellcolor[HTML]{DAE8FC}4.61 & 4.20 & \multicolumn{1}{r|}{3.76} & \cellcolor[HTML]{DAE8FC}3.80 & \cellcolor[HTML]{DAE8FC}\textbf{4.22} & \cellcolor[HTML]{DAE8FC}\textbf{4.67} & \cellcolor[HTML]{DAE8FC}\textbf{4.13} & \multicolumn{1}{r|}{\cellcolor[HTML]{DAE8FC}4.30} & 0.70 \\
 & \multicolumn{1}{l|}{Long-Context} & \cellcolor[HTML]{DAE8FC}4.14 & \cellcolor[HTML]{DAE8FC}4.26 & \cellcolor[HTML]{DAE8FC}4.48 & 4.32 & \multicolumn{1}{r|}{4.25} & 3.53 & 4.08 & 4.47 & 3.83 & \multicolumn{1}{r|}{3.14} & 3.65 & 4.00 & 4.53 & 3.85 & \multicolumn{1}{r|}{4.04} & 0.65 \\
 & \multicolumn{1}{l|}{RAG-\textit{All}} & \cellcolor[HTML]{DAE8FC}4.17 & \cellcolor[HTML]{DAE8FC}4.30 & \cellcolor[HTML]{DAE8FC}4.55 & \cellcolor[HTML]{DAE8FC}4.45 & \multicolumn{1}{r|}{4.29} & 3.72 & 4.18 & 4.59 & 3.99 & \multicolumn{1}{r|}{3.44} & \cellcolor[HTML]{DAE8FC}3.80 & \cellcolor[HTML]{DAE8FC}4.14 & \cellcolor[HTML]{DAE8FC}4.65 & \cellcolor[HTML]{DAE8FC}4.03 & \multicolumn{1}{r|}{\cellcolor[HTML]{DAE8FC}4.23} & 0.66 \\
 & \multicolumn{1}{l|}{RAG-\textit{Doc}} & \cellcolor[HTML]{DAE8FC}4.23 & \cellcolor[HTML]{DAE8FC}\textbf{4.31} & \cellcolor[HTML]{DAE8FC}4.56 & \cellcolor[HTML]{DAE8FC}4.41 & \multicolumn{1}{r|}{\cellcolor[HTML]{DAE8FC}4.41} & 3.76 & 4.16 & 4.59 & 4.07 & \multicolumn{1}{r|}{3.45} & 3.66 & \cellcolor[HTML]{DAE8FC}4.15 & \cellcolor[HTML]{DAE8FC}4.59 & \cellcolor[HTML]{DAE8FC}4.03 & \multicolumn{1}{r|}{\cellcolor[HTML]{DAE8FC}4.19} & 0.66 \\
 & \multicolumn{1}{l|}{Hierarchical} & \cellcolor[HTML]{DAE8FC}4.23 & \cellcolor[HTML]{DAE8FC}4.24 & \cellcolor[HTML]{DAE8FC}\textbf{4.59} & \cellcolor[HTML]{DAE8FC}4.51 & \multicolumn{1}{r|}{\cellcolor[HTML]{DAE8FC}4.36} & 3.75 & 4.19 & 4.59 & 4.06 & \multicolumn{1}{r|}{3.47} & 3.67 & \cellcolor[HTML]{DAE8FC}4.10 & \cellcolor[HTML]{DAE8FC}4.62 & 3.90 & \multicolumn{1}{r|}{\cellcolor[HTML]{DAE8FC}4.22} & 0.65 \\
 & \multicolumn{1}{l|}{Increm-\textit{All}} & 3.95 & 4.14 & 4.44 & 4.30 & \multicolumn{1}{r|}{4.15} & 3.48 & 4.04 & 4.49 & 3.76 & \multicolumn{1}{r|}{3.14} & 3.71 & \cellcolor[HTML]{DAE8FC}4.09 & \cellcolor[HTML]{DAE8FC}4.62 & \cellcolor[HTML]{DAE8FC}4.02 & \multicolumn{1}{r|}{\cellcolor[HTML]{DAE8FC}4.16} & 0.65 \\
 & \multicolumn{1}{l|}{Increm-\textit{Topic}} & \cellcolor[HTML]{DAE8FC}4.20 & \cellcolor[HTML]{DAE8FC}4.23 & 4.43 & \cellcolor[HTML]{DAE8FC}4.42 & \multicolumn{1}{r|}{\cellcolor[HTML]{DAE8FC}4.38} & 3.93 & \cellcolor[HTML]{DAE8FC}4.25 & \cellcolor[HTML]{DAE8FC}\textbf{4.66} & 4.17 & \multicolumn{1}{r|}{3.60} & 3.47 & 3.85 & 4.39 & 3.69 & \multicolumn{1}{r|}{3.83} & 0.69 \\
 & \multicolumn{1}{l|}{Cluster} & 3.92 & 4.03 & 4.17 & 4.20 & \multicolumn{1}{r|}{4.09} & 3.68 & 4.13 & 4.47 & 3.99 & \multicolumn{1}{r|}{3.51} & 2.36 & 2.73 & 3.62 & 2.27 & \multicolumn{1}{r|}{2.54} & 0.68 \\
\multirow{-10}{*}{4} & \multicolumn{1}{l|}{RAG+Cluster} & \cellcolor[HTML]{DAE8FC}4.21 & \cellcolor[HTML]{DAE8FC}4.20 & 4.44 & \cellcolor[HTML]{DAE8FC}4.44 & \multicolumn{1}{r|}{4.28} & 3.99 & \cellcolor[HTML]{DAE8FC}4.28 & \cellcolor[HTML]{DAE8FC}\textbf{4.66} & 4.26 & \multicolumn{1}{r|}{\cellcolor[HTML]{DAE8FC}3.83} & 2.56 & 3.05 & 3.95 & 2.58 & \multicolumn{1}{r|}{2.69} & 0.71 \\ \midrule
 & \multicolumn{1}{l|}{\textbf{\modelTopic}} & \cellcolor[HTML]{DAE8FC}4.17 & \cellcolor[HTML]{DAE8FC}4.24 & \cellcolor[HTML]{DAE8FC}4.35 & \cellcolor[HTML]{DAE8FC}\textbf{4.51} & \multicolumn{1}{r|}{\cellcolor[HTML]{DAE8FC}4.35} & \cellcolor[HTML]{DAE8FC}\textbf{4.08} & \cellcolor[HTML]{DAE8FC}\textbf{4.33} & \cellcolor[HTML]{DAE8FC}\textbf{4.69} & \cellcolor[HTML]{DAE8FC}\textbf{4.40} & \multicolumn{1}{r|}{\cellcolor[HTML]{DAE8FC}\textbf{3.97}} & \cellcolor[HTML]{DAE8FC}\textbf{4.15} & \cellcolor[HTML]{DAE8FC}\textbf{4.43} & \cellcolor[HTML]{DAE8FC}\textbf{4.82} & \cellcolor[HTML]{DAE8FC}\textbf{4.44} & \multicolumn{1}{r|}{\cellcolor[HTML]{DAE8FC}4.52} & 0.76 \\
 & \multicolumn{1}{l|}{\textbf{\modelAll}} & \cellcolor[HTML]{DAE8FC}\textbf{4.25} & \cellcolor[HTML]{DAE8FC}4.22 & \cellcolor[HTML]{DAE8FC}4.41 & \cellcolor[HTML]{DAE8FC}4.44 & \multicolumn{1}{r|}{\cellcolor[HTML]{DAE8FC}4.39} & 3.89 & 4.24 & 4.60 & 4.21 & \multicolumn{1}{r|}{3.69} & \cellcolor[HTML]{DAE8FC}4.14 & \cellcolor[HTML]{DAE8FC}4.37 & \cellcolor[HTML]{DAE8FC}4.77 & \cellcolor[HTML]{DAE8FC}\textbf{4.44} & \multicolumn{1}{r|}{\cellcolor[HTML]{DAE8FC}4.50} & 0.74 \\
 & \multicolumn{1}{l|}{Long-Context} & 3.98 & 4.11 & 4.28 & 4.29 & \multicolumn{1}{r|}{4.12} & 3.50 & 4.10 & 4.46 & 3.83 & \multicolumn{1}{r|}{3.02} & 3.90 & \cellcolor[HTML]{DAE8FC}4.35 & \cellcolor[HTML]{DAE8FC}4.71 & 4.22 & \multicolumn{1}{r|}{4.37} & 0.69 \\
 & \multicolumn{1}{l|}{RAG-\textit{All}} & \cellcolor[HTML]{DAE8FC}4.11 & \cellcolor[HTML]{DAE8FC}4.24 & \cellcolor[HTML]{DAE8FC}\textbf{4.48} & \cellcolor[HTML]{DAE8FC}4.48 & \multicolumn{1}{r|}{4.28} & 3.69 & 4.18 & 4.56 & 3.99 & \multicolumn{1}{r|}{3.39} & \cellcolor[HTML]{DAE8FC}4.02 & \cellcolor[HTML]{DAE8FC}4.39 & \cellcolor[HTML]{DAE8FC}4.80 & \cellcolor[HTML]{DAE8FC}4.36 & \multicolumn{1}{r|}{\cellcolor[HTML]{DAE8FC}4.46} & 0.71 \\
 & \multicolumn{1}{l|}{RAG-\textit{Doc}} & \cellcolor[HTML]{DAE8FC}4.12 & \cellcolor[HTML]{DAE8FC}4.20 & \cellcolor[HTML]{DAE8FC}\textbf{4.48} & \cellcolor[HTML]{DAE8FC}4.42 & \multicolumn{1}{r|}{\cellcolor[HTML]{DAE8FC}\textbf{4.50}} & 3.74 & 4.21 & 4.57 & 4.01 & \multicolumn{1}{r|}{3.42} & \cellcolor[HTML]{DAE8FC}3.96 & \cellcolor[HTML]{DAE8FC}4.36 & \cellcolor[HTML]{DAE8FC}4.78 & \cellcolor[HTML]{DAE8FC}4.32 & \multicolumn{1}{r|}{\cellcolor[HTML]{DAE8FC}4.41} & 0.70 \\
 & \multicolumn{1}{l|}{Hierarchical} & \cellcolor[HTML]{DAE8FC}4.07 & \cellcolor[HTML]{DAE8FC}\textbf{4.27} & 4.47 & \cellcolor[HTML]{DAE8FC}4.42 & \multicolumn{1}{r|}{\cellcolor[HTML]{DAE8FC}4.41} & 3.69 & 4.17 & 4.55 & 4.01 & \multicolumn{1}{r|}{3.39} & \cellcolor[HTML]{DAE8FC}4.07 & \cellcolor[HTML]{DAE8FC}4.35 & \cellcolor[HTML]{DAE8FC}4.80 & \cellcolor[HTML]{DAE8FC}4.37 & \multicolumn{1}{r|}{\cellcolor[HTML]{DAE8FC}\textbf{4.56}} & 0.70 \\
 & \multicolumn{1}{l|}{Increm-\textit{All}} & 3.83 & 4.09 & \cellcolor[HTML]{DAE8FC}4.35 & 4.27 & \multicolumn{1}{r|}{4.05} & 3.38 & 4.00 & 4.42 & 3.66 & \multicolumn{1}{r|}{2.98} & \cellcolor[HTML]{DAE8FC}4.06 & \cellcolor[HTML]{DAE8FC}4.41 & \cellcolor[HTML]{DAE8FC}4.74 & \cellcolor[HTML]{DAE8FC}4.29 & \multicolumn{1}{r|}{\cellcolor[HTML]{DAE8FC}4.44} & 0.69 \\
 & \multicolumn{1}{l|}{Increm-\textit{Topic}} & \cellcolor[HTML]{DAE8FC}4.05 & \cellcolor[HTML]{DAE8FC}4.22 & \cellcolor[HTML]{DAE8FC}4.34 & \cellcolor[HTML]{DAE8FC}4.34 & \multicolumn{1}{r|}{4.25} & 3.86 & 4.24 & 4.64 & 4.14 & \multicolumn{1}{r|}{3.57} & 3.69 & 4.00 & 4.52 & 3.96 & \multicolumn{1}{r|}{4.11} & 0.73 \\
 & \multicolumn{1}{l|}{Cluster} & 3.92 & 3.88 & 3.94 & 4.10 & \multicolumn{1}{r|}{4.07} & 3.74 & 4.09 & 4.46 & 4.00 & \multicolumn{1}{r|}{3.50} & 2.27 & 2.68 & 3.55 & 2.41 & \multicolumn{1}{r|}{2.48} & 0.73 \\
\multirow{-10}{*}{5} & \multicolumn{1}{l|}{RAG+Cluster} & 4.00 & 4.08 & 4.30 & 4.28 & \multicolumn{1}{r|}{4.29} & \cellcolor[HTML]{DAE8FC}4.00 & \cellcolor[HTML]{DAE8FC}4.30 & \cellcolor[HTML]{DAE8FC}4.66 & 4.28 & \multicolumn{1}{r|}{3.77} & 2.63 & 3.01 & 3.88 & 2.66 & \multicolumn{1}{r|}{2.62} & 0.75 \\ \bottomrule
\end{tabular}
\caption{\label{appendix:table:llm_cqa} Interest, Coherence, Relevance, Coverage, and Diversity scores from Prometheus for summaries, topic paragraphs, and topics on ConflictingQA. Best scores are \textbf{bold}, significant scores in \colorbox{myblue}{blue} (2-sample $t$-test, $p<0.05$)}
\end{table*}
\begin{table*}[]
\definecolor{myblue}{HTML}{DAE8FC}
\small
\centering
\setlength{\tabcolsep}{3.5pt}
\renewcommand{\arraystretch}{0.8}
\begin{tabular}{@{}clrrrrrrrrrrrrrrrc@{}}
\multicolumn{1}{l}{} &  & \multicolumn{5}{c}{\textit{Summary Quality}} & \multicolumn{5}{c}{\textit{Topic Paragraph Quality}} & \multicolumn{5}{c}{\textit{Topic Quality}} & \multicolumn{1}{l}{\textit{Sep.}} \\ \midrule
\textbf{\# Topics} & \multicolumn{1}{l|}{\textbf{Model}} & \multicolumn{1}{c}{\textbf{Int}} & \multicolumn{1}{c}{\textbf{Coh}} & \multicolumn{1}{c}{\textbf{Rel}} & \multicolumn{1}{l}{\textbf{Cov}} & \multicolumn{1}{l|}{\textbf{Div}} & \multicolumn{1}{c}{\textbf{Int}} & \multicolumn{1}{c}{\textbf{Coh}} & \multicolumn{1}{c}{\textbf{Rel}} & \multicolumn{1}{l}{\textbf{Cov}} & \multicolumn{1}{l|}{\textbf{Div}} & \multicolumn{1}{c}{\textbf{Int}} & \multicolumn{1}{c}{\textbf{Coh}} & \multicolumn{1}{c}{\textbf{Rel}} & \multicolumn{1}{l}{\textbf{Cov}} & \multicolumn{1}{l|}{\textbf{Div}} & \textbf{SB} \\ \midrule
 & \multicolumn{1}{l|}{\textbf{\modelTopic}} & \cellcolor[HTML]{DAE8FC}\textbf{4.16} & \cellcolor[HTML]{DAE8FC}4.13 & \cellcolor[HTML]{DAE8FC}4.53 & \cellcolor[HTML]{DAE8FC}\textbf{4.34} & \multicolumn{1}{r|}{\cellcolor[HTML]{DAE8FC}\textbf{4.15}} & \cellcolor[HTML]{DAE8FC}\textbf{4.03} & \cellcolor[HTML]{DAE8FC}4.20 & \cellcolor[HTML]{DAE8FC}\textbf{4.62} & \cellcolor[HTML]{DAE8FC}\textbf{4.22} & \multicolumn{1}{r|}{\cellcolor[HTML]{DAE8FC}\textbf{3.89}} & \cellcolor[HTML]{DAE8FC}\textbf{3.28} & \cellcolor[HTML]{DAE8FC}3.98 & \cellcolor[HTML]{DAE8FC}4.62 & \cellcolor[HTML]{DAE8FC}2.93 & \multicolumn{1}{r|}{\cellcolor[HTML]{DAE8FC}3.56} & 0.50 \\
 & \multicolumn{1}{l|}{\textbf{\modelAll}} & \cellcolor[HTML]{DAE8FC}3.98 & \cellcolor[HTML]{DAE8FC}4.10 & \cellcolor[HTML]{DAE8FC}4.45 & \cellcolor[HTML]{DAE8FC}\textbf{4.34} & \multicolumn{1}{r|}{\cellcolor[HTML]{DAE8FC}4.09} & \cellcolor[HTML]{DAE8FC}3.88 & \cellcolor[HTML]{DAE8FC}4.20 & 4.50 & \cellcolor[HTML]{DAE8FC}4.16 & \multicolumn{1}{r|}{\cellcolor[HTML]{DAE8FC}3.75} & \cellcolor[HTML]{DAE8FC}3.23 & \cellcolor[HTML]{DAE8FC}\textbf{4.01} & \cellcolor[HTML]{DAE8FC}4.61 & \cellcolor[HTML]{DAE8FC}\textbf{3.11} & \multicolumn{1}{r|}{\cellcolor[HTML]{DAE8FC}3.56} & 0.49 \\
 & \multicolumn{1}{l|}{Long-Context} & 3.79 & \cellcolor[HTML]{DAE8FC}4.07 & \cellcolor[HTML]{DAE8FC}4.48 & \cellcolor[HTML]{DAE8FC}4.19 & \multicolumn{1}{r|}{3.83} & 3.57 & \cellcolor[HTML]{DAE8FC}4.15 & \cellcolor[HTML]{DAE8FC}4.53 & 3.90 & \multicolumn{1}{r|}{3.28} & \cellcolor[HTML]{DAE8FC}3.15 & \cellcolor[HTML]{DAE8FC}3.81 & \cellcolor[HTML]{DAE8FC}4.56 & 2.70 & \multicolumn{1}{r|}{\cellcolor[HTML]{DAE8FC}3.51} & 0.46 \\
 & \multicolumn{1}{l|}{RAG-\textit{All}} & \cellcolor[HTML]{DAE8FC}4.02 & \cellcolor[HTML]{DAE8FC}4.08 & \cellcolor[HTML]{DAE8FC}4.46 & \cellcolor[HTML]{DAE8FC}4.20 & \multicolumn{1}{r|}{\cellcolor[HTML]{DAE8FC}3.96} & 3.72 & 4.11 & \cellcolor[HTML]{DAE8FC}4.54 & 4.04 & \multicolumn{1}{r|}{3.61} & \cellcolor[HTML]{DAE8FC}3.23 & \cellcolor[HTML]{DAE8FC}3.78 & \cellcolor[HTML]{DAE8FC}4.56 & \cellcolor[HTML]{DAE8FC}2.97 & \multicolumn{1}{r|}{\cellcolor[HTML]{DAE8FC}3.46} & 0.46 \\
 & \multicolumn{1}{l|}{RAG-\textit{Doc}} & 3.90 & \cellcolor[HTML]{DAE8FC}\textbf{4.18} & \cellcolor[HTML]{DAE8FC}4.54 & \cellcolor[HTML]{DAE8FC}4.28 & \multicolumn{1}{r|}{3.86} & 3.74 & 4.10 & \cellcolor[HTML]{DAE8FC}4.52 & 4.03 & \multicolumn{1}{r|}{3.60} & \cellcolor[HTML]{DAE8FC}3.08 & \cellcolor[HTML]{DAE8FC}3.97 & \cellcolor[HTML]{DAE8FC}4.63 & \cellcolor[HTML]{DAE8FC}2.95 & \multicolumn{1}{r|}{\cellcolor[HTML]{DAE8FC}3.55} & 0.47 \\
 & \multicolumn{1}{l|}{Hierarchical} & 4.08 & \cellcolor[HTML]{DAE8FC}4.16 & \cellcolor[HTML]{DAE8FC}\textbf{4.55} & \cellcolor[HTML]{DAE8FC}4.28 & \multicolumn{1}{r|}{\cellcolor[HTML]{DAE8FC}4.05} & \cellcolor[HTML]{DAE8FC}3.94 & \cellcolor[HTML]{DAE8FC}4.21 & \cellcolor[HTML]{DAE8FC}4.56 & 4.07 & \multicolumn{1}{r|}{3.62} & \cellcolor[HTML]{DAE8FC}3.13 & \cellcolor[HTML]{DAE8FC}3.90 & \cellcolor[HTML]{DAE8FC}\textbf{4.64} & \cellcolor[HTML]{DAE8FC}3.06 & \multicolumn{1}{r|}{\cellcolor[HTML]{DAE8FC}\textbf{3.60}} & 0.47 \\
 & \multicolumn{1}{l|}{Increm-\textit{All}} & 3.81 & \cellcolor[HTML]{DAE8FC}4.04 & \cellcolor[HTML]{DAE8FC}4.50 & \cellcolor[HTML]{DAE8FC}4.25 & \multicolumn{1}{r|}{\cellcolor[HTML]{DAE8FC}3.93} & 3.65 & 4.08 & \cellcolor[HTML]{DAE8FC}4.51 & 3.79 & \multicolumn{1}{r|}{3.40} & \cellcolor[HTML]{DAE8FC}3.15 & \cellcolor[HTML]{DAE8FC}3.99 & \cellcolor[HTML]{DAE8FC}4.62 & \cellcolor[HTML]{DAE8FC}2.92 & \multicolumn{1}{r|}{\cellcolor[HTML]{DAE8FC}3.58} & 0.45 \\
 & \multicolumn{1}{l|}{Increm-\textit{Topic}} & \cellcolor[HTML]{DAE8FC}3.91 & \cellcolor[HTML]{DAE8FC}4.18 & \cellcolor[HTML]{DAE8FC}4.54 & \cellcolor[HTML]{DAE8FC}4.19 & \multicolumn{1}{r|}{\cellcolor[HTML]{DAE8FC}4.12} & \cellcolor[HTML]{DAE8FC}3.92 & \cellcolor[HTML]{DAE8FC}\textbf{4.25} & \cellcolor[HTML]{DAE8FC}4.57 & \cellcolor[HTML]{DAE8FC}4.14 & \multicolumn{1}{r|}{3.70} & 2.86 & 3.59 & 4.19 & 2.64 & \multicolumn{1}{r|}{3.09} & 0.48 \\
 & \multicolumn{1}{l|}{Cluster} & \cellcolor[HTML]{DAE8FC}3.91 & 4.01 & 4.35 & 4.09 & \multicolumn{1}{r|}{3.87} & 3.75 & 4.01 & 4.36 & 3.90 & \multicolumn{1}{r|}{3.50} & 2.72 & 3.40 & 4.03 & 2.38 & \multicolumn{1}{r|}{3.02} & 0.45 \\
\multirow{-10}{*}{2} & \multicolumn{1}{l|}{RAG+Cluster} & \cellcolor[HTML]{DAE8FC}3.98 & \cellcolor[HTML]{DAE8FC}4.11 & \cellcolor[HTML]{DAE8FC}4.44 & \cellcolor[HTML]{DAE8FC}4.27 & \multicolumn{1}{r|}{\cellcolor[HTML]{DAE8FC}3.99} & 3.72 & \cellcolor[HTML]{DAE8FC}4.17 & \cellcolor[HTML]{DAE8FC}4.56 & 4.04 & \multicolumn{1}{r|}{3.62} & 2.96 & 3.73 & \cellcolor[HTML]{DAE8FC}4.56 & 2.57 & \multicolumn{1}{r|}{3.26} & 0.48 \\ \midrule
 & \multicolumn{1}{l|}{\textbf{\modelTopic}} & \cellcolor[HTML]{DAE8FC}4.02 & \cellcolor[HTML]{DAE8FC}4.20 & 4.49 & \cellcolor[HTML]{DAE8FC}\textbf{4.44} & \multicolumn{1}{r|}{\cellcolor[HTML]{DAE8FC}4.34} & \cellcolor[HTML]{DAE8FC}\textbf{3.97} & \cellcolor[HTML]{DAE8FC}\textbf{4.21} & \cellcolor[HTML]{DAE8FC}\textbf{4.55} & \cellcolor[HTML]{DAE8FC}\textbf{4.14} & \multicolumn{1}{r|}{\cellcolor[HTML]{DAE8FC}\textbf{3.82}} & \cellcolor[HTML]{DAE8FC}3.54 & \cellcolor[HTML]{DAE8FC}4.09 & \cellcolor[HTML]{DAE8FC}4.64 & 3.39 & \multicolumn{1}{r|}{\cellcolor[HTML]{DAE8FC}3.93} & 0.67 \\
 & \multicolumn{1}{l|}{\textbf{\modelAll}} & \cellcolor[HTML]{DAE8FC}4.11 & \cellcolor[HTML]{DAE8FC}4.21 & \cellcolor[HTML]{DAE8FC}4.60 & \cellcolor[HTML]{DAE8FC}4.34 & \multicolumn{1}{r|}{\cellcolor[HTML]{DAE8FC}\textbf{4.36}} & \cellcolor[HTML]{DAE8FC}3.83 & \cellcolor[HTML]{DAE8FC}4.15 & \cellcolor[HTML]{DAE8FC}4.51 & \cellcolor[HTML]{DAE8FC}4.10 & \multicolumn{1}{r|}{3.63} & \cellcolor[HTML]{DAE8FC}\textbf{3.61} & \cellcolor[HTML]{DAE8FC}4.11 & \cellcolor[HTML]{DAE8FC}4.67 & \cellcolor[HTML]{DAE8FC}\textbf{3.71} & \multicolumn{1}{r|}{\cellcolor[HTML]{DAE8FC}4.02} & 0.64 \\
 & \multicolumn{1}{l|}{Long-Context} & \cellcolor[HTML]{DAE8FC}3.94 & \cellcolor[HTML]{DAE8FC}4.13 & \cellcolor[HTML]{DAE8FC}4.54 & 4.32 & \multicolumn{1}{r|}{4.14} & 3.54 & 4.09 & \cellcolor[HTML]{DAE8FC}4.46 & 3.80 & \multicolumn{1}{r|}{3.17} & \cellcolor[HTML]{DAE8FC}3.36 & \cellcolor[HTML]{DAE8FC}4.09 & \cellcolor[HTML]{DAE8FC}4.69 & 3.36 & \multicolumn{1}{r|}{\cellcolor[HTML]{DAE8FC}4.04} & 0.59 \\
 & \multicolumn{1}{l|}{RAG-\textit{All}} & \cellcolor[HTML]{DAE8FC}4.04 & \cellcolor[HTML]{DAE8FC}4.20 & \cellcolor[HTML]{DAE8FC}4.59 & \cellcolor[HTML]{DAE8FC}4.25 & \multicolumn{1}{r|}{4.14} & 3.62 & 4.06 & \cellcolor[HTML]{DAE8FC}4.49 & 3.87 & \multicolumn{1}{r|}{3.47} & \cellcolor[HTML]{DAE8FC}3.56 & \cellcolor[HTML]{DAE8FC}4.11 & \cellcolor[HTML]{DAE8FC}4.64 & 3.46 & \multicolumn{1}{r|}{\cellcolor[HTML]{DAE8FC}3.97} & 0.59 \\
 & \multicolumn{1}{l|}{RAG-\textit{Doc}} & \cellcolor[HTML]{DAE8FC}4.19 & \cellcolor[HTML]{DAE8FC}\textbf{4.25} & \cellcolor[HTML]{DAE8FC}4.59 & \cellcolor[HTML]{DAE8FC}4.33 & \multicolumn{1}{r|}{4.08} & 3.59 & 4.06 & \cellcolor[HTML]{DAE8FC}4.49 & 3.88 & \multicolumn{1}{r|}{3.36} & \cellcolor[HTML]{DAE8FC}3.56 & \cellcolor[HTML]{DAE8FC}4.10 & \cellcolor[HTML]{DAE8FC}4.62 & \cellcolor[HTML]{DAE8FC}3.51 & \multicolumn{1}{r|}{\cellcolor[HTML]{DAE8FC}3.97} & 0.59 \\
 & \multicolumn{1}{l|}{Hierarchical} & \cellcolor[HTML]{DAE8FC}4.15 & \cellcolor[HTML]{DAE8FC}4.17 & \cellcolor[HTML]{DAE8FC}\textbf{4.69} & \cellcolor[HTML]{DAE8FC}4.35 & \multicolumn{1}{r|}{\cellcolor[HTML]{DAE8FC}4.33} & 3.74 & 4.09 & \cellcolor[HTML]{DAE8FC}4.53 & 3.96 & \multicolumn{1}{r|}{3.48} & \cellcolor[HTML]{DAE8FC}3.56 & \cellcolor[HTML]{DAE8FC}\textbf{4.22} & \cellcolor[HTML]{DAE8FC}\textbf{4.70} & \cellcolor[HTML]{DAE8FC}3.63 & \multicolumn{1}{r|}{\cellcolor[HTML]{DAE8FC}\textbf{4.16}} & 0.58 \\
 & \multicolumn{1}{l|}{Increm-\textit{All}} & 3.92 & 4.08 & 4.52 & \cellcolor[HTML]{DAE8FC}4.29 & \multicolumn{1}{r|}{4.08} & 3.50 & 3.98 & \cellcolor[HTML]{DAE8FC}4.46 & 3.75 & \multicolumn{1}{r|}{3.25} & \cellcolor[HTML]{DAE8FC}3.36 & \cellcolor[HTML]{DAE8FC}4.12 & \cellcolor[HTML]{DAE8FC}4.61 & 3.25 & \multicolumn{1}{r|}{3.75} & 0.58 \\
 & \multicolumn{1}{l|}{Increm-\textit{Topic}} & \cellcolor[HTML]{DAE8FC}\textbf{4.25} & \cellcolor[HTML]{DAE8FC}4.19 & \cellcolor[HTML]{DAE8FC}4.61 & \cellcolor[HTML]{DAE8FC}4.41 & \multicolumn{1}{r|}{\cellcolor[HTML]{DAE8FC}4.23} & \cellcolor[HTML]{DAE8FC}3.91 & \cellcolor[HTML]{DAE8FC}4.17 & \cellcolor[HTML]{DAE8FC}\textbf{4.55} & \cellcolor[HTML]{DAE8FC}4.06 & \multicolumn{1}{r|}{\cellcolor[HTML]{DAE8FC}3.68} & 3.09 & 3.66 & 4.30 & 3.03 & \multicolumn{1}{r|}{3.56} & 0.60 \\
 & \multicolumn{1}{l|}{Cluster} & 3.92 & 3.97 & 4.34 & 4.08 & \multicolumn{1}{r|}{4.06} & 3.64 & 3.95 & 4.31 & 3.82 & \multicolumn{1}{r|}{3.39} & 2.67 & 3.41 & 3.97 & 2.53 & \multicolumn{1}{r|}{3.16} & 0.59 \\
\multirow{-10}{*}{3} & \multicolumn{1}{l|}{RAG+Cluster} & \cellcolor[HTML]{DAE8FC}4.11 & \cellcolor[HTML]{DAE8FC}4.16 & 4.49 & \cellcolor[HTML]{DAE8FC}4.37 & \multicolumn{1}{r|}{\cellcolor[HTML]{DAE8FC}4.23} & \cellcolor[HTML]{DAE8FC}3.83 & \cellcolor[HTML]{DAE8FC}4.18 & \cellcolor[HTML]{DAE8FC}4.54 & \cellcolor[HTML]{DAE8FC}4.11 & \multicolumn{1}{r|}{\cellcolor[HTML]{DAE8FC}3.69} & 3.08 & 3.80 & 4.40 & 2.87 & \multicolumn{1}{r|}{3.31} & 0.61 \\ \midrule
 & \multicolumn{1}{l|}{\textbf{\modelTopic}} & \cellcolor[HTML]{DAE8FC}4.15 & \cellcolor[HTML]{DAE8FC}4.08 & \cellcolor[HTML]{DAE8FC}4.45 & \cellcolor[HTML]{DAE8FC}4.37 & \multicolumn{1}{r|}{\cellcolor[HTML]{DAE8FC}\textbf{4.40}} & \cellcolor[HTML]{DAE8FC}\textbf{4.06} & \cellcolor[HTML]{DAE8FC}\textbf{4.20} & \cellcolor[HTML]{DAE8FC}4.54 & \cellcolor[HTML]{DAE8FC}\textbf{4.20} & \multicolumn{1}{r|}{\cellcolor[HTML]{DAE8FC}\textbf{3.94}} & \cellcolor[HTML]{DAE8FC}3.80 & \cellcolor[HTML]{DAE8FC}4.12 & \cellcolor[HTML]{DAE8FC}4.68 & \cellcolor[HTML]{DAE8FC}4.11 & \multicolumn{1}{r|}{\cellcolor[HTML]{DAE8FC}4.19} & 0.71 \\
 & \multicolumn{1}{l|}{\textbf{\modelAll}} & \cellcolor[HTML]{DAE8FC}\textbf{4.21} & \cellcolor[HTML]{DAE8FC}4.14 & \cellcolor[HTML]{DAE8FC}\textbf{4.48} & \cellcolor[HTML]{DAE8FC}\textbf{4.39} & \multicolumn{1}{r|}{\cellcolor[HTML]{DAE8FC}4.30} & 3.82 & \cellcolor[HTML]{DAE8FC}4.12 & \cellcolor[HTML]{DAE8FC}4.49 & 4.02 & \multicolumn{1}{r|}{3.68} & \cellcolor[HTML]{DAE8FC}\textbf{3.93} & \cellcolor[HTML]{DAE8FC}\textbf{4.18} & 4.58 & \cellcolor[HTML]{DAE8FC}4.07 & \multicolumn{1}{r|}{\cellcolor[HTML]{DAE8FC}4.21} & 0.69 \\
 & \multicolumn{1}{l|}{Long-Context} & 3.92 & \cellcolor[HTML]{DAE8FC}4.07 & \cellcolor[HTML]{DAE8FC}4.40 & 4.15 & \multicolumn{1}{r|}{4.14} & 3.48 & 4.04 & 4.43 & 3.70 & \multicolumn{1}{r|}{3.07} & \cellcolor[HTML]{DAE8FC}3.83 & \cellcolor[HTML]{DAE8FC}4.14 & 4.56 & \cellcolor[HTML]{DAE8FC}4.02 & \multicolumn{1}{r|}{\cellcolor[HTML]{DAE8FC}4.21} & 0.65 \\
 & \multicolumn{1}{l|}{RAG-\textit{All}} & 3.93 & \cellcolor[HTML]{DAE8FC}4.04 & \cellcolor[HTML]{DAE8FC}4.36 & \cellcolor[HTML]{DAE8FC}4.27 & \multicolumn{1}{r|}{\cellcolor[HTML]{DAE8FC}4.18} & 3.55 & 4.02 & 4.45 & 3.83 & \multicolumn{1}{r|}{3.30} & \cellcolor[HTML]{DAE8FC}3.79 & \cellcolor[HTML]{DAE8FC}4.16 & \cellcolor[HTML]{DAE8FC}4.64 & \cellcolor[HTML]{DAE8FC}4.02 & \multicolumn{1}{r|}{\cellcolor[HTML]{DAE8FC}4.21} & 0.66 \\
 & \multicolumn{1}{l|}{RAG-\textit{Doc}} & 3.96 & \cellcolor[HTML]{DAE8FC}3.99 & \cellcolor[HTML]{DAE8FC}4.31 & \cellcolor[HTML]{DAE8FC}4.34 & \multicolumn{1}{r|}{\cellcolor[HTML]{DAE8FC}4.24} & 3.64 & 4.05 & \cellcolor[HTML]{DAE8FC}4.51 & 3.87 & \multicolumn{1}{r|}{3.31} & \cellcolor[HTML]{DAE8FC}3.80 & \cellcolor[HTML]{DAE8FC}4.08 & \cellcolor[HTML]{DAE8FC}4.63 & \cellcolor[HTML]{DAE8FC}\textbf{4.15} & \multicolumn{1}{r|}{\cellcolor[HTML]{DAE8FC}4.14} & 0.66 \\
 & \multicolumn{1}{l|}{Hierarchical} & \cellcolor[HTML]{DAE8FC}4.05 & \cellcolor[HTML]{DAE8FC}\textbf{4.16} & \cellcolor[HTML]{DAE8FC}4.44 & \cellcolor[HTML]{DAE8FC}4.34 & \multicolumn{1}{r|}{\cellcolor[HTML]{DAE8FC}4.37} & 3.63 & 4.07 & \cellcolor[HTML]{DAE8FC}4.49 & 3.87 & \multicolumn{1}{r|}{3.44} & \cellcolor[HTML]{DAE8FC}3.80 & \cellcolor[HTML]{DAE8FC}4.15 & \cellcolor[HTML]{DAE8FC}\textbf{4.75} & \cellcolor[HTML]{DAE8FC}4.04 & \multicolumn{1}{r|}{\cellcolor[HTML]{DAE8FC}\textbf{4.28}} & 0.66 \\
 & \multicolumn{1}{l|}{Increm-\textit{All}} & 3.93 & \cellcolor[HTML]{DAE8FC}4.06 & \cellcolor[HTML]{DAE8FC}4.36 & \cellcolor[HTML]{DAE8FC}4.19 & \multicolumn{1}{r|}{\cellcolor[HTML]{DAE8FC}4.19} & 3.45 & 4.02 & 4.45 & 3.68 & \multicolumn{1}{r|}{3.24} & \cellcolor[HTML]{DAE8FC}3.82 & \cellcolor[HTML]{DAE8FC}4.12 & \cellcolor[HTML]{DAE8FC}4.66 & \cellcolor[HTML]{DAE8FC}4.09 & \multicolumn{1}{r|}{\cellcolor[HTML]{DAE8FC}4.09} & 0.65 \\
 & \multicolumn{1}{l|}{Increm-\textit{Topic}} & \cellcolor[HTML]{DAE8FC}4.05 & \cellcolor[HTML]{DAE8FC}4.08 & 4.25 & \cellcolor[HTML]{DAE8FC}4.34 & \multicolumn{1}{r|}{\cellcolor[HTML]{DAE8FC}4.33} & 3.90 & \cellcolor[HTML]{DAE8FC}4.18 & \cellcolor[HTML]{DAE8FC}\textbf{4.56} & \cellcolor[HTML]{DAE8FC}4.12 & \multicolumn{1}{r|}{3.67} & 3.61 & \cellcolor[HTML]{DAE8FC}3.98 & 4.39 & 3.77 & \multicolumn{1}{r|}{3.93} & 0.69 \\
 & \multicolumn{1}{l|}{Cluster} & \cellcolor[HTML]{DAE8FC}3.97 & \cellcolor[HTML]{DAE8FC}4.01 & 4.17 & 4.05 & \multicolumn{1}{r|}{4.15} & 3.70 & 4.00 & 4.32 & 3.82 & \multicolumn{1}{r|}{3.48} & 3.01 & 3.56 & 4.07 & 3.02 & \multicolumn{1}{r|}{3.37} & 0.66 \\
\multirow{-10}{*}{4} & \multicolumn{1}{l|}{RAG+Cluster} & \cellcolor[HTML]{DAE8FC}4.15 & \cellcolor[HTML]{DAE8FC}3.97 & \cellcolor[HTML]{DAE8FC}4.32 & \cellcolor[HTML]{DAE8FC}4.23 & \multicolumn{1}{r|}{\cellcolor[HTML]{DAE8FC}4.26} & 3.83 & 4.09 & \cellcolor[HTML]{DAE8FC}4.54 & 4.04 & \multicolumn{1}{r|}{3.56} & 3.29 & 3.60 & 4.17 & 3.24 & \multicolumn{1}{r|}{3.50} & 0.66 \\ \midrule
 & \multicolumn{1}{l|}{\textbf{\modelTopic}} & \cellcolor[HTML]{DAE8FC}\textbf{4.16} & \cellcolor[HTML]{DAE8FC}\textbf{4.14} & \cellcolor[HTML]{DAE8FC}4.36 & \cellcolor[HTML]{DAE8FC}\textbf{4.28} & \multicolumn{1}{r|}{\cellcolor[HTML]{DAE8FC}\textbf{4.40}} & \cellcolor[HTML]{DAE8FC}\textbf{4.05} & \cellcolor[HTML]{DAE8FC}\textbf{4.25} & \cellcolor[HTML]{DAE8FC}\textbf{4.58} & \cellcolor[HTML]{DAE8FC}\textbf{4.27} & \multicolumn{1}{r|}{\cellcolor[HTML]{DAE8FC}\textbf{3.89}} & \cellcolor[HTML]{DAE8FC}4.04 & \cellcolor[HTML]{DAE8FC}4.37 & \cellcolor[HTML]{DAE8FC}4.77 & \cellcolor[HTML]{DAE8FC}4.33 & \multicolumn{1}{r|}{\cellcolor[HTML]{DAE8FC}4.49} & 0.75 \\
 & \multicolumn{1}{l|}{\textbf{\modelAll}} & \cellcolor[HTML]{DAE8FC}4.07 & \cellcolor[HTML]{DAE8FC}4.03 & \cellcolor[HTML]{DAE8FC}\textbf{4.37} & \cellcolor[HTML]{DAE8FC}4.25 & \multicolumn{1}{r|}{\cellcolor[HTML]{DAE8FC}4.29} & 3.79 & 4.09 & 4.48 & 3.99 & \multicolumn{1}{r|}{3.59} & \cellcolor[HTML]{DAE8FC}4.05 & \cellcolor[HTML]{DAE8FC}4.35 & \cellcolor[HTML]{DAE8FC}4.81 & \cellcolor[HTML]{DAE8FC}4.38 & \multicolumn{1}{r|}{\cellcolor[HTML]{DAE8FC}\textbf{4.54}} & 0.73 \\
 & \multicolumn{1}{l|}{Long-Context} & 3.79 & \cellcolor[HTML]{DAE8FC}3.99 & \cellcolor[HTML]{DAE8FC}4.20 & \cellcolor[HTML]{DAE8FC}4.14 & \multicolumn{1}{r|}{3.99} & 3.47 & 4.01 & 4.44 & 3.69 & \multicolumn{1}{r|}{3.05} & \cellcolor[HTML]{DAE8FC}4.07 & 4.27 & 4.74 & 4.23 & \multicolumn{1}{r|}{\cellcolor[HTML]{DAE8FC}4.40} & 0.70 \\
 & \multicolumn{1}{l|}{RAG-\textit{All}} & 3.90 & \cellcolor[HTML]{DAE8FC}3.91 & \cellcolor[HTML]{DAE8FC}4.23 & \cellcolor[HTML]{DAE8FC}4.14 & \multicolumn{1}{r|}{4.15} & 3.59 & 4.00 & 4.44 & 3.72 & \multicolumn{1}{r|}{3.33} & \cellcolor[HTML]{DAE8FC}\textbf{4.17} & \cellcolor[HTML]{DAE8FC}\textbf{4.44} & \cellcolor[HTML]{DAE8FC}4.87 & \cellcolor[HTML]{DAE8FC}4.36 & \multicolumn{1}{r|}{\cellcolor[HTML]{DAE8FC}4.52} & 0.70 \\
 & \multicolumn{1}{l|}{RAG-\textit{Doc}} & \cellcolor[HTML]{DAE8FC}3.93 & \cellcolor[HTML]{DAE8FC}3.98 & \cellcolor[HTML]{DAE8FC}4.30 & \cellcolor[HTML]{DAE8FC}4.23 & \multicolumn{1}{r|}{4.14} & 3.61 & 4.05 & 4.47 & 3.81 & \multicolumn{1}{r|}{3.31} & \cellcolor[HTML]{DAE8FC}4.05 & \cellcolor[HTML]{DAE8FC}4.43 & \cellcolor[HTML]{DAE8FC}4.84 & \cellcolor[HTML]{DAE8FC}\textbf{4.50} & \multicolumn{1}{r|}{\cellcolor[HTML]{DAE8FC}4.50} & 0.70 \\
 & \multicolumn{1}{l|}{Hierarchical} & 3.90 & \cellcolor[HTML]{DAE8FC}3.96 & \cellcolor[HTML]{DAE8FC}4.23 & \cellcolor[HTML]{DAE8FC}4.16 & \multicolumn{1}{r|}{4.09} & 3.60 & 4.09 & 4.48 & 3.85 & \multicolumn{1}{r|}{3.38} & \cellcolor[HTML]{DAE8FC}4.16 & \cellcolor[HTML]{DAE8FC}4.43 & \cellcolor[HTML]{DAE8FC}\textbf{4.87} & \cellcolor[HTML]{DAE8FC}4.52 & \multicolumn{1}{r|}{\cellcolor[HTML]{DAE8FC}4.52} & 0.70 \\
 & \multicolumn{1}{l|}{Increm-\textit{All}} & 3.80 & \cellcolor[HTML]{DAE8FC}4.07 & \cellcolor[HTML]{DAE8FC}4.23 & \cellcolor[HTML]{DAE8FC}4.09 & \multicolumn{1}{r|}{4.04} & 3.41 & 3.95 & 4.40 & 3.56 & \multicolumn{1}{r|}{3.09} & \cellcolor[HTML]{DAE8FC}4.08 & 4.26 & \cellcolor[HTML]{DAE8FC}4.76 & \cellcolor[HTML]{DAE8FC}4.36 & \multicolumn{1}{r|}{\cellcolor[HTML]{DAE8FC}4.37} & 0.68 \\
 & \multicolumn{1}{l|}{Increm-\textit{Topic}} & \cellcolor[HTML]{DAE8FC}4.04 & \cellcolor[HTML]{DAE8FC}4.10 & \cellcolor[HTML]{DAE8FC}4.21 & \cellcolor[HTML]{DAE8FC}4.16 & \multicolumn{1}{r|}{\cellcolor[HTML]{DAE8FC}4.16} & 3.84 & 4.16 & \cellcolor[HTML]{DAE8FC}4.55 & 4.06 & \multicolumn{1}{r|}{3.56} & 3.69 & 3.98 & 4.51 & 3.88 & \multicolumn{1}{r|}{3.99} & 0.73 \\
 & \multicolumn{1}{l|}{Cluster} & 3.86 & 3.90 & 3.96 & 3.98 & \multicolumn{1}{r|}{4.12} & 3.68 & 4.01 & 4.36 & 3.87 & \multicolumn{1}{r|}{3.42} & 3.14 & 3.60 & 4.19 & 3.31 & \multicolumn{1}{r|}{3.52} & 0.71 \\
\multirow{-10}{*}{5} & \multicolumn{1}{l|}{RAG+Cluster} & \cellcolor[HTML]{DAE8FC}4.03 & \cellcolor[HTML]{DAE8FC}3.96 & \cellcolor[HTML]{DAE8FC}4.19 & \cellcolor[HTML]{DAE8FC}4.19 & \multicolumn{1}{r|}{\cellcolor[HTML]{DAE8FC}4.21} & 3.79 & 4.10 & 4.50 & 4.04 & \multicolumn{1}{r|}{3.57} & 3.46 & 3.86 & 4.40 & 3.62 & \multicolumn{1}{r|}{3.69} & 0.72 \\ \bottomrule
\end{tabular}
\caption{\label{appendix:table:llm_debate} Interest, Coherence, Relevance, Coverage, and Diversity scores from Prometheus for summaries, topic paragraphs, and topics on DebateQFS. Best scores are \textbf{bold}, significant scores in \colorbox{myblue}{blue} (2-sample $t$-test, $p<0.05$)}
\end{table*}
\begin{table*}[]
\small
\centering
\setlength{\tabcolsep}{3.5pt}
\renewcommand{\arraystretch}{0.8}
\begin{tabular}{@{}cl|ccccc@{}}
\toprule
\textbf{\# Topics} & \textbf{Model} & \multicolumn{1}{l}{\textbf{\# Input Tokens}} & \multicolumn{1}{l}{\textbf{\# Output Tokens}} & \multicolumn{1}{l}{\textbf{\# LLM Calls}} & \multicolumn{1}{l}{\textbf{Cost (GPT-4)}} & \multicolumn{1}{l}{\textbf{Time (seconds)}} \\ \midrule
\multirow{3}{*}{2} & \modelTopic & 21383.08 & 3412.02 & 25.45 & 0.32 & 117.60 \\
 & Hierarchical & 31130.02 & 2536.66 & 13.15 & 0.39 & 83.13 \\
 & Incremental-\textit{Topic} & 59010.66 & 6115.04 & 15.15 & 0.77 & 214.39 \\ \midrule
\multirow{3}{*}{3} & \modelTopic & 30208.20 & 5040.38 & 37.38 & 0.45 & 149.54 \\
 & Hierarchical & 31144.83 & 2649.78 & 13.15 & 0.39 & 68.60 \\
 & Incremental-\textit{Topic} & 61344.07 & 8442.54 & 16.15 & 0.87 & 197.33 \\ \midrule
\multirow{3}{*}{4} & \modelTopic & 38286.40 & 6440.23 & 47.91 & 0.58 & 163.91 \\
 & Hierarchical & 31144.31 & 2740.31 & 13.15 & 0.39 & 88.75 \\
 & Incremental-\textit{Topic} & 62877.46 & 9966.45 & 17.15 & 0.93 & 312.55 \\ \midrule
\multirow{3}{*}{5} & \modelTopic & 47008.59 & 7918.92 & 58.94 & 0.71 & 186.32 \\
 & Hierarchical & 31160.88 & 2850.24 & 13.15 & 0.40 & 61.70 \\
 & Incremental-\textit{Topic} & 64893.95 & 11965.84 & 18.15 & 1.01 & 262.07 \\ \bottomrule
\end{tabular}
\caption{\label{appendix:table:cost_cqa} Number of LLM input/output tokens, LLM calls, GPT-4 Cost (USD), and Time (seconds) needed to run inference on a single DFQS example on ConflictingQA with the top-3 models. We report 5 runs and 20 examples.}
\end{table*}

\begin{table*}[]
\small
\centering
\setlength{\tabcolsep}{3.5pt}
\renewcommand{\arraystretch}{0.8}
\begin{tabular}{@{}cl|ccccc@{}}
\toprule
\multicolumn{1}{l}{\textbf{Dataset}} & \textbf{Model} & \multicolumn{1}{l}{\textbf{\# Input Tokens}} & \multicolumn{1}{l}{\textbf{\# Output Tokens}} & \multicolumn{1}{l}{\textbf{\# LLM Calls}} & \multicolumn{1}{l}{\textbf{Cost (GPT-4)}} & \multicolumn{1}{l}{\textbf{Time (seconds)}} \\ \midrule
\multirow{3}{*}{2} & \modelTopic & 17183.75 & 2722.40 & 20.30 & 0.25 & 94.81 \\
 & Hierarchical & 19181.59 & 2040.39 & 10.25 & 0.25 & 63.68 \\
 & Incremental-\textit{Topic} & 41656.87 & 5062.44 & 12.25 & 0.57 & 182.19 \\ 
 \midrule
\multirow{3}{*}{3} & \modelTopic & 24801.22 & 4136.12 & 30.40 & 0.37 & 126.83 \\
 & Hierarchical & 19182.58 & 2141.91 & 10.25 & 0.26 & 53.32 \\
 & Incremental-\textit{Topic} & 43119.51 & 6532.92 & 13.25 & 0.63 & 152.44 \\ \midrule
\multirow{3}{*}{4} & \modelTopic & 30677.67 & 5037.31 & 38.00 & 0.46 & 120.64 \\
 & Hierarchical & 19203.30 & 2253.17 & 10.25 & 0.26 & 73.35 \\
 & Incremental-\textit{Topic} & 43922.02 & 7327.88 & 14.25 & 0.66 & 241.54 \\ \midrule
\multirow{3}{*}{5} & \modelTopic & 36988.41 & 6049.93 & 46.09 & 0.55 & 139.71 \\
 & Hierarchical & 19211.74 & 2356.01 & 10.25 & 0.26 & 49.41 \\
 & Incremental-\textit{Topic} & 45113.12 & 8504.59 & 15.25 & 0.71 & 186.40 \\ \bottomrule
\end{tabular}
\caption{\label{appendix:table:cost_debate} Number of LLM input/output tokens, LLM calls, GPT-4 Cost (USD), and Time (seconds) needed to run inference on a single DFQS example on DebateQFS with the top-3 models. We report 5 runs and 20 examples.}
\end{table*}

\begin{table*}[]
\small
\centering
\setlength{\tabcolsep}{3.5pt}
\renewcommand{\arraystretch}{0.8}
\begin{tabular}{@{}cl|ccccc@{}}
\toprule
\multicolumn{1}{l}{\textbf{\# Topics}} & \textbf{Model} & \multicolumn{1}{l}{\textbf{\# Input Tokens}} & \multicolumn{1}{l}{\textbf{\# Output Tokens}} & \multicolumn{1}{l}{\textbf{\# LLM Calls}} & \multicolumn{1}{l}{\textbf{Cost (GPT-4)}} & \multicolumn{1}{l}{\textbf{Time (seconds)}} \\ 
\midrule
\multirow{3}{*}{ConflictingQA} & \modelTopic & 47008.59 & 7918.92 & 58.94 & 0.71 & 186.32 \\
 & \modelTopic Pick All & 53733.70 & 9596.75 & 71.75 & 0.83 & 303.13 \\
 & Hierarchical-\emph{Topic} & 168160.85 & 7485.50 & 66.75 & 1.91 & 210.80 \\ \midrule
\multirow{3}{*}{DebateQFS} & \modelTopic & 36988.41 & 6049.93 & 46.09 & 0.55 & 139.71 \\
& \modelTopic Pick All & 43098.85 & 7612.45 & 57.25 & 0.66 & 242.35 \\
& Hierarchical-\emph{Topic} & 105237.25 & 5278.35 & 52.25 & 1.21 & 139.96 \\ \bottomrule
\end{tabular}
\caption{\label{appendix:table:cost_weird} Number of LLM input/output tokens, LLM calls, GPT-4 Cost (USD), and Time (seconds) needed to run inference on a single DFQS example on ConflictingQA and DebateQFS with \modelTopic, the version of \modelTopic with no Moderator, and the version of Hierarchical merging that runs on each topic paragraph ($m=5$). We report 5 runs and 20 examples.}
\end{table*}

\clearpage
\hypersetup{
    colorlinks=true, 
    linkcolor=black, 
    citecolor=black, 
    urlcolor=black 
}

\begin{summary}[title={\modelTopic Summary: Are Audiobooks Considered Real Reading? (ConflictingQA)}, label=summary1]
\textbf{Topic 1: Audiobooks and cognitive engagement}\\
Audiobooks provide cognitive benefits similar to traditional reading, such as improved vocabulary, comprehension, and critical thinking skills, and engage the same cognitive processes as reading print books [1][2][3]. While listening to audiobooks, children and adults can visualize, make predictions, and synthesize information, indicating active cognitive engagement, although the lack of visual stimulation may affect the engagement for some individuals [3][4][8]. High-quality production and narration in audiobooks can enhance the cognitive experience, making them a valid form of reading for people with visual impairments, dyslexia, or those who multitask [7][9][11].\\
\\
\textbf{Topic 2: Audiobooks accessibility and inclusivity}\\
Audiobooks enhance accessibility and inclusivity by supporting children with different learning needs, allowing individuals with ADHD, dyslexia, and other attention disorders to enjoy literature and improve comprehension [1][2][11]. They enable multitasking and comprehension for various audiences, including those with visual impairments or who find traditional reading challenging, and are cost-effective to produce, increasing their accessibility [7][12]. Despite some cultural biases that may undervalue audiobooks compared to printed text, they are considered a valid form of reading, especially for those who may not engage with traditional reading methods [9][8].\\
\\
\textbf{Topic 3: Audiobooks versus physical reading}\\
Audiobooks provide similar cognitive benefits to traditional reading, such as improved vocabulary and comprehension, and engage the same cognitive processes as reading with eyes, making them a valid form of reading [1][2][3]. They are particularly beneficial for those with different learning needs or physical limitations, allowing for a rich, accessible, and multitask-friendly reading experience [3][6][7][11]. Despite some cultural perceptions that audiobooks are less valuable than printed text, they are increasingly popular, economically significant, and considered a legitimate form of storytelling and information absorption [8][9][12].
\end{summary}

\begin{summary}[title={\modelTopic Summary: Is infinite scrolling a good web design technique? (ConflictingQA)}, label=summary2]
\textbf{Topic 1: User Engagement Increase}\\
Infinite scrolling can potentially increase user engagement and time spent on a site by continuously loading new content, which exploits users' automatic behavior and keeps them engaged [9]. However, it has been found to decrease user engagement in some cases, such as on Etsy, and can negatively impact users with disabilities and mental health, leading to a high cognitive load and potential mental health issues [1][5][8]. Additionally, infinite scrolling can lead to control issues and user frustration due to less controllable pages and jumping glitches [6].\\
\\
\textbf{Topic 2: Content Accessibility Concerns}\\
Infinite scrolling can lead to content accessibility issues, as it breaks the expected behavior of scrollbars and makes it difficult for users to gauge the length of the page, and it poses significant challenges for users with assistive technologies, often excluding footers and making navigation stressful [1][6][7]. While it can keep users engaged on eCommerce platforms, it has been associated with increased stress levels and negative mental health outcomes, particularly in young social media users [3][6][8]. Moreover, strategies like role='feed' have failed to address these accessibility problems effectively [5].\\
\\
\textbf{Topic 3: Mental Health Implications}\\
Infinite scrolling can exploit human psychological phenomena such as automaticity, leading to behaviors like doom-scrolling that may contribute to mental health issues by causing users to lose track of time and continue scrolling unconsciously [9]. The design can also induce stress by preventing users from reaching a perceived end, leading to information overload, and overwhelming them with choices, which can result in frustration, anxiety, and a reduced motivation to engage with content [6][7]. However, some studies suggest that engaging in mindful scrolling practices can mitigate these negative mental health outcomes, indicating that the impact of infinite scrolling may vary based on user behavior [8].
\end{summary}

\clearpage

\begin{summary}[title={\modelTopic Summary: Is EU expansion and EU membership itself a good idea? (DebateQFS)}, label=summary3]
\textbf{Topic 1: Economic gains from accession}\\
The 1997 study by the Centre for Economic Policy Research predicted economic gains for both the EU-15 and new Central and Eastern European members, with an estimated €10 billion and €23 billion increase respectively [1]. However, concerns about high budget and trade deficits in accession countries, such as Estonia and Hungary, and the potential for increased unemployment and social costs, suggest that EU expansion could also exacerbate economic disparities and put fiscal pressure on both new and existing members [5][6]. Additionally, the enlargement is expected to shift regional funds towards new members, potentially reducing support for poorer regions within the EU(15) and necessitating a significant increase in the EU's regional funding budget to address growing economic and social needs [6].\\
\\
\textbf{Topic 2: EU enlargement political challenges}\\
EU enlargement is seen as beneficial, with studies indicating potential GDP growth for new and existing members, strategic interests in stabilizing regions like the Western Balkans and Turkey, and necessary controls in place to manage economic migration and regional subsidies [1][2][6]. However, public opposition in some member states, the slow process of enlargement due to political complexities, and concerns over social contradictions and international conflicts [2][5][6] present significant challenges. The Treaty of Lisbon is deemed necessary for further enlargement, although there are differing opinions on whether its ratification should delay the process [4].\\
\\
\textbf{Topic 3: Regional disparities and funding}\\
EU expansion has been estimated to bring economic gains for both old and new member states, with the EU-15 seeing a €10 billion increase and new Central and Eastern European members gaining €23 billion [1]. However, regional disparities pose challenges, as unemployment rates have risen in accession countries and the wealth gap between regions may widen, with 98 million inhabitants in applicant states living in regions with GDP less than 75\% of the EU average [5][6]. Despite the potential for increased regional funding, there are concerns that existing poorer regions within the EU(15) may receive less support as a result of the expansion [6].
\end{summary}

\begin{summary}[title={\modelTopic Summary: Is going to law school a good idea? (DebateQFS)}, label=summary4]

\textbf{Topic 1: Law School ROI Analysis}\\
Attending law school can lead to a variety of career opportunities and the acquisition of valuable skills, with some graduates finding employment directly from campus and others benefiting from practical skills-oriented courses [4][5][6]. However, the financial burden of law school is significant, with many students accruing substantial debt, facing uncertain job markets, and questioning the return on investment, especially if they do not graduate from top-tier schools or are not at the top of their class [8][10][15][19]. Despite the potential for high starting salaries in some legal jobs, the competitiveness of the market and the cost of tuition may not justify the investment for all students, particularly when considering the psychological toll and the oversupply of law graduates [12][14][16].\\
\\
\textbf{Topic 2: Legal Career Job Market}\\
The legal job market presents a mixed outlook, with some documents indicating an increase in law firm hiring practices and a demand for legal services in certain areas, while others highlight the oversaturation of law graduates, underemployment, and the potential for job dissatisfaction and misleading employment statistics from law schools [4][17][8][10][11][18][19]. Graduates from prestigious law schools or those in the top of their class may have better job prospects and higher starting salaries, but many face significant debt and struggle to find well-paying jobs to manage that debt [16][19]. The rise of legal process outsourcing and the hiring of law school graduates directly by companies suggest evolving trends in the legal job market that could affect future employment opportunities for lawyers [6][5].\\
\\
\textbf{Topic 3: Law Education Value Debate}\\
Law school provides a range of non-monetary benefits, such as personal growth, maturity, and the development of transferable skills like critical thinking and argumentation, which are applicable in various fields beyond traditional legal practice [3][9]. However, the financial implications of law school, including high tuition costs, significant student debt, and an uncertain job market, challenge the notion that a legal education is a sound financial investment for all students [13][14][17][19]. Despite these concerns, there is a demand for legal professionals, and law school can prepare graduates for diverse career paths, including roles that address complex societal challenges and ensure access to justice [2][7][16].
\end{summary}

\end{document}